\documentclass[11pt]{article}

\usepackage[preprint]{acl}

\usepackage{times}
\usepackage{latexsym}

\usepackage[T1]{fontenc}
\usepackage[utf8]{inputenc}

\usepackage{microtype}

\usepackage{inconsolata}

\usepackage{graphicx}
\usepackage{subcaption}

\usepackage{amsmath}
\usepackage{amssymb}
\usepackage{booktabs}
\usepackage{multirow}
\usepackage[table]{xcolor}
\usepackage[most]{tcolorbox}
\usepackage{makecell}
\usepackage{diagbox}
\usepackage{hyperref}

\title{Breaking Failure Cascades: Step-Aware Reinforcement Learning for Medical Multimodal Reasoning}

\author{
  \textbf{Junha Jung\textsuperscript{1,6}\thanks{Equal contribution.}},
  \textbf{Minbyul Jeong\textsuperscript{2,}}\footnotemark[1],
  \textbf{Suhyeon Lim\textsuperscript{4}},
  \textbf{Sungwook Jung\textsuperscript{1}},
\\
  \textbf{Jaehoon Yun\textsuperscript{5}},
  \textbf{Taeyun Roh\textsuperscript{1}},
  \textbf{Mujeen Sung\textsuperscript{3}\thanks{Corresponding authors.}},
  \textbf{Jaewoo Kang\textsuperscript{1,6,}}\footnotemark[2]
\\
  \textsuperscript{1}Korea University,
  \textsuperscript{2}Meridian Intelligence Global Inc.,
  \textsuperscript{3}Kyung Hee University,
  \textsuperscript{4}KAIST,
\\
  \textsuperscript{5}Hanyang University College of Medicine,
  \textsuperscript{6}AIGEN Sciences
\\
  \texttt{goodjungjun@korea.ac.kr}
\\
  \texttt{mujeensung@khu.ac.kr}, \texttt{kangj@korea.ac.kr}
}

\begin{document}
\maketitle
\begin{abstract}
Recent multimodal large language models have shown great promise in clinical image reasoning, but existing post-training pipelines remain predominantly outcome-centric, relying on final answer correctness or sequence-level preferences.
This suffers from sparse credit assignment, making it difficult to optimize the reasoning process essential for clinical applications. Our analysis reveals that cascading errors from early-stage reasoning failures are a leading cause of incorrect predictions in medical visual question answering (VQA) benchmarks.
% Motivated by this, we propose \textbf{M}edical \textbf{R}easoning-aware \textbf{P}olicy \textbf{O}ptimization \textbf{(MRPO)}, an RL algorithm that incorporates step-wise process rewards. When the final answer is incorrect, MRPO assigns exponentially larger penalties to tokens in earlier invalid reasoning steps, breaking failure cascades without compromising successful paths. Across three multimodal LLM backbones, MRPO consistently outperforms standard GRPO and recent RL alternative while reducing early-stage reasoning failures from 64.0\% to 13.0\%, addressing the cascading failure problem.
Motivated by this, we propose \textbf{M}edical \textbf{R}easoning-aware \textbf{P}olicy \textbf{O}ptimization \textbf{(MRPO)}, an RL algorithm that incorporates step-wise process rewards. When the final answer is incorrect, MRPO assigns exponentially larger penalties to tokens in earlier invalid reasoning steps, breaking failure cascades without compromising successful paths. Across four multimodal LLM backbones, MRPO consistently outperforms standard GRPO and a recent RL baseline, and on Qwen3-VL-8B-Thinking even surpasses substantially larger medical MLLMs such as HuatuoGPT-Vision-34B by 4.59 points. Moreover, MRPO reduces early-stage reasoning failures from 58.6\% to 13.4\%, showing that targeted mitigation of cascading failures improves both reasoning quality and final answer accuracy. Our code is available \href{https://github.com/dmis-lab/MRPO}{here}.

\end{abstract}

\section{Introduction}
% Recent advances in multimodal large language models (MLLMs) have extended their capabilities beyond general vision-language tasks to clinical image reasoning~\cite{chen2024huatuogptvisioninjectingmedicalvisual, lasateam2025lingshugeneralistfoundationmodel}. Clinical or medical image reasoning requires multi-dimensional reasoning that spans medical image interpretation, identification of salient evidence, and integration of medical knowledge. To improve such reasoning abilities, prior work has explored chain-of-thought fine-tuning on curated reasoning traces~\cite{sun2025chirono1ignitingmultimodallarge, Kim2025} as well as reinforcement learning (RL) post-training~\cite{fan2025chestxreasoneradvancingradiologyfoundation, lai2025medr1reinforcementlearninggeneralizable}.
Recent advances in multimodal large language models (MLLMs) have extended their capabilities to clinical image reasoning~\cite{chen2024huatuogptvisioninjectingmedicalvisual, lasateam2025lingshugeneralistfoundationmodel}, where prior work has explored chain-of-thought fine-tuning on curated reasoning traces~\cite{sun2025chirono1ignitingmultimodallarge, Kim2025} as well as reinforcement learning (RL) post-training~\cite{fan2025chestxreasoneradvancingradiologyfoundation, lai2025medr1reinforcementlearninggeneralizable}.
% However, most post-training techniques for medical MLLMs remain outcome-centric, providing supervision primarily through final answer correctness or sequence-level preferences~\cite{huang2025medvlthinkersimplebaselinesmultimodal, liu2025breakingrewardcollapseadaptive}. This approach inherently suffers from sparse credit assignment, as the learning signal cannot identify which intermediate steps caused the failure when a long reasoning trajectory fails~\cite{mu2025medcegreinforcingverifiablemedical, xie2025capoenhancingllmreasoning}. The limitation is particularly severe in free-form generation settings, where rewards are sparse and delayed, typically observed only after the entire response is generated~\cite{chaudhari2024rlhfdecipheredcriticalanalysis, lightman2023letsverifystepstep}. Since real clinical environments predominantly involve open-ended queries, addressing this limitation is essential for practical deployment of medical MLLMs.
However, most post-training techniques for medical MLLMs remain outcome-centric, supervising primarily through final answer correctness or sequence-level preferences~\cite{huang2025medvlthinkersimplebaselinesmultimodal, liu2025breakingrewardcollapseadaptive}. This inherently suffers from sparse credit assignment problem, when a reasoning trajectory fails, the learning signal cannot identify which intermediate steps caused the failure~\cite{mu2025medcegreinforcingverifiablemedical, xie2025capoenhancingllmreasoning}. This is particularly severe in free-form generation, where rewards are sparse and delayed, typically observed only after the entire response is generated~\cite{chaudhari2024rlhfdecipheredcriticalanalysis}. Consequently, learning signals are distributed uniformly across all tokens, making it difficult for models to learn how to reason correctly step by step. Since real clinical environments predominantly involve open-ended queries, addressing this limitation is essential for practical deployment of medical MLLMs.

Beyond sparse credit assignment, we identify a structural failure mode in medical VQA reasoning: early-stage reasoning failure tends to propagate and accumulate, forming failure cascades that drive final prediction errors. We analyze sentence-level reasoning traces from existing MLLMs on open-ended medical VQA benchmarks. The analysis reveals a strong correlation between the position of the first invalid reasoning step and the likelihood of an incorrect final answer. Once an early reasoning step becomes invalid, subsequent steps are significantly more likely to fail, even when later reasoning capabilities would otherwise suffice.

% To mitigate this issue, we propose a new RL algorithm, \textbf{M}edical \textbf{R}easoning-aware \textbf{P}olicy \textbf{O}ptimization \textbf{(MRPO)}. The key idea of MRPO is to reshape the GRPO~\cite{shao2024deepseekmathpushinglimitsmathematical}-based advantage to assign larger penalties to tokens in earlier failed reasoning steps, thereby reducing early-stage failures and preventing subsequent error accumulation. To this end, MRPO leverages an answer reward and a step-wise reasoning process reward. The step-wise reasoning process reward is computed by an external judge model that evaluates each reasoning step as valid or invalid. Based on these reward signals, when the answer reward indicates an incorrect prediction, we reshape the advantage to assign exponentially larger penalties to earlier invalid reasoning steps. This preserves successful reasoning trajectories while increasing the probability of generating correct tokens at the first invalid step in failed traces. As a result, MRPO corrects early-stage reasoning failures and encourages valid reasoning in later steps, ultimately improving both answer accuracy and reasoning quality.

To mitigate this, we propose \textbf{M}edical \textbf{R}easoning-aware \textbf{P}olicy \textbf{O}ptimization \textbf{(MRPO)}, an RL algorithm that reshapes the GRPO~\cite{shao2024deepseekmathpushinglimitsmathematical}-based advantage to assign larger penalties to tokens in earlier failed reasoning steps. MRPO leverages an answer reward and a step-wise reasoning process reward computed by an external judge model that evaluates each step as valid or invalid. When the answer reward indicates an incorrect prediction, we reshape the advantage to assign exponentially larger penalties to earlier invalid steps, preserving successful reasoning trajectories while increasing the probability of generating correct tokens at the first invalid step in failed traces. As a result, MRPO corrects early-stage reasoning failures and encourages valid reasoning in later steps, improving both answer accuracy and reasoning quality.

To demonstrate its effectiveness, we apply MRPO to four multimodal LLM backbones, Qwen2.5-VL-7B-Instruct~\cite{Qwen2.5-VL}, Qwen3-VL-8B-Instruct~\cite{Qwen3-VL}, Qwen3-VL-8B-Thinking~\cite{Qwen3-VL}, and InternVL3-8B-Instruct~\cite{zhu2025internvl3exploringadvancedtraining}, on diverse open-ended medical VQA benchmarks. With only 13K training samples, MRPO consistently achieves the highest average performance across all four backbones, outperforming GRPO and the recent RL baseline GDPO~\cite{liu2026gdpogrouprewarddecouplednormalization}. On Qwen3-VL-8B-Thinking, MRPO outperforms larger medical MLLMs including HuatuoGPT-Vision-34B by 4.59 points, suggesting that targeted reasoning supervision can be a competitive alternative to large-scale medical instruction tuning. Moreover, reasoning failure analysis shows that MRPO substantially improves the reasoning failure pattern, reducing early-stage failures from 58.6\% to 13.4\% and mitigating downstream failure accumulation.

Our contributions are summarized as follows:
\begin{itemize}
\item Through systematic analysis of sentence-level reasoning traces on open-ended medical VQA, we identify cascading failures as a dominant cause of incorrect predictions, where early-stage reasoning failures propagate and accumulate to derail the final answer.
\item We propose MRPO, a GRPO-based RL algorithm that incorporates step-wise process rewards and reshapes advantages with exponentially larger penalties on earlier failed steps, directly targeting cascading failures.
\item Across four backbones, MRPO consistently achieves the highest average performance over standard GRPO and GDPO. On Qwen3-VL-8B-Thinking, it outperforms larger medical MLLMs such as HuatuoGPT-Vision-34B by 4.59 points. Moreover, MRPO improves the reasoning failure pattern, reducing early-stage failures from 58.6\% to 13.4\% and mitigating downstream failure accumulation.
\end{itemize}

\section{Related Works}
\label{sec:related_works}

Multimodal large language models~\cite{vteam2025glm45vglm41vthinkingversatilemultimodal, zhu2025internvl3exploringadvancedtraining} have been adapted to medical vision-language tasks~\cite{li2023llavamedtraininglargelanguageandvision, sellergren2025medgemmatechnicalreport}, evolving from supervised fine-tuning~\cite{sun2025chirono1ignitingmultimodallarge, Kim2025} to GRPO-based RL~\cite{lai2025medr1reinforcementlearninggeneralizable, pan2025medvlmr1incentivizingmedicalreasoning, su2025gmaivlr1harnessingreinforcementlearning} following DeepSeek-R1~\cite{Guo_2025}. To move beyond final-answer supervision, recent work evaluates individual reasoning steps and uses them as process rewards in RL~\cite{fan2025chestxreasoneradvancingradiologyfoundation, zhi2025medgr2breakingdatabarrier}, as inference-time verifiers~\cite{yun2025medprmmedicalreasoningmodels}, or as offline preferences~\cite{yang2026medreflmedicalreasoningenhancement}. However, none redistributes the learning signal by where a failure occurs, and thus cannot directly correct the failed step.
This is studied more directly in general-domain reasoning, where token-level credit assignment such as FSPO~\cite{li2025reasoningmodelshallucinatemore} and CAPO~\cite{xie2025capoenhancingllmreasoning} allocates learning signals selectively across the trajectory, but penalizes invalid steps regardless of their position. MRPO combines this token-level credit assignment with the medical step-level evaluation. By assigning exponentially stronger penalties to earlier invalid steps, MRPO corrects early reasoning failures before they cascade and steers the model toward valid reasoning. A detailed overview is in Appendix~\ref{app:related_work}.

\begin{figure*}[t]
  \centering

  \begin{subfigure}[t]{0.48\textwidth}
    \centering
    \includegraphics[width=\linewidth]{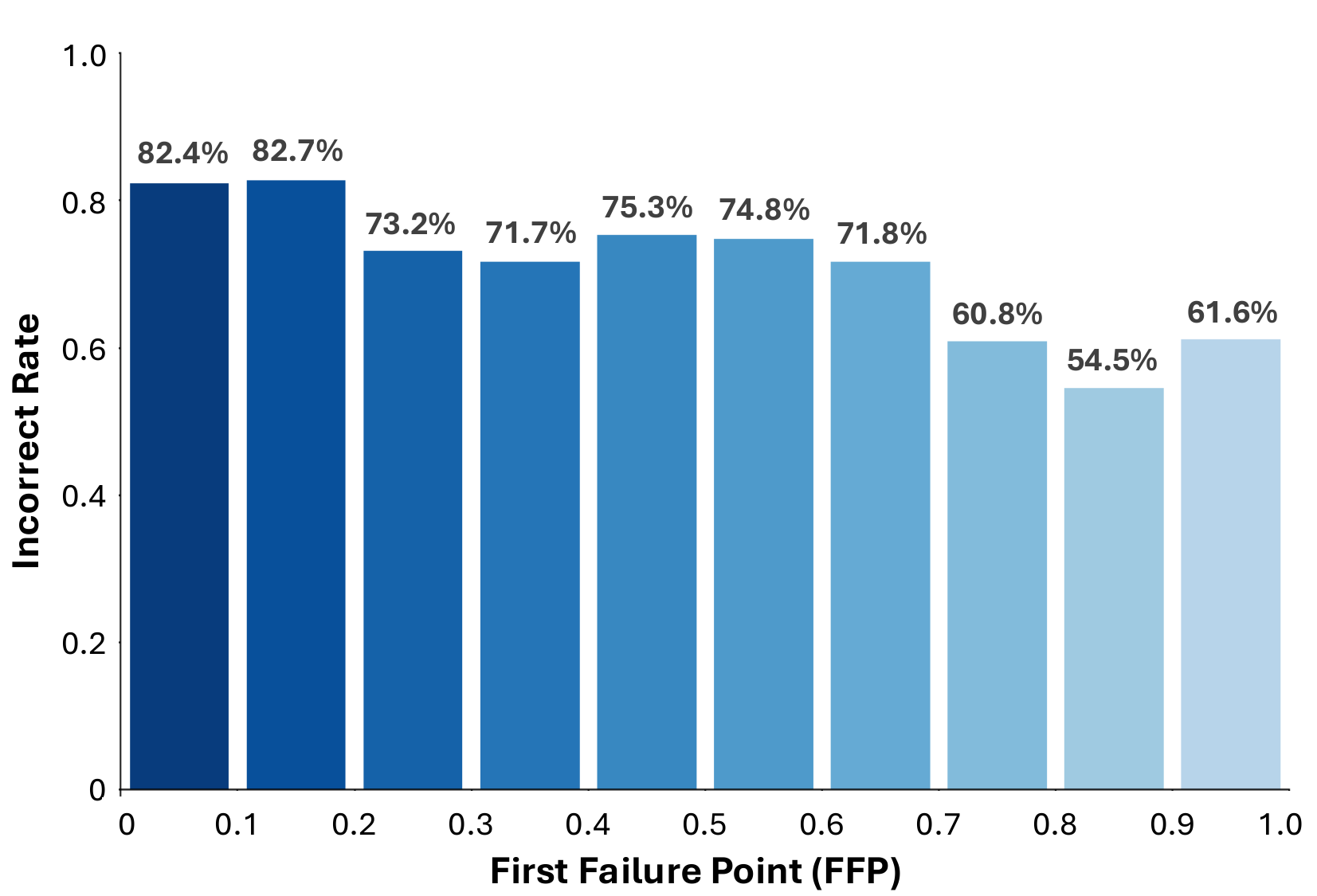}
    \caption{Incorrect rate across First Failure Point (FFP) bins.}
    \label{fig:preliminary-1}
  \end{subfigure}
  \hfill
  \begin{subfigure}[t]{0.48\textwidth}
    \centering
    \includegraphics[width=\linewidth]{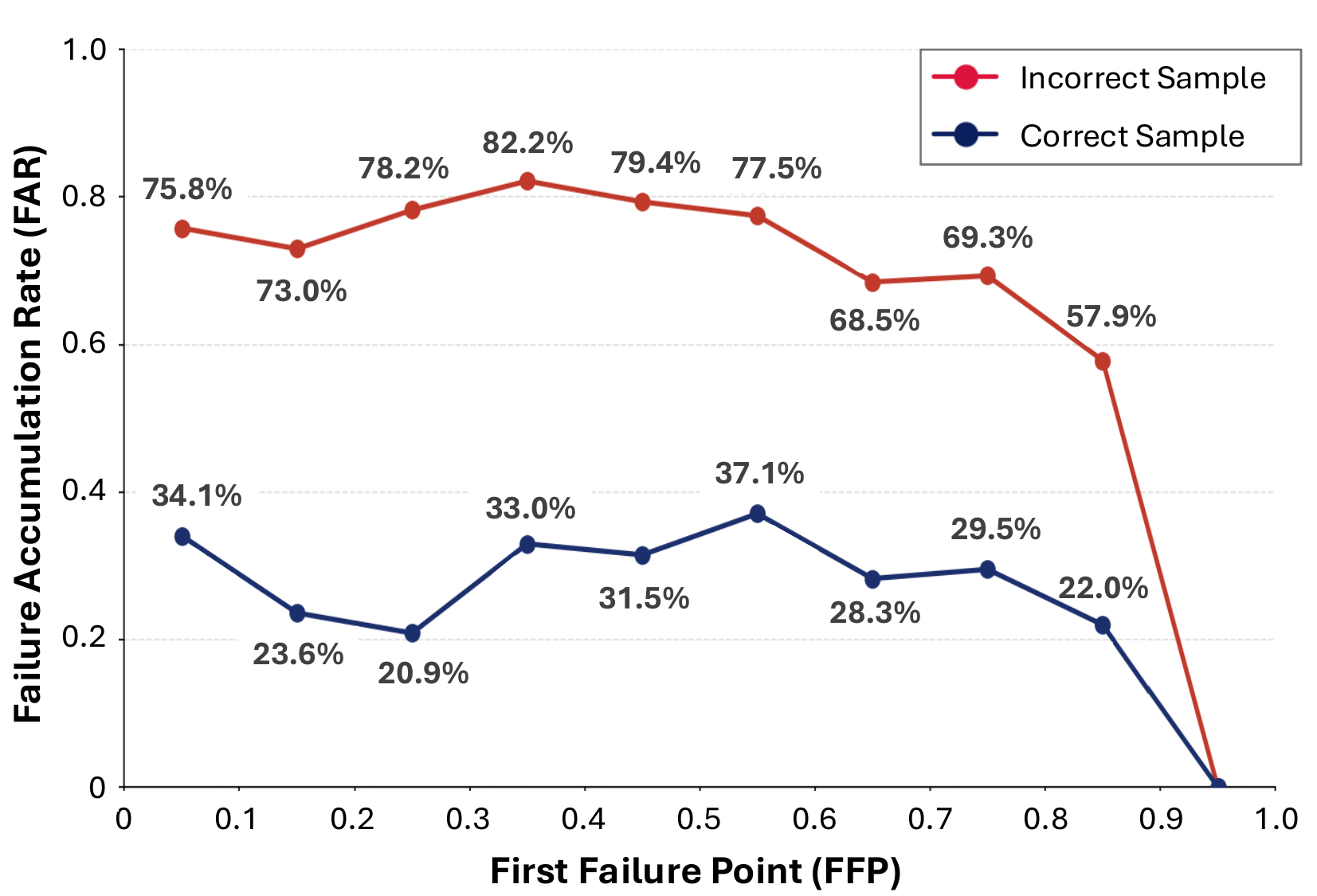}
    \caption{Failure Accumulation Rate (FAR) across FFP bins.}
    \label{fig:preliminary-2}
  \end{subfigure}

  \caption{
    \textbf{Step-wise medical multimodal reasoning analysis.}
    (A) Incorrect rate across FFP bins. Earlier first failures are associated with substantially higher incorrect rates.
    (B) FAR across FFP bins for correct and incorrect instances. Incorrect predictions show greater failure accumulation, particularly when the first failure occurs early.
  }
  \label{fig:ffp-analysis}
  \vspace{-0.3cm}
\end{figure*}

\section{Preliminary Experiments}
\label{sec:prelim}

To analyze structural failures in reasoning patterns for open-ended medical VQA, we conduct preliminary experiments across a diverse set of multimodal LLMs and medical VQA benchmarks.

\paragraph{Experiment Purpose. }
We first analyze the relationship between the point of first reasoning failure and the final answer incorrect rate. 
To this end, we define the \emph{First Failure Point} (FFP), which represents the relative position in a reasoning trace where the first invalid step occurs. Given $K$ total reasoning steps and the index $k$ of the first invalid step, we compute $\mathrm{FFP} = k/K$. A lower FFP indicates that reasoning fails earlier in the trace. We further examine how reasoning behaves after the first failure by defining the \emph{Failure Accumulation Rate} (FAR), which measures the proportion of failed steps among the remaining steps after the first failed step. We compute FAR as follows:

\begin{equation}
\mathrm{FAR}
=
\frac{
\textnormal{\# failed steps after the first failure}
}{
\textnormal{\# total steps after the first failure}
}
\end{equation}

\paragraph{Models and Benchmarks.}
We evaluate four MLLMs spanning general-purpose and medical-specialized domains: (i) General MLLMs: Qwen3-VL-8B-Instruct~\cite{Qwen3-VL}, InternVL3-8B-Instruct~\cite{zhu2025internvl3exploringadvancedtraining}; (ii) Medical MLLMs: HuatuoGPT-Vision-7B~\cite{chen2024huatuogptvisioninjectingmedicalvisual}, Lingshu-7B~\cite{lasateam2025lingshugeneralistfoundationmodel}. We use three medical VQA benchmarks, VQA-RAD~\cite{vqarad2018}, SLAKE~\cite{liu2021slakesemanticallylabeledknowledgeenhanceddataset}, and PathVQA~\cite{he2020pathvqa30000questionsmedical}. 
To obtain gold rationales for each medical VQA instance, we use MedThink~\cite{gai2024medthinkexplainingmedicalvisual}, which provides medical decision-making rationales for these three benchmarks. We align each MedThink gold reasoning one-to-one with its corresponding test instance. We then exclude binary and multiple-choice questions to focus our experiment on open-ended instances.

\paragraph{Evaluation Metrics.}
We assess the correctness of answers to open-ended VQA using an LLM-as-judge approach~\citep{zheng2023judging} with GPT-5-mini\footnote{We use \texttt{gpt-5-mini-2025-08-07}}, following the evaluation prompt design in PeFoMed~\cite{he2025pefomedparameterefficientfinetuning}. The prompt used for this evaluation is provided in Appendix~\ref{app:appendix_prompts_answer}.
% Evaluating the validity of each reasoning step in medical VQA is inherently challenging, as it requires identifying domain-specific key findings, anatomical landmarks, and pathological features relevant to the given question. Without explicit guidance on what constitutes valid observations or inferences, an LLM judge lacks sufficient context to assess whether a step contributes meaningfully to the diagnostic reasoning process.

% To address this challenge, we introduce two complementary metrics that together define step validity:

% Evaluating the validity of each reasoning step in medical VQA is inherently challenging, as it requires identifying domain-specific key findings, anatomical landmarks, and pathological features that an LLM judge alone cannot reliably assess. 
Evaluating the validity of each reasoning step in medical VQA is inherently challenging, as it requires identifying domain-specific key findings, anatomical landmarks, and pathological features relevant to the question. Without explicit guidance on what constitutes valid observations or inferences, an LLM judge lacks sufficient context to assess whether a step contributes meaningfully to the diagnostic process.
We thus introduce two complementary metrics that together define step validity. 
\textbf{(i) Gold Alignment} measures whether a reasoning step is consistent with the gold reasoning trajectory. By providing expert-annotated reasoning as a reference, we anchor the evaluation to clinically appropriate observations and diagnostic directions.
\textbf{(ii) Answer Contribution:} measures whether a reasoning step directly contributes to deriving the ground-truth answer. This metric captures cases where a step may diverge from the gold reasoning path yet still validly supports the correct conclusion, acknowledging that multiple reasoning trajectories can lead to the same conclusion.
Each metric is scored binarily at the sentence level using GPT-5-mini, and a step is \emph{valid} if it scores 1 on either metric, reflecting that multiple valid diagnostic pathways can lead to the same conclusion in medical practice. The prompt is in Appendix~\ref{app:appendix_process_reward_prompt}.

% All judgments are produced at the sentence level using an LLM-as-judge approach with GPT-5-mini, where each metric is scored binarily as 0 or 1 for every reasoning step. We consider a reasoning step \emph{valid} if it receives a score of 1 on either gold alignment or answer contribution, reflecting that multiple valid diagnostic pathways can lead to the same conclusion in medical practice. Conversely, a step is deemed \emph{invalid} only when it neither aligns with the reference reasoning nor contributes to the correct answer. The prompt used for this step-level evaluation is provided in Appendix~\ref{app:appendix_process_reward_prompt}.

To validate the reliability of our two LLM-as-judge protocols, we conduct human evaluation on answer correctness and step-wise reasoning quality. Both protocols achieve substantial agreement with human judgments, as detailed in Appendix~\ref{app:human_eval}.

\begin{figure*}[t]
  \centering
  \includegraphics[width=\textwidth]{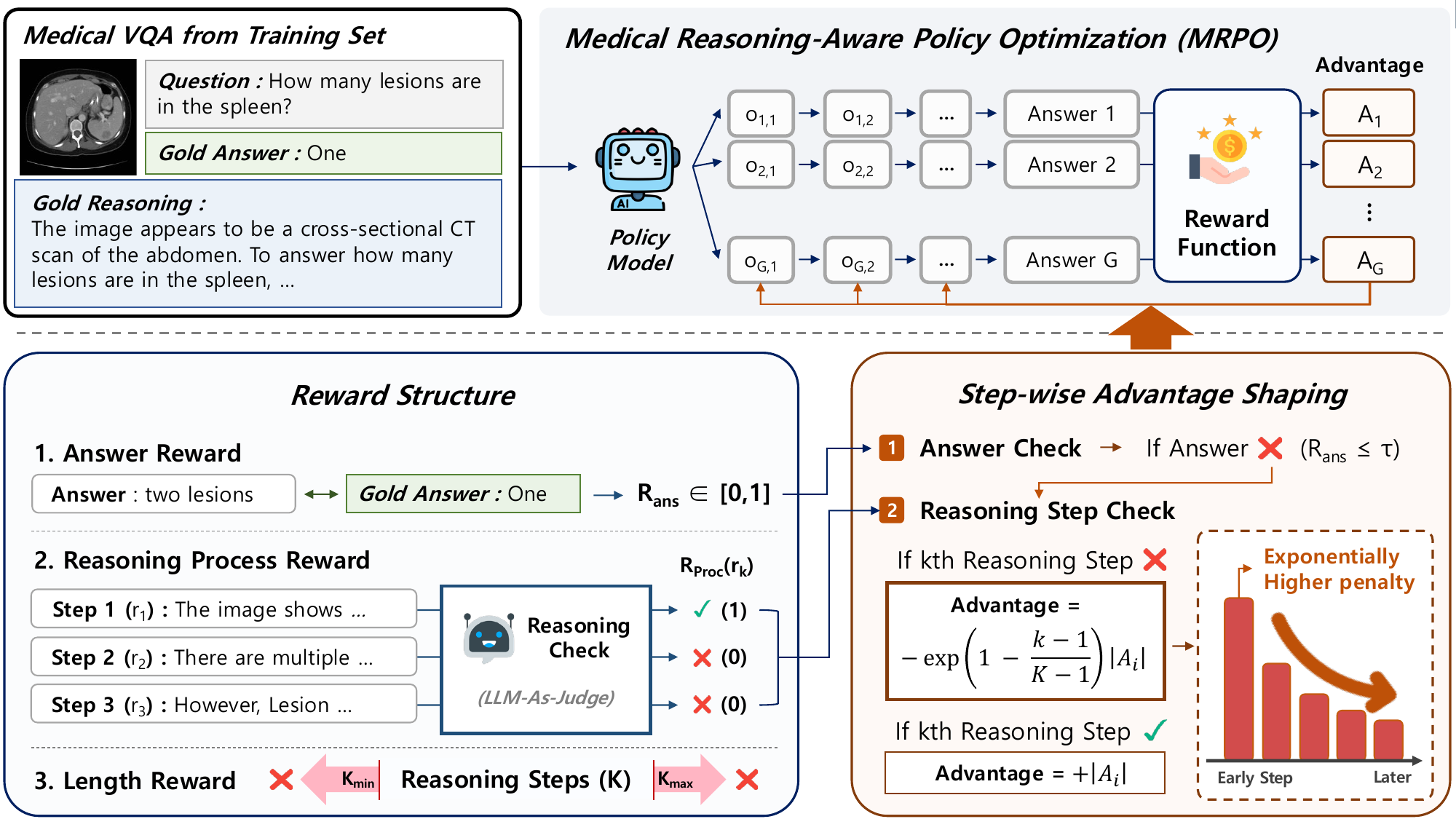}
  \caption{\textbf{Overview of the MRPO algorithm.} The policy model generates multiple reasoning paths, each evaluated by answer, step-wise reasoning process reward, and length reward. When the answer is judged incorrect, MRPO assigns larger penalties to earlier failed steps to correct early-stage reasoning failures.}
  \label{fig:main_figure}
  \vspace{-0.3cm}
\end{figure*}

\paragraph{Results and Analysis.}

% Figure~\ref{fig:preliminary-1} groups instances into bins of width 0.1 based on FFP and visualizes the incorrect rate within each bin. Aggregating results from all four models, we observe a clear trend in which earlier FFPs correspond to earlier failures in step-wise reasoning and are associated with a higher probability of incorrect answers.

Figure~\ref{fig:preliminary-1} groups instances into bins of width 0.1 based on FFP and visualizes the incorrect rate within each bin. Aggregating all four models, we observe a clear trend in which earlier FFPs correspond to earlier failures in step-wise reasoning and are associated with a higher probability of incorrect answers. 
Figure~\ref{fig:preliminary-2} visualizes FAR across FFP bins separately for correct and incorrect instances, revealing two observations. First, across all bins, incorrect instances exhibit higher FAR than correct ones, suggesting a strong link between failure accumulation and answer correctness. Second, for incorrect instances, earlier FFPs show even higher FAR, indicating stronger downstream failure accumulation when the first failure occurs early in the trajectory.

% In summary, our analysis demonstrates that initial reasoning failures trigger an error cascade that leads to incorrect medical VQA outcomes. Standard GRPO-based methods struggle to address this, as they distribute learning signals uniformly across all tokens, failing to pinpoint early errors. We thus propose a medical reasoning-aware RL algorithm that overcomes this limitation by evaluating reasoning step-wise. By assigning larger penalties to tokens in the initial stages of a failed trajectory, our method specifically targets and corrects the root causes of reasoning failure.

In summary, our analysis demonstrates that initial reasoning failures trigger an error cascade that leads to incorrect medical VQA outcomes. Standard GRPO-based methods struggle to address this, as they distribute learning signals uniformly across all tokens, failing to pinpoint early errors. We thus propose a medical reasoning-aware RL algorithm that overcomes this limitation by evaluating reasoning step-wise and assigning larger penalties to tokens in the earlier invalid stages of a failed trajectory, targeting the root causes of reasoning failure.

\section{Approach}
We propose \textbf{M}edical \textbf{R}easoning-aware \textbf{P}olicy \textbf{O}ptimization \textbf{(MRPO)}, which improves medical VQA reasoning by correcting failures in the early stages of the reasoning process.
As shown in Figure~\ref{fig:main_figure}, MRPO incorporates both an answer reward and a step-wise reasoning reward during policy optimization. Based on these signals, MRPO shapes advantages to assign larger penalties to tokens in reasoning steps that fail earlier when the final answer is judged incorrect. This discourages early-stage reasoning failures and encourages valid reasoning in subsequent steps, thereby improving both reasoning quality and answer accuracy.

% In this section, we describe our approach for effectively improving medical VQA reasoning by correcting failures that arise in the early stages of the reasoning process. To this end, we propose \textbf{M}edical \textbf{R}easoning-aware \textbf{P}olicy \textbf{O}ptimization \textbf{(MRPO)}. 
% As shown in Figure~\ref{fig:main_figure}, MRPO incorporates both an answer reward and a step-wise reasoning reward during policy optimization. Based on these signals, MRPO shapes advantages to assign larger penalties to tokens in reasoning steps that fail earlier, when the final answer is judged incorrect. This approach discourages early-stage reasoning failures and encourages valid reasoning in subsequent steps, thereby improving both reasoning quality and answer accuracy.

\subsection{Reward structure}
\label{sec:reward_structure}
We define three reward components, (1) Answer Reward, (2) Reasoning Process Reward, and (3) Length Reward. The final reward is computed as a weighted combination of these components.

\paragraph{Answer Reward. }
To assess how well the generated answer aligns with the reference answer~\citep{jeong2024olaph}, we adopt a weighted combination of lexical overlap ($\lambda_1 = 0.25$) and semantic similarity ($\lambda_2 = 0.5$):
\begin{equation}
\begin{aligned}
R_{\mathrm{ans}} =\;&
\lambda_1 \cdot (\mathrm{ROUGE\text{-}1}
+ \mathrm{BLEU\text{-}1}) \\
&+ \lambda_2 \cdot \mathrm{BERTScore}
\end{aligned}
\end{equation}
ROUGE-1~\cite{lin-2004-rouge} and BLEU-1~\cite{papineni-etal-2002-bleu} capture lexical overlap, but can be sparse for short open-ended medical answers.
We therefore include BERTScore~\citep{zhang2019bertscore} to provide a denser semantic signal, computed using BiomedBERT~\cite{chakraborty-etal-2020-biomedbert}.

\paragraph{Reasoning Process Reward.}
We segment the reasoning text $r$ into $K$ sentence-level steps $\{r_k\}_{k=1}^K$, and evaluate each step $r_k$ given the ground-truth answer $a^\star$ and gold reasoning $r^\star$ using two criteria.
\textbf{(i) Gold Alignment} assesses whether step $r_k$ is consistent with $r^\star$. A step is aligned if it identifies key findings in $r^\star$, maintains correct anatomical localization, and follows the diagnostic pathway. A step is misaligned if it contradicts $r^\star$, identifies incorrect anatomical location or laterality, or gives only generic instead of specific findings.
\textbf{(ii) Answer Contribution} assesses whether step $r_k$ directly contributes to deriving $a^\star$. A step is contributive if it explicitly mentions $a^\star$ or identifies findings required to derive it. This captures cases where a step may diverge from the gold reasoning path yet still supports the correct conclusion, acknowledging that multiple reasoning trajectories can lead to the same diagnosis.

These criteria match the reasoning evaluation metrics in Section~\ref{sec:prelim}. Using GPT-5-mini as an LLM judge, we assign a binary score $\mathrm{score}(r_k, c)\in\{0,1\}$ for each criterion $c$, and set $R_{\mathrm{proc}}(r_k)=1$ if either criterion scores 1, and $R_{\mathrm{proc}}(r_k)=0$ otherwise. The prompt is provided in Appendix~\ref{app:appendix_process_reward_prompt}.

\begin{equation}
\resizebox{0.88 \linewidth}{!}{$
R_{\mathrm{proc}}(r_k) =
\begin{cases}
1, & \text{if } \mathrm{score}(r_k, c)=1, \ \exists c \in \mathcal{C} \\
0, & \text{otherwise}
\end{cases}
$}
\end{equation}

\paragraph{Length Reward.}
Reasoning traces with insufficient length may omit essential diagnostic steps, while excessively lengthy outputs introduce redundant information. To prevent such behaviors, we introduce a length-based reward $R_{\mathrm{len}}$ that regularizes the number of reasoning steps. Given a trace of $K$ sentence-level steps, we define:
\begin{equation}
R_{\mathrm{len}}(r) =
\begin{cases}
-\frac{K_{\min} - K}{K_{\min}}, & \text{if } K < K_{\min} \\[6pt]
-\frac{K - K_{\max}}{K_{\max}}, & \text{if } K > K_{\max} \\[6pt]
0, & \text{otherwise}
\end{cases}
\end{equation}
% where $K_{\min}$ and $K_{\max}$ denote the minimum and maximum acceptable step counts. In our experiments, we set $K_{\min}$ to 4 and $K_{\max}$ to 10. This reward imposes a linear penalty proportional to the deviation from the acceptable range, encouraging the model to generate reasoning traces of appropriate length.
where $K_{\min}$ and $K_{\max}$ denote the minimum and maximum acceptable step counts, set to 4 and 10 in our experiments. This reward imposes a linear penalty proportional to the deviation from the acceptable range, encouraging the model to generate reasoning traces of appropriate length.

% The reasoning process reward $R_{\mathrm{proc}}(r_k)$ is averaged over all reasoning steps to obtain the final reasoning process reward. Together with the answer reward and the length reward, this constitutes the total reward for training.

The total reward combines the answer reward, the step-averaged reasoning process reward, and the length reward as follows:

\begin{equation}
R_{\mathrm{tot}} = R_{\mathrm{ans}} + \frac{1}{K} \sum_{k=1}^{K} R_{\mathrm{proc}}(r_k) + R_{\mathrm{len}}
\end{equation}

\subsection{Medical Reasoning-aware Advantage Shaping}
\label{sec:MRPO}
We adopt GRPO~\cite{shao2024deepseekmathpushinglimitsmathematical} as our base policy optimization algorithm. Given an input $x$, GRPO generates a set of $G$ candidate outputs $\{y_1,\ldots,y_G\}$, with rewards $\{R_1,\ldots,R_G\}$, and computes a group-normalized advantage for each candidate output $y_i$ as
\begin{equation}
A_i = \frac{R_i - \mathrm{mean}(\{R_1, \ldots, R_G\})}
           {\mathrm{std}(\{R_1, \ldots, R_G\})}
\end{equation}
Most GRPO-based methods compute the advantage at the sequence level and apply the same learning signal uniformly to all tokens, which fails to provide differentiated supervision across reasoning steps.
Consequently, it is inadequate for mitigating the early-stage reasoning failures and ensuing error accumulation observed in Section~\ref{sec:prelim}. To address this, MRPO reshapes advantages step-wise to assign larger penalties to earlier failed reasoning steps when the final answer is judged incorrect.

\begin{table*}[t]
\centering
\footnotesize
\setlength{\tabcolsep}{4pt}
\renewcommand{\arraystretch}{1.0}
\begin{tabular}{p{5.5cm}ccccccc}
\toprule
\multirow{2}{*}{\textbf{Model}} & \multicolumn{5}{c}{\textbf{Benchmarks}} & \multirow{2}{*}{\textbf{AVG}} \\
\cmidrule(lr){2-6}
& \textbf{PMC-VQA} & \textbf{VQA-Med} & \textbf{Quilt-VQA} & \textbf{Rad-VQA} & \textbf{MIMIC-VQA} & \\
\midrule
\multicolumn{7}{l}{\textit{General MLLMs}} \\
Qwen2.5-VL-7B-Instruct  & 28.35 & 5.18 & 19.06 & 27.20 & 15.08 & 22.28 \\
Qwen3-VL-8B-Instruct    & 31.00 & 9.41 & 23.90 & 33.15 & 15.31 & 25.61 \\
Qwen3-VL-8B-Thinking    & 30.75 & \underline{10.82} & 22.92 & 31.20 & \underline{22.39} & 26.73 \\
InternVL3-8B-Instruct   & 30.00 & 7.59 & 22.24 & 37.60 & 15.20 & 26.29 \\
LLaVA-v1.6-7B           & 11.45 & 3.29 & 9.94 & 21.60 & 7.19 & 12.67 \\
LLaVA-v1.6-34B          & 15.90 & 4.71 & 13.12 & 24.90 & 9.80 & 16.00 \\
\midrule
\multicolumn{7}{l}{\textit{Medical MLLMs}} \\
LLaVA-Med-v1.5-7B       & 20.65 & 3.53 & 14.92 & 23.60 & 9.57 & 17.07 \\
HuatuoGPT-Vision-7B     & 27.90 & 8.94 & 19.61 & 32.70 & 16.06 & 24.28 \\
HuatuoGPT-Vision-34B    & 31.30 & 9.65 & 21.27 & 33.60 & 17.63 & 26.15 \\
Chiron-o1-8B            & 29.30 & 6.82 & 20.86 & 29.85 & 16.82 & 24.05 \\
QoQ-Med-7B              & 28.90 & 7.76 & 21.82 & 26.60 & 14.56 & 22.58 \\
MedGemma-27B            & 31.25 & 8.00 & 20.72 & 31.90 & 18.10 & 25.59 \\
\midrule
\multicolumn{7}{l}{\textit{Medical Reasoning MLLMs (GRPO)}} \\
MedVLM-R1-2B            & 23.35 & 5.41 & 17.54 & 25.10 & 12.88 & 19.51 \\
MedVLThinker-7B         & 28.90 & 6.82 & 20.30 & 28.85 & 15.66 & 23.29 \\
\midrule
\multicolumn{7}{l}{\textit{Our Method(MRPO)}} \\
\textbf{Qwen2.5-VL-7B-Instruct(MRPO)}       & 30.85 & 5.65 & 22.79 & 29.65 & 18.65 & 25.04 \\
\textbf{Qwen3-VL-8B-Instruct(MRPO)}       & \underline{33.00} & \textbf{12.00} & 23.90 & \underline{40.20} & 17.46 & \underline{28.94} \\
\textbf{Qwen3-VL-8B-Thinking(MRPO)}       & \textbf{33.80} & 7.76 & \textbf{25.00} & \textbf{40.55} & \textbf{23.90} & \textbf{30.74} \\
\textbf{InternVL3-8B-Instruct(MRPO)}       & 31.60 & 7.59 & \underline{24.31} & 39.15 & 20.42 & 28.75 \\
\bottomrule
\end{tabular}
\caption{\textbf{Performance comparison of MRPO against existing MLLMs.} All models are evaluated on five out-of-distribution benchmarks, where VQA-Med denotes VQA-Med-2021, Rad-VQA denotes RadImageNet-VQA, and MIMIC-VQA denotes MIMIC-Ext-MIMIC-CXR-VQA. The best result in each column is in \textbf{bold} and the second-best is \underline{underlined}. AVG denotes the average across all five benchmarks.}
\vspace{-0.3cm}
\label{main-result}
\end{table*}

\paragraph{Step-wise Advantage Shaping.}

We adjust token-level advantages using the answer reward $R_{\mathrm{ans}}$ and the step-wise reasoning process reward $R_{\text{proc}}(r_k)$ for each step $r_k$.
% We judge the final answer as correct if $R_{\mathrm{ans}} > \tau$ and as incorrect otherwise. To set $\tau$, we use the samples labeled as correct or incorrect by the LLM-as-judge in Section~\ref{sec:prelim}, and compute the mean $R_{\mathrm{ans}}$ for each group. We set $\tau$ to the midpoint between these two means, yielding $\tau = 0.6$.
We judge the final answer as correct if $R_{\mathrm{ans}} > \tau$ and as incorrect otherwise, where $\tau$ is set to the midpoint between the mean $R_{\mathrm{ans}}$ of samples labeled correct and incorrect by the LLM-as-judge in Section~\ref{sec:prelim}, yielding $\tau = 0.6$.

When the final answer is incorrect and a reasoning step is evaluated as invalid ($R_{\mathrm{proc}}(r_k)=0$), we reshape the advantage to assign exponentially larger penalties to tokens in earlier failed steps. Concretely, for an output $y_i$, we modify the advantage of each token $o_{i,t}\in r_k$ as follows:

\begin{equation}
\resizebox{0.88 \linewidth}{!}{$
\hat{A}_{i,t} =
\begin{cases}
\scalebox{0.8}{$-\exp\!\left(1-\dfrac{k-1}{K-1}\right)|A_i|$}
& \text{if } \substack{R_{\mathrm{ans}} \le \tau \\ \text{and } R_{\mathrm{proc}}(r_k)=0} \\[12pt]
\scalebox{0.8}{$+|A_i|$}
& \text{if } \substack{R_{\mathrm{ans}} \le \tau \\ \text{and } R_{\mathrm{proc}}(r_k)=1} \\[12pt]
\scalebox{0.8}{$A_i$}
& \text{otherwise}
\end{cases}
$}
\end{equation}

% By reshaping the advantage as $\hat{A}_{i,t}$, we more strongly decrease the probability of inaccurate tokens occurring in earlier failed reasoning steps. This substantially increases the probability of generating correct tokens at the first failure point, enabling correction of the failed step and facilitating valid reasoning in subsequent steps. In contrast, when the final answer is judged correct, we do not reweight the reasoning tokens. This choice reflects the intuition that minor imperfections in the generated reasoning did not materially hinder arriving at the correct answer. With this design, MRPO selectively corrects reasoning only for incorrect predictions, thereby improving failed reasoning traces without disrupting successful reasoning trajectories. Based on the adjusted advantages, we optimize the policy using the following GRPO-based training objective for each input prompt $x$:

This reshaping more strongly decreases the probability of inaccurate tokens in earlier failed reasoning steps, substantially increasing the probability of generating correct tokens at the first failure point and facilitating valid reasoning in subsequent steps. In contrast, when the final answer is judged correct, we do not reweight the reasoning tokens, reflecting the intuition that minor imperfections did not materially hinder arriving at the correct answer. With this design, MRPO selectively corrects reasoning only for incorrect predictions, improving failed reasoning traces without disrupting successful trajectories. Based on the adjusted advantages, we optimize the policy using the following GRPO-based training objective for each input prompt $x$:

\begin{equation}
\resizebox{0.88\linewidth}{!}{$
\begin{aligned}
\mathcal{J}_{\mathrm{MRPO}}(\theta)
= \mathbb{E}_{\{y_i\}_{i=1}^{G}\sim \pi_{\theta_{\mathrm{old}}}(x)}
\frac{1}{G}\sum_{i=1}^{G}\frac{1}{|y_i|}
\sum_{t=1}^{|y_i|} \\
\Big[\min\Big(
\rho_{i,t}(\theta)\,\hat{A}_{i,t},\;
\mathrm{clip}\!\left(\rho_{i,t}(\theta),\,1-\epsilon,\,1+\epsilon\right)\hat{A}_{i,t}
\Big)\\
\;-\; \beta\, D_{\mathrm{KL}}\!\left(\pi_\theta \,\|\, \pi_{\mathrm{ref}}\right)
\Big]
\end{aligned}
$}
\end{equation}

\begin{equation}
\rho_{i,t}(\theta)
= \frac{\pi_\theta\!\left(o_{i,t}\mid x, y_{i,<t}\right)}
{\pi_{\theta_{\mathrm{old}}}\!\left(o_{i,t}\mid x, y_{i,<t}\right)}
\end{equation}

\begin{table*}[t]
\centering

\renewcommand{\arraystretch}{1.08}

\resizebox{\textwidth}{!}{%
\begin{tabular}{lccccccccc}
\toprule
\multirow{2}{*}{\textbf{Method}} &
\multicolumn{3}{c}{\textbf{In-Distribution}} &
\multicolumn{5}{c}{\textbf{Out-of-Distribution}} &
\multirow{2}{*}{\textbf{AVG}} \\
\cmidrule(lr){2-4}\cmidrule(lr){5-9}
& \textbf{VQA-RAD}
& \textbf{SLAKE}
& \textbf{PathVQA}
& \textbf{PMC-VQA}
& \textbf{VQA-Med}
& \textbf{Quilt-VQA}
& \textbf{Rad-VQA}
& \textbf{MIMIC-VQA}
& \\

\midrule

\multicolumn{10}{@{}l}{\textit{Qwen2.5-VL-7B-Instruct}} \\
Base Model
& 42.00 & 58.29 & 17.10
& 28.35 & 5.18 & 19.06
& 27.20 & 15.08 & 23.36 \\
SFT
& 38.50 & 63.88 & 18.83
& 28.50 & 4.71 & 18.37 & 26.20 & 15.02 & 23.59 \\
GRPO
& \textbf{44.00} & 65.27 & \underline{20.52}
& \textbf{31.05} & 3.76 & \underline{20.99}
& 29.30 & \underline{16.71} & \underline{26.06} \\
GDPO
& \underline{42.50} & \textbf{66.29} & 19.60
& 28.60 & \textbf{7.76} & 20.17
& \textbf{32.00} & 16.47 & 25.92 \\
\rowcolor{cyan!12}
MRPO
& 41.50 & \underline{65.89} & \textbf{21.30}
& \underline{30.85} & \underline{5.65} & \textbf{22.79}
& \underline{29.65} & \textbf{18.65} & \textbf{26.79} \\

\midrule
\multicolumn{10}{@{}l}{\textit{Qwen3-VL-8B-Instruct}} \\
Base Model
& \underline{43.00} & 59.69 & \textbf{21.48}
& 31.00 & 9.41 & \underline{23.90}
& 33.15 & 15.31 & 26.83 \\
SFT
& \textbf{45.00} & 66.29 & \underline{21.27}
& 30.35 & 7.76 & 19.48 & 33.40 & 15.20 & 26.79 \\
GRPO
& \textbf{45.00} & 67.14 & 20.11
& 30.25 & 9.65 & \textbf{24.03}
& \textbf{40.60} & \underline{18.79} & 28.69 \\
GDPO
& 42.50 & \underline{67.28} & 19.09
& \underline{32.10} & \textbf{12.47} & 22.51
& 40.10 & \textbf{21.46} & \underline{29.01} \\
\rowcolor{cyan!12}
MRPO
& 41.50 & \textbf{68.27} & 20.43
& \textbf{33.00} & \underline{12.00} & \underline{23.90}
& \underline{40.20} & 17.46 & \textbf{29.09} \\

\midrule
\multicolumn{10}{@{}l}{\textit{Qwen3-VL-8B-Thinking}} \\
Base Model
& 38.50 & 62.60 & \textbf{21.51}
& 30.75 & \textbf{10.82} & 22.92
& 31.20 & 22.39 & 27.64 \\
SFT
& \textbf{44.50} & 66.29 & 19.78
& 30.30 & 7.06 & 21.41 & 33.20 & 20.42 & 27.19 \\
GRPO
& 41.50 & \textbf{67.14} & \underline{21.21}
& \underline{33.40} & \underline{9.65} & \underline{24.31}
& \underline{35.50} & \underline{25.99} & \underline{29.74} \\
GDPO
& \textbf{44.50} & 66.29 & 20.79
& 32.85 & 7.06 & 23.62
& 34.40 & \textbf{26.33} & 29.23 \\
\rowcolor{cyan!12}
MRPO
& \underline{43.00} & \underline{67.00} & \textbf{21.51}
& \textbf{33.80} & 7.76 & \textbf{25.00}
& \textbf{40.55} & 23.90 & \textbf{30.48} \\

\midrule
\multicolumn{10}{@{}l}{\textit{InternVL3-8B-Instruct}} \\
Base Model
& \textbf{45.50} & 64.73 & 22.97
& 30.00 & 7.59 & 22.24
& 37.60 & 15.20 & 28.07 \\
SFT
& \underline{44.00} & 68.83 & 26.57
& 29.50 & 4.71 & 17.12 & 35.20 & 17.11 & 27.94 \\
GRPO
& 42.00 & 68.27 & \underline{28.15}
& 30.50 & \underline{7.76} & \underline{23.48}
& \underline{38.75} & \underline{19.49} & \underline{30.84} \\
GDPO
& 39.50 & \textbf{71.95} & 26.18
& \underline{30.90} & \textbf{10.12} & 23.20
& 37.85 & 17.46 & 30.11 \\
\rowcolor{cyan!12}
MRPO
& 41.00 & \underline{71.78} & \textbf{29.54}
& \textbf{31.60} & 7.59 & \textbf{24.31}
& \textbf{39.15} & \textbf{20.42} & \textbf{31.94} \\

\bottomrule
\end{tabular}%
}

\vspace{0.1cm}
\caption{
\textbf{Cross-backbone ablation of training methods.}
We compare MRPO against the base model, SFT, GRPO, and GDPO on four backbones, namely Qwen2.5-VL-7B-Instruct, Qwen3-VL-8B-Instruct, Qwen3-VL-8B-Thinking, and InternVL3-8B-Instruct.
All methods are evaluated on three in-distribution and five out-of-distribution benchmarks.
The best result in each column is in \textbf{bold} and the second-best is \underline{underlined}.
AVG denotes the average across all benchmarks.
}
\label{tab:main-result-ablation-all}
\vspace{-0.3cm}
\end{table*}

\section{Experiments}

\subsection{Experimental Setup}
\label{sec:experimental_setup}

\paragraph{Dataset.}
Our training set is derived from the training splits of three medical VQA benchmarks, VQA-RAD~\cite{vqarad2018}, SLAKE~\cite{liu2021slakesemanticallylabeledknowledgeenhanceddataset}, and PathVQA~\cite{he2020pathvqa30000questionsmedical}, filtered to open-ended instances, and augmented with MedThink~\cite{gai2024medthinkexplainingmedicalvisual}, which provides gold reasoning annotations.
For evaluation, test splits of these three benchmarks serve as our in-distribution test sets. We further adopt five out-of-distribution benchmarks spanning diverse imaging modalities unseen during training, including PMC-VQA~\cite{zhang2024pmcvqavisualinstructiontuning}, VQA-Med-2021~\cite{ImageCLEF-VQA-Med2021}, Quilt-VQA~\cite{seyfioglu2025quiltllavavisualinstructiontuning}, RadImageNet-VQA~\cite{butsanets2026radimagenetvqalargescalectmri}, and MIMIC-Ext-MIMIC-CXR-VQA~\cite{PhysioNet-mimic-ext-mimic-cxr-vqa-1.0.0}. Details are provided in Appendix~\ref{app:dataset}.

\paragraph{Backbones.}
We apply MRPO to four open-source multimodal LLM backbones of comparable scale, namely Qwen2.5-VL-7B-Instruct~\cite{Qwen2.5-VL}, Qwen3-VL-8B-Instruct~\cite{Qwen3-VL}, Qwen3-VL-8B-Thinking~\cite{Qwen3-VL}, and InternVL3-8B-Instruct~\cite{zhu2025internvl3exploringadvancedtraining}.

\paragraph{Evaluation Metrics.}
To evaluate \textbf{answer accuracy}, we adopt an LLM-as-judge approach using GPT-5-mini to assess answer correctness with a binary label. Accuracy is reported as the proportion of examples judged correct. The evaluation prompt is in Appendix~\ref{app:appendix_prompts_answer}, and its reliability is validated through a human alignment study in Appendix~\ref{app:human_eval}.

\paragraph{Baselines.}
We compare against three categories of baselines. \textbf{(i) General MLLMs,} general-purpose open-source MLLMs of varying sizes, such as Qwen2.5-VL~\cite{Qwen2.5-VL}, Qwen3-VL~\cite{Qwen3-VL}, InternVL3~\cite{zhu2025internvl3exploringadvancedtraining}, and LLaVA-v1.6~\cite{liu2023visualinstructiontuning}; \textbf{(ii) Medical MLLMs,} medical-domain MLLMs trained on specialized biomedical datasets, including LLaVA-Med~\cite{li2023llavamedtraininglargelanguageandvision}, HuatuoGPT-Vision~\cite{chen2024huatuogptvisioninjectingmedicalvisual}, Chiron-o1~\cite{sun2025chirono1ignitingmultimodallarge}, QoQ-Med~\cite{dai2025qoqmedbuildingmultimodalclinical}, and MedGemma~\cite{sellergren2025medgemmatechnicalreport}; \textbf{(iii) Medical Reasoning MLLMs,} medical reasoning MLLMs trained with GRPO, including MedVLM-R1~\cite{pan2025medvlmr1incentivizingmedicalreasoning} and MedVLThinker~\cite{huang2025medvlthinkersimplebaselinesmultimodal}.

\subsection{Main Results}
\label{sec:main_results}

\paragraph{Comparison against MLLMs.}
% Table~\ref{main-result} compares MRPO against existing MLLMs on diverse medical VQA benchmarks. MRPO consistently improves over the baseline across all three backbones, with the most pronounced gain on Qwen3-VL-8B-Instruct, where the average rises from 25.61 to 28.94, an improvement of 3.33 points. Built on this backbone, MRPO attains the highest average among all evaluated models.
% Compared to general-purpose MLLMs, our best-performing variant on Qwen3-VL-8B-Instruct outperforms all baselines regardless of scale, including the strongest reasoning variant Qwen3-VL-8B-Thinking at 26.73. Against medical MLLMs, it outperforms the best baseline HuatuoGPT-Vision-34B at 26.15 by 2.79 points despite a substantially smaller 8B backbone and only 13K training samples. This suggests that targeted reasoning supervision can be a competitive alternative to large-scale medical instruction tuning for inducing transferable clinical reasoning. Consistent with this, it obtains the highest score on three of the five benchmarks, indicating reasonable out-of-distribution generalization rather than overfitting to a narrow subset of clinical contexts.

Table~\ref{main-result} compares MRPO against existing MLLMs on diverse medical VQA benchmarks. MRPO consistently improves over the baseline across all four backbones, most notably on Qwen3-VL-8B-Thinking, where the average rises from 26.73 to 30.74, an improvement of 4.01 points. Built on this backbone, MRPO attains the highest average among all evaluated models. It obtains the top score on four of the five benchmarks, with a notable 9.35 point gain over the baseline on RadImageNet-VQA. Compared to general-purpose MLLMs, it outperforms all baselines regardless of scale. Against medical MLLMs, it outperforms the largest baseline HuatuoGPT-Vision-34B (26.15) by 4.59 points, and also surpasses the recent MedGemma-27B (25.59) by 5.15 points, despite its substantially smaller 8B backbone and only 13K training samples. Together, these results show that targeted reasoning supervision can be a competitive alternative to large-scale medical instruction tuning, inducing transferable reasoning that generalizes across diverse out-of-distribution benchmarks rather than overfitting to narrow clinical contexts.

\paragraph{Cross-backbone Ablation.}
To verify that the gains of MRPO are not specific to a particular backbone, we compare MRPO against the base model, SFT, GRPO, and GDPO~\cite{liu2026gdpogrouprewarddecouplednormalization} across four backbones in Table~\ref{tab:main-result-ablation-all}. GDPO is a recent variant of GRPO that decouples reward normalization across groups.
Training details are provided in Appendix~\ref{app:implementation_detail}.
MRPO achieves the highest average score across all eight benchmarks on all four backbones, reaching 26.79 on Qwen2.5-VL-7B-Instruct, 29.09 on Qwen3-VL-8B-Instruct, 30.48 on Qwen3-VL-8B-Thinking, and 31.94 on InternVL3-8B-Instruct. While SFT improves in-distribution performance, it fails to transfer to out-of-distribution benchmarks and even degrades several of them. In contrast, RL-based methods including MRPO yield consistent improvements on both in- and out-of-distribution benchmarks, indicating that RL induces transferable reasoning capability rather than overfitting to the training distribution. Among the RL methods, MRPO consistently outperforms both GRPO and GDPO, with average improvements of 0.73, 0.40, 0.74, and 1.10 points over GRPO on the four backbones. The improvement over GRPO is largest on InternVL3-8B-Instruct, where MRPO achieves the best score on five of the eight benchmarks. This suggests that step-wise advantage reshaping provides finer-grained credit assignment than the sequence-level optimization of GRPO and GDPO. Additional ablations are provided in Appendix~\ref{app:ablation_study}.

\begin{figure}[t]
  \centering
  \includegraphics[width=\linewidth]{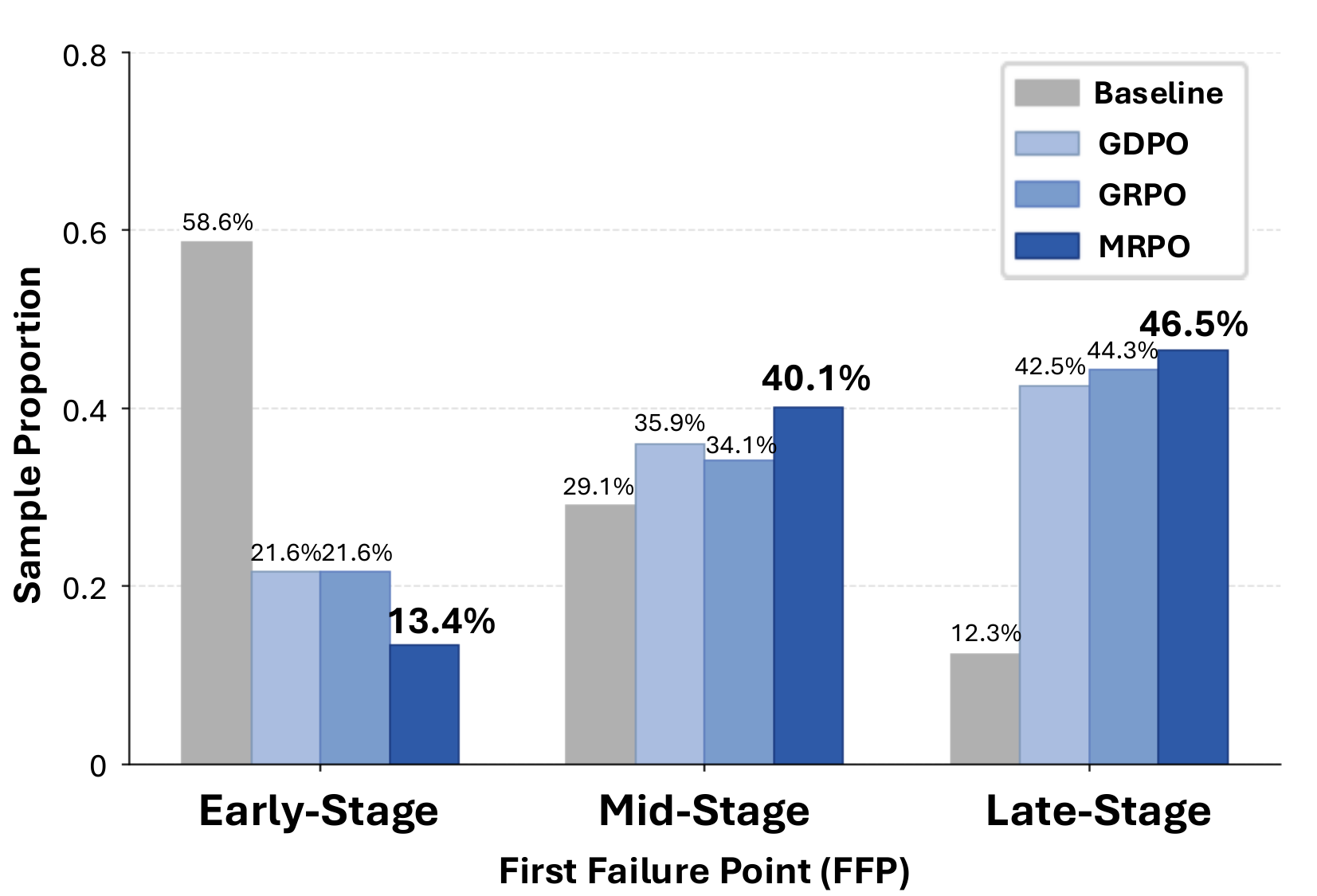}
  \caption{
    \textbf{Sample distribution across First Failure Point (FFP) stages.}
    Grouped into Early (0.0--0.4), Mid (0.4--0.7), and Late-Stage (0.7--1.0).
  }
  \label{fig:Reasoning_ablation_1}
  \vspace{-0.3cm}
\end{figure}

\subsection{Reasoning Analysis}
\label{sec:reasoning_quality_analysis}

\paragraph{First Failure Point Analysis.}
To verify that MRPO addresses the cascading failure problem from Section~\ref{sec:prelim}, we analyze how the distribution of reasoning failures changes under MRPO. We include GRPO and GDPO under identical conditions to isolate the effect of step-wise advantage reshaping. Following the evaluation protocol in Section~\ref{sec:prelim}, we assess step validity and compare the proportion of samples in each of three FFP stages: Early (0.0--0.4), Mid (0.4--0.7), and Late-Stage (0.7--1.0)

Figure~\ref{fig:Reasoning_ablation_1} shows the FFP distribution averaged across the four backbones. The baseline concentrates 58.6\% of failures in the early-stage range, indicating frequent failures at the beginning of the trajectory. All RL methods reduce this, but MRPO lowers early-stage failures the most, to 13.4\% versus 21.6\% for GRPO and GDPO. Correspondingly, MRPO shifts the largest share of failures to the late-stage at 46.5\%, compared to 12.3\% in the baseline. These results demonstrate that MRPO's exponential penalty on early invalid steps effectively prevents early-stage failures and shifts the failure distribution toward later stages.

\paragraph{Failure Accumulation Analysis.}

% Figure~\ref{fig:Reasoning_ablation_2} compares the FAR across FFP bins at 0.2 intervals, averaged over the three backbones, where a lower FAR at a given FFP indicates that the model recovers more effectively after the first failure. In the Early-to-Mid-Stage range from FFP 0.0 to 0.6, the baseline model exhibits consistently high FAR values, indicating that an early failure tends to propagate through most of the subsequent steps. In contrast, MRPO records the lowest FAR in this range across all methods, with a particularly pronounced reduction in the 0.0--0.2 bin where its FAR of 43.3\% is markedly lower than the baseline's 64.6\% and also clearly below those of GRPO and GDPO. This pattern indicates that MRPO not only delays the onset of reasoning failures but also substantially reduces their downstream propagation, with the strongest effect precisely when failures begin early. Together with the FFP results in Figure~\ref{fig:Reasoning_ablation_1}, these findings confirm that MRPO's step-wise advantage reshaping addresses cascading failures along two complementary axes, delaying the onset of failures and reducing their accumulation thereafter. Detailed results for each backbone and qualitative analyses are provided in Appendix~\ref{app:reasoning_analysis}.

\begin{figure}[t]
  \centering
  \includegraphics[width=\linewidth]{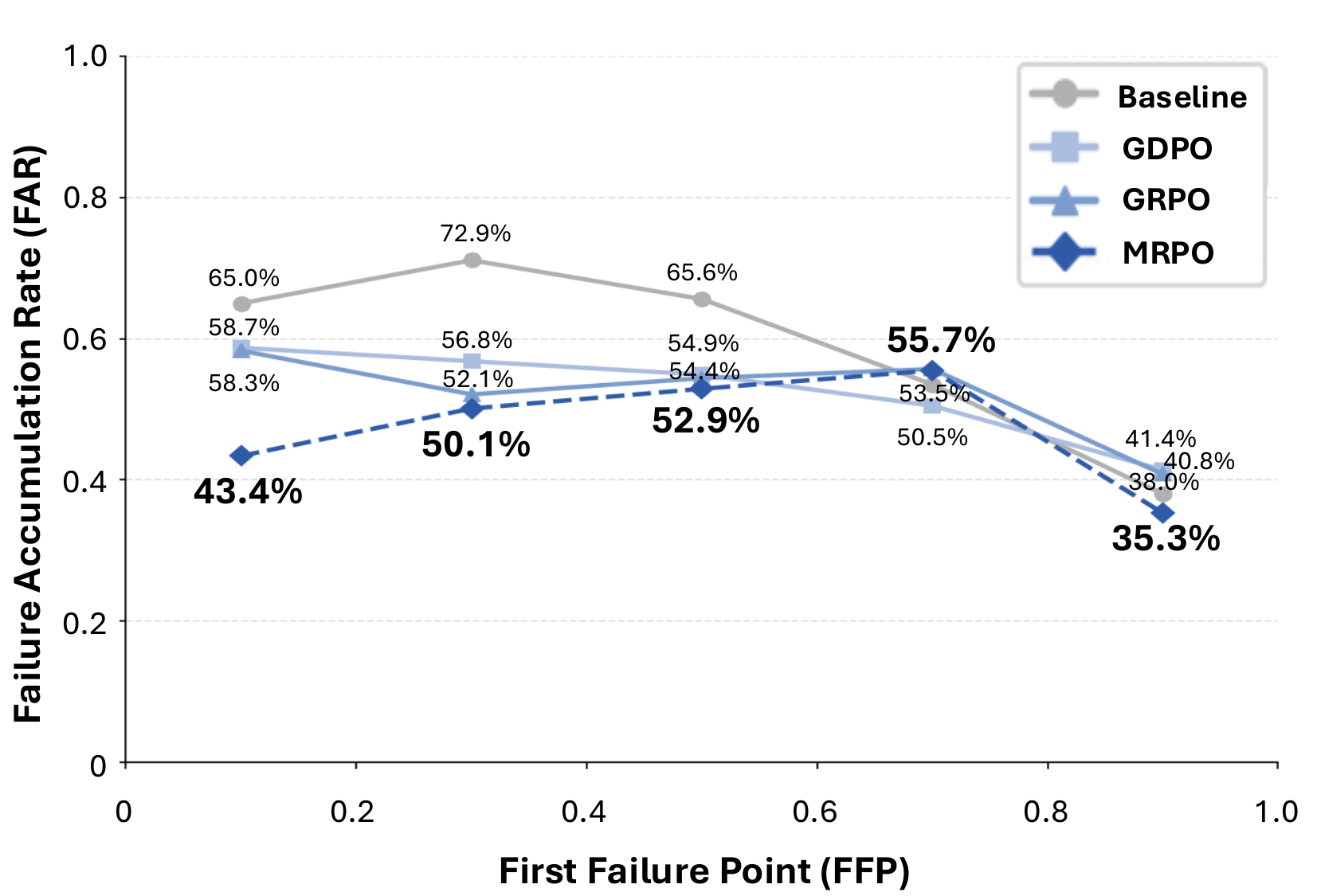}
  \caption{
    \textbf{Failure Accumulation Rate (FAR) across FFP bins.}
    MRPO shows the lowest FAR across all methods, indicating reduced failure accumulation.
  }
  \label{fig:Reasoning_ablation_2}
  \vspace{-0.3cm}
\end{figure}

Figure~\ref{fig:Reasoning_ablation_2} compares the FAR across FFP bins at 0.2 intervals, averaged over the four backbones, where a lower FAR indicates more effective recovery after the first failure. In the Early-to-Mid-Stage range from FFP 0.0 to 0.6, the baseline exhibits consistently high FAR, from 65.0\% to 72.9\%. In contrast, MRPO records the lowest FAR across this range, with a particularly pronounced reduction in the 0.0--0.2 bin, where its FAR of 43.4\% falls markedly below the baseline's 65.0\% as well as GRPO (58.3\%) and GDPO (58.7\%). This indicates that MRPO reduces downstream failure propagation, with the strongest effect when failures begin early.

Together with the FFP results in Figure~\ref{fig:Reasoning_ablation_1}, MRPO mitigates cascading failures along two complementary axes: delaying failure onset and reducing subsequent accumulation. Detailed per-backbone results, an instance-level paired comparison between GRPO and MRPO, and qualitative analyses are provided in Appendix~\ref{app:reasoning_analysis} and Appendix~\ref{app:qualitative_analysis}.

\section{Conclusion}
\label{sec:conclusion}
In this paper, we show that in open-ended medical VQA, early-stage reasoning failures systematically propagate and dominate final prediction errors, exposing a key limitation of outcome-centric approaches.
To address this, we propose MRPO, an RL algorithm that targets the first point of reasoning failure by integrating step-wise reasoning rewards into policy optimization. When the final answer is incorrect, MRPO assigns exponentially larger penalties to earlier invalid steps, correcting root causes of failure while preserving successful trajectories.
Across four multimodal LLM backbones, MRPO consistently outperforms standard GRPO and GDPO, achieving competitive performance with substantially larger medical MLLMs using only 13K training samples. Reasoning analysis confirms that MRPO reduces early-stage reasoning failures and mitigates downstream failure accumulation, validating that our approach addresses the cascading failure problem. These findings suggest that explicitly addressing early reasoning failures offers a promising direction for developing more reliable medical multimodal reasoning.

\section*{Limitations}
\label{sec:limitations}
% Our work has several limitations that warrant discussion. First, MRPO relies on an external LLM judge, specifically GPT-5-mini, to compute the step-wise reasoning process reward during RL training. While this design enables fine-grained sentence-level supervision without training a separate process reward model, it incurs additional API cost and introduces a dependency on the judge's reliability. As shown in Appendix~\ref{app:local_prm}, replacing GPT-5-mini with smaller local process reward models leads to notable performance degradation, indicating that a sufficiently strong judge is required for the step-wise reward signal to be effective. This observation motivates a promising future direction of developing dedicated process reward models for medical VQA that can match the judgment quality of frontier LLMs while eliminating the dependency on external APIs.

Our work has several limitations.

First, MRPO relies on an external LLM judge, GPT-5-mini, to compute the step-wise reasoning process reward during RL training. While this design enables fine-grained sentence-level supervision without training a separate process reward model, it incurs additional API cost and a dependency on the judge; a detailed cost breakdown is in Appendix~\ref{app:api_cost}. As shown in Appendix~\ref{app:local_prm}, GPT-5-mini provides the highest absolute performance ceiling among the four process reward models we evaluate, which is why we adopt it as the default. Conversely, the weakest judge yields a notable degradation, indicating that a sufficiently strong judge is required for the step-wise reward signal to be effective. Strong open-source judges such as MedGemma-27B and Qwen3-VL-30B narrow the gap, making them viable alternatives when API access is constrained, though they do not reach the ceiling of GPT-5-mini. This motivates a promising future direction of training or distilling dedicated medical VQA process reward models that match frontier-LLM judgment quality while eliminating the dependency on external APIs.

Second, the construction of step-wise reasoning rewards in our framework relies on gold reasoning annotations from MedThink~\cite{gai2024medthinkexplainingmedicalvisual}. Extending MRPO to settings without such gold rationales would require annotation-free reward signals or weaker forms of step-level supervision, both promising directions for future investigation.

Third, our evaluation is currently confined to medical VQA. The cascading failure problem and the step-wise advantage reshaping mechanism may generalize to other domains requiring multi-step reasoning, such as scientific question answering or legal reasoning, but we leave the empirical verification of this generalization to future work.

Finally, our models are research prototypes not validated for clinical use; their outputs may contain errors and require expert verification before any clinical application.

\section*{Acknowledgments}

This research was supported by the National Research Foundation of Korea (NRF2023R1A2C3004176), the Ministry of Health \& Welfare, Republic of Korea (HR20C002103), the Ministry of Science and ICT through Seoul National University Hospital (RS-2023-00262002), the National Research Foundation of Korea(NRF) grant funded by the Korea government(MSIT and MOE) (No. RS-2025-16652968), and the Institute of Information \& Communications Technology Planning \& Evaluation(IITP)-ICT Creative Consilience Program grant funded by the Korea government(MSIT) (IITP-2026-RS-2020-II201819).
This research was also supported by the Ministry of Science and ICT(MSIT) and National IT Industry Promotion Agency(NIPA) through the Hyperscale AI Healthcare Service Support Program (grant number : H0701-25-1003), by a grant of the Korea Health Technology R\&D Project through the Korea Health Industry Development Institute (KHIDI), funded by the Ministry of Health \& Welfare, Republic of Korea (grant numbers: RS-2025-25462758 and RS-2022-KH129295), and by the ``Advanced GPU Utilization Support Program'' funded by the Government of the Republic of Korea (Ministry of Science and ICT).
M.S. was supported by (1) No. RS-2024-00509257: Global AI Frontier Lab, and (2) the MSIT(Ministry of Science and ICT), Korea, under the ITRC(Information Technology Research Center) support program(IITP-2026-RS-2024-00438239, 10\%).

% Bibliography entries for the entire Anthology, followed by custom entries
%\bibliography{anthology,custom}
% Custom bibliography entries only
\bibliography{custom}

@inproceedings{chakraborty-etal-2020-biomedbert,
    title = "{B}io{M}ed{BERT}: A Pre-trained Biomedical Language Model for {QA} and {IR}",
    author = "Chakraborty, Souradip  and
      Bisong, Ekaba  and
      Bhatt, Shweta  and
      Wagner, Thomas  and
      Elliott, Riley  and
      Mosconi, Francesco",
    editor = "Scott, Donia  and
      Bel, Nuria  and
      Zong, Chengqing",
    booktitle = "Proceedings of the 28th International Conference on Computational Linguistics",
    month = dec,
    year = "2020",
    address = "Barcelona, Spain (Online)",
    publisher = "International Committee on Computational Linguistics",
    url = "https://aclanthology.org/2020.coling-main.59/",
    doi = "10.18653/v1/2020.coling-main.59",
    pages = "669--679"
}

@article{zheng2023judging,
  title={Judging llm-as-a-judge with mt-bench and chatbot arena},
  author={Zheng, Lianmin and Chiang, Wei-Lin and Sheng, Ying and Zhuang, Siyuan and Wu, Zhanghao and Zhuang, Yonghao and Lin, Zi and Li, Zhuohan and Li, Dacheng and Xing, Eric and others},
  journal={Advances in neural information processing systems},
  year={2023}
}

@misc{shao2024deepseekmathpushinglimitsmathematical,
      title={DeepSeekMath: Pushing the Limits of Mathematical Reasoning in Open Language Models}, 
      author={Zhihong Shao and Peiyi Wang and Qihao Zhu and Runxin Xu and Junxiao Song and Xiao Bi and Haowei Zhang and Mingchuan Zhang and Y. K. Li and Y. Wu and Daya Guo},
      year={2024},
      eprint={2402.03300},
      archivePrefix={arXiv},
      primaryClass={cs.CL},
      url={https://arxiv.org/abs/2402.03300}, 
}

@misc{he2025pefomedparameterefficientfinetuning,
      title={PeFoMed: Parameter Efficient Fine-tuning of Multimodal Large Language Models for Medical Imaging}, 
      author={Jinlong He and Pengfei Li and Gang Liu and Genrong He and Zhaolin Chen and Shenjun Zhong},
      year={2025},
      eprint={2401.02797},
      archivePrefix={arXiv},
      primaryClass={cs.CL},
      url={https://arxiv.org/abs/2401.02797}, 
}

@article{Qwen2.5-VL,
  title={Qwen2.5-VL Technical Report},
  author={Bai, Shuai and Chen, Keqin and Liu, Xuejing and Wang, Jialin and Ge, Wenbin and Song, Sibo and Dang, Kai and Wang, Peng and Wang, Shijie and Tang, Jun and Zhong, Humen and Zhu, Yuanzhi and Yang, Mingkun and Li, Zhaohai and Wan, Jianqiang and Wang, Pengfei and Ding, Wei and Fu, Zheren and Xu, Yiheng and Ye, Jiabo and Zhang, Xi and Xie, Tianbao and Cheng, Zesen and Zhang, Hang and Yang, Zhibo and Xu, Haiyang and Lin, Junyang},
  journal={arXiv preprint arXiv:2502.13923},
  year={2025}
}

@article{Qwen3-VL,
      title={Qwen3-VL Technical Report}, 
      author={Shuai Bai and Yuxuan Cai and Ruizhe Chen and Keqin Chen and Xionghui Chen and Zesen Cheng and Lianghao Deng and Wei Ding and Chang Gao and Chunjiang Ge and Wenbin Ge and Zhifang Guo and Qidong Huang and Jie Huang and Fei Huang and Binyuan Hui and Shutong Jiang and Zhaohai Li and Mingsheng Li and Mei Li and Kaixin Li and Zicheng Lin and Junyang Lin and Xuejing Liu and Jiawei Liu and Chenglong Liu and Yang Liu and Dayiheng Liu and Shixuan Liu and Dunjie Lu and Ruilin Luo and Chenxu Lv and Rui Men and Lingchen Meng and Xuancheng Ren and Xingzhang Ren and Sibo Song and Yuchong Sun and Jun Tang and Jianhong Tu and Jianqiang Wan and Peng Wang and Pengfei Wang and Qiuyue Wang and Yuxuan Wang and Tianbao Xie and Yiheng Xu and Haiyang Xu and Jin Xu and Zhibo Yang and Mingkun Yang and Jianxin Yang and An Yang and Bowen Yu and Fei Zhang and Hang Zhang and Xi Zhang and Bo Zheng and Humen Zhong and Jingren Zhou and Fan Zhou and Jing Zhou and Yuanzhi Zhu and Ke Zhu},
	  journal={arXiv preprint arXiv:2511.21631},
      year={2025}
}

@article{zhang2019bertscore,
  title={Bertscore: Evaluating text generation with bert},
  author={Zhang, Tianyi and Kishore, Varsha and Wu, Felix and Weinberger, Kilian Q and Artzi, Yoav},
  journal={arXiv preprint arXiv:1904.09675},
  year={2019}
}

@article{jeong2024olaph,
  title={Olaph: Improving factuality in biomedical long-form question answering},
  author={Jeong, Minbyul and Hwang, Hyeon and Yoon, Chanwoong and Lee, Taewhoo and Kang, Jaewoo},
  journal={arXiv preprint arXiv:2405.12701},
  year={2024}
}

@misc{zhu2025internvl3exploringadvancedtraining,
      title={InternVL3: Exploring Advanced Training and Test-Time Recipes for Open-Source Multimodal Models}, 
      author={Jinguo Zhu and Weiyun Wang and Zhe Chen and Zhaoyang Liu and Shenglong Ye and Lixin Gu and Hao Tian and Yuchen Duan and Weijie Su and Jie Shao and Zhangwei Gao and Erfei Cui and Xuehui Wang and Yue Cao and Yangzhou Liu and Xingguang Wei and Hongjie Zhang and Haomin Wang and Weiye Xu and Hao Li and Jiahao Wang and Nianchen Deng and Songze Li and Yinan He and Tan Jiang and Jiapeng Luo and Yi Wang and Conghui He and Botian Shi and Xingcheng Zhang and Wenqi Shao and Junjun He and Yingtong Xiong and Wenwen Qu and Peng Sun and Penglong Jiao and Han Lv and Lijun Wu and Kaipeng Zhang and Huipeng Deng and Jiaye Ge and Kai Chen and Limin Wang and Min Dou and Lewei Lu and Xizhou Zhu and Tong Lu and Dahua Lin and Yu Qiao and Jifeng Dai and Wenhai Wang},
      year={2025},
      eprint={2504.10479},
      archivePrefix={arXiv},
      primaryClass={cs.CV},
      url={https://arxiv.org/abs/2504.10479}, 
}

@misc{chen2024huatuogptvisioninjectingmedicalvisual,
      title={HuatuoGPT-Vision, Towards Injecting Medical Visual Knowledge into Multimodal LLMs at Scale}, 
      author={Junying Chen and Chi Gui and Ruyi Ouyang and Anningzhe Gao and Shunian Chen and Guiming Hardy Chen and Xidong Wang and Ruifei Zhang and Zhenyang Cai and Ke Ji and Guangjun Yu and Xiang Wan and Benyou Wang},
      year={2024},
      eprint={2406.19280},
      archivePrefix={arXiv},
      primaryClass={cs.CV},
      url={https://arxiv.org/abs/2406.19280}, 
}

@misc{lasateam2025lingshugeneralistfoundationmodel,
      title={Lingshu: A Generalist Foundation Model for Unified Multimodal Medical Understanding and Reasoning}, 
      author={LASA Team and Weiwen Xu and Hou Pong Chan and Long Li and Mahani Aljunied and Ruifeng Yuan and Jianyu Wang and Chenghao Xiao and Guizhen Chen and Chaoqun Liu and Zhaodonghui Li and Yu Sun and Junao Shen and Chaojun Wang and Jie Tan and Deli Zhao and Tingyang Xu and Hao Zhang and Yu Rong},
      year={2025},
      eprint={2506.07044},
      archivePrefix={arXiv},
      primaryClass={cs.CL},
      url={https://arxiv.org/abs/2506.07044}, 
}

@article{vqarad2018,
  title        = {A dataset of clinically generated visual questions and answers about radiology images},
  author       = {Lau, Jason J. and Gayen, Soumya and Ben Abacha, Asma and Demner-Fushman, Dina},
  journal      = {Scientific Data},
  year         = {2018},
  volume       = {5},
  number       = {1},
  pages        = {180251},
  doi          = {10.1038/sdata.2018.251},
  url          = {https://doi.org/10.1038/sdata.2018.251},
  issn         = {2052-4463},
  date         = {2018-11-20}
}

@misc{liu2021slakesemanticallylabeledknowledgeenhanceddataset,
      title={SLAKE: A Semantically-Labeled Knowledge-Enhanced Dataset for Medical Visual Question Answering}, 
      author={Bo Liu and Li-Ming Zhan and Li Xu and Lin Ma and Yan Yang and Xiao-Ming Wu},
      year={2021},
      eprint={2102.09542},
      archivePrefix={arXiv},
      primaryClass={cs.CV},
      url={https://arxiv.org/abs/2102.09542}, 
}

@misc{he2020pathvqa30000questionsmedical,
      title={PathVQA: 30000+ Questions for Medical Visual Question Answering}, 
      author={Xuehai He and Yichen Zhang and Luntian Mou and Eric Xing and Pengtao Xie},
      year={2020},
      eprint={2003.10286},
      archivePrefix={arXiv},
      primaryClass={cs.CL},
      url={https://arxiv.org/abs/2003.10286}, 
}

@misc{gai2024medthinkexplainingmedicalvisual,
      title={MedThink: Explaining Medical Visual Question Answering via Multimodal Decision-Making Rationale}, 
      author={Xiaotang Gai and Chenyi Zhou and Jiaxiang Liu and Yang Feng and Jian Wu and Zuozhu Liu},
      year={2024},
      eprint={2404.12372},
      archivePrefix={arXiv},
      primaryClass={cs.CV},
      url={https://arxiv.org/abs/2404.12372}, 
}

@misc{sun2025chirono1ignitingmultimodallarge,
      title={Chiron-o1: Igniting Multimodal Large Language Models towards Generalizable Medical Reasoning via Mentor-Intern Collaborative Search}, 
      author={Haoran Sun and Yankai Jiang and Wenjie Lou and Yujie Zhang and Wenjie Li and Lilong Wang and Mianxin Liu and Lei Liu and Xiaosong Wang},
      year={2025},
      eprint={2506.16962},
      archivePrefix={arXiv},
      primaryClass={cs.CV},
      url={https://arxiv.org/abs/2506.16962}, 
}

@misc{huang2025medvlthinkersimplebaselinesmultimodal,
      title={MedVLThinker: Simple Baselines for Multimodal Medical Reasoning}, 
      author={Xiaoke Huang and Juncheng Wu and Hui Liu and Xianfeng Tang and Yuyin Zhou},
      year={2025},
      eprint={2508.02669},
      archivePrefix={arXiv},
      primaryClass={cs.CV},
      url={https://arxiv.org/abs/2508.02669}, 
}

@misc{fan2025chestxreasoneradvancingradiologyfoundation,
      title={ChestX-Reasoner: Advancing Radiology Foundation Models with Reasoning through Step-by-Step Verification}, 
      author={Ziqing Fan and Cheng Liang and Chaoyi Wu and Ya Zhang and Yanfeng Wang and Weidi Xie},
      year={2025},
      eprint={2504.20930},
      archivePrefix={arXiv},
      primaryClass={cs.AI},
      url={https://arxiv.org/abs/2504.20930}, 
}

@misc{shen2025vlmr1stablegeneralizabler1style,
      title={VLM-R1: A Stable and Generalizable R1-style Large Vision-Language Model}, 
      author={Haozhan Shen and Peng Liu and Jingcheng Li and Chunxin Fang and Yibo Ma and Jiajia Liao and Qiaoli Shen and Zilun Zhang and Kangjia Zhao and Qianqian Zhang and Ruochen Xu and Tiancheng Zhao},
      year={2025},
      eprint={2504.07615},
      archivePrefix={arXiv},
      primaryClass={cs.CV},
      url={https://arxiv.org/abs/2504.07615}, 
}

@misc{lai2025medr1reinforcementlearninggeneralizable,
      title={Med-R1: Reinforcement Learning for Generalizable Medical Reasoning in Vision-Language Models}, 
      author={Yuxiang Lai and Jike Zhong and Ming Li and Shitian Zhao and Yuheng Li and Konstantinos Psounis and Xiaofeng Yang},
      year={2025},
      eprint={2503.13939},
      archivePrefix={arXiv},
      primaryClass={cs.CV},
      url={https://arxiv.org/abs/2503.13939}, 
}

@misc{liu2025breakingrewardcollapseadaptive,
      title={Breaking Reward Collapse: Adaptive Reinforcement for Open-ended Medical Reasoning with Enhanced Semantic Discrimination}, 
      author={Yizhou Liu and Jingwei Wei and Zizhi Chen and Minghao Han and Xukun Zhang and Keliang Liu and Lihua Zhang},
      year={2025},
      eprint={2508.12957},
      archivePrefix={arXiv},
      primaryClass={cs.CV},
      url={https://arxiv.org/abs/2508.12957}, 
}

@article{Kim2025,
  title        = {Small language models learn enhanced reasoning skills from medical textbooks},
  author       = {Kim, Hyunjae and Hwang, Hyeon and Lee, Jiwoo and Park, Sihyeon and Kim, Dain and Lee, Taewhoo and Yoon, Chanwoong and Sohn, Jiwoong and Park, Jungwoo and Reykhart, Olga and Fetherston, Thomas and Choi, Donghee and Kwak, Soo Heon and Chen, Qingyu and Kang, Jaewoo},
  journal      = {npj Digital Medicine},
  year         = {2025},
  volume       = {8},
  number       = {1},
  pages        = {240},
  doi          = {10.1038/s41746-025-01653-8},
  url          = {https://doi.org/10.1038/s41746-025-01653-8},
  issn         = {2398-6352},
  month        = may
}

@misc{chaudhari2024rlhfdecipheredcriticalanalysis,
      title={RLHF Deciphered: A Critical Analysis of Reinforcement Learning from Human Feedback for LLMs}, 
      author={Shreyas Chaudhari and Pranjal Aggarwal and Vishvak Murahari and Tanmay Rajpurohit and Ashwin Kalyan and Karthik Narasimhan and Ameet Deshpande and Bruno Castro da Silva},
      year={2024},
      eprint={2404.08555},
      archivePrefix={arXiv},
      primaryClass={cs.LG},
      url={https://arxiv.org/abs/2404.08555}, 
}

@misc{mu2025medcegreinforcingverifiablemedical,
      title={MedCEG: Reinforcing Verifiable Medical Reasoning with Critical Evidence Graph}, 
      author={Linjie Mu and Yannian Gu and Zhongzhen Huang and Yakun Zhu and Shaoting Zhang and Xiaofan Zhang},
      year={2025},
      eprint={2512.13510},
      archivePrefix={arXiv},
      primaryClass={cs.AI},
      url={https://arxiv.org/abs/2512.13510}, 
}

@misc{xie2025capoenhancingllmreasoning,
      title={CAPO: Towards Enhancing LLM Reasoning through Generative Credit Assignment}, 
      author={Guofu Xie and Yunsheng Shi and Hongtao Tian and Ting Yao and Xiao Zhang},
      year={2025},
      eprint={2508.02298},
      archivePrefix={arXiv},
      primaryClass={cs.LG},
      url={https://arxiv.org/abs/2508.02298}, 
}

@inproceedings{lin-2004-rouge,
    title = "{ROUGE}: A Package for Automatic Evaluation of Summaries",
    author = "Lin, Chin-Yew",
    booktitle = "Text Summarization Branches Out",
    month = jul,
    year = "2004",
    address = "Barcelona, Spain",
    publisher = "Association for Computational Linguistics",
    url = "https://aclanthology.org/W04-1013/",
    pages = "74--81"
}

@inproceedings{papineni-etal-2002-bleu,
    title = "{B}leu: a Method for Automatic Evaluation of Machine Translation",
    author = "Papineni, Kishore  and
      Roukos, Salim  and
      Ward, Todd  and
      Zhu, Wei-Jing",
    editor = "Isabelle, Pierre  and
      Charniak, Eugene  and
      Lin, Dekang",
    booktitle = "Proceedings of the 40th Annual Meeting of the Association for Computational Linguistics",
    month = jul,
    year = "2002",
    address = "Philadelphia, Pennsylvania, USA",
    publisher = "Association for Computational Linguistics",
    url = "https://aclanthology.org/P02-1040/",
    doi = "10.3115/1073083.1073135",
    pages = "311--318"
}

@misc{zhang2024pmcvqavisualinstructiontuning,
      title={PMC-VQA: Visual Instruction Tuning for Medical Visual Question Answering}, 
      author={Xiaoman Zhang and Chaoyi Wu and Ziheng Zhao and Weixiong Lin and Ya Zhang and Yanfeng Wang and Weidi Xie},
      year={2024},
      eprint={2305.10415},
      archivePrefix={arXiv},
      primaryClass={cs.CV},
      url={https://arxiv.org/abs/2305.10415}, 
}

@Inproceedings{ImageCLEF-VQA-Med2021,
                author = {Asma {Ben Abacha} and Mourad Sarrouti and Dina Demner-Fushman and Sadid A. Hasan and Henning M\"uller},
                title = {Overview of the VQA-Med Task at ImageCLEF 2021: Visual Question Answering and Generation in the Medical Domain},
                booktitle = {CLEF 2021 Working Notes},
                series = {{CEUR} Workshop Proceedings},
                year = {2021},
                publisher = {CEUR-WS.org},
                month = {September 21-24},
                address = {Bucharest, Romania} 
}

@misc{seyfioglu2025quiltllavavisualinstructiontuning,
      title={Quilt-LLaVA: Visual Instruction Tuning by Extracting Localized Narratives from Open-Source Histopathology Videos}, 
      author={Mehmet Saygin Seyfioglu and Wisdom O. Ikezogwo and Fatemeh Ghezloo and Ranjay Krishna and Linda Shapiro},
      year={2025},
      eprint={2312.04746},
      archivePrefix={arXiv},
      primaryClass={cs.CV},
      url={https://arxiv.org/abs/2312.04746}, 
}

@misc{li2023llavamedtraininglargelanguageandvision,
      title={LLaVA-Med: Training a Large Language-and-Vision Assistant for Biomedicine in One Day}, 
      author={Chunyuan Li and Cliff Wong and Sheng Zhang and Naoto Usuyama and Haotian Liu and Jianwei Yang and Tristan Naumann and Hoifung Poon and Jianfeng Gao},
      year={2023},
      eprint={2306.00890},
      archivePrefix={arXiv},
      primaryClass={cs.CV},
      url={https://arxiv.org/abs/2306.00890}, 
}

@misc{dai2025qoqmedbuildingmultimodalclinical,
      title={QoQ-Med: Building Multimodal Clinical Foundation Models with Domain-Aware GRPO Training}, 
      author={Wei Dai and Peilin Chen and Chanakya Ekbote and Paul Pu Liang},
      year={2025},
      eprint={2506.00711},
      archivePrefix={arXiv},
      primaryClass={cs.LG},
      url={https://arxiv.org/abs/2506.00711}, 
}

@misc{pan2025medvlmr1incentivizingmedicalreasoning,
      title={MedVLM-R1: Incentivizing Medical Reasoning Capability of Vision-Language Models (VLMs) via Reinforcement Learning}, 
      author={Jiazhen Pan and Che Liu and Junde Wu and Fenglin Liu and Jiayuan Zhu and Hongwei Bran Li and Chen Chen and Cheng Ouyang and Daniel Rueckert},
      year={2025},
      eprint={2502.19634},
      archivePrefix={arXiv},
      primaryClass={cs.CV},
      url={https://arxiv.org/abs/2502.19634}, 
}

@misc{paszke2019pytorchimperativestylehighperformance,
      title={PyTorch: An Imperative Style, High-Performance Deep Learning Library}, 
      author={Adam Paszke and Sam Gross and Francisco Massa and Adam Lerer and James Bradbury and Gregory Chanan and Trevor Killeen and Zeming Lin and Natalia Gimelshein and Luca Antiga and Alban Desmaison and Andreas Köpf and Edward Yang and Zach DeVito and Martin Raison and Alykhan Tejani and Sasank Chilamkurthy and Benoit Steiner and Lu Fang and Junjie Bai and Soumith Chintala},
      year={2019},
      eprint={1912.01703},
      archivePrefix={arXiv},
      primaryClass={cs.LG},
      url={https://arxiv.org/abs/1912.01703}, 
}

@misc{dao2023flashattention2fasterattentionbetter,
      title={FlashAttention-2: Faster Attention with Better Parallelism and Work Partitioning}, 
      author={Tri Dao},
      year={2023},
      eprint={2307.08691},
      archivePrefix={arXiv},
      primaryClass={cs.LG},
      url={https://arxiv.org/abs/2307.08691}, 
}

@misc{vteam2025glm45vglm41vthinkingversatilemultimodal,
      title={GLM-4.5V and GLM-4.1V-Thinking: Towards Versatile Multimodal Reasoning with Scalable Reinforcement Learning}, 
      author={V Team and Wenyi Hong and Wenmeng Yu and Xiaotao Gu and Guo Wang and Guobing Gan and Haomiao Tang and Jiale Cheng and Ji Qi and Junhui Ji and Lihang Pan and Shuaiqi Duan and Weihan Wang and Yan Wang and Yean Cheng and Zehai He and Zhe Su and Zhen Yang and Ziyang Pan and Aohan Zeng and Baoxu Wang and Bin Chen and Boyan Shi and Changyu Pang and Chenhui Zhang and Da Yin and Fan Yang and Guoqing Chen and Haochen Li and Jiale Zhu and Jiali Chen and Jiaxing Xu and Jiazheng Xu and Jing Chen and Jinghao Lin and Jinhao Chen and Jinjiang Wang and Junjie Chen and Leqi Lei and Letian Gong and Leyi Pan and Mingdao Liu and Mingde Xu and Mingzhi Zhang and Qinkai Zheng and Ruiliang Lyu and Shangqin Tu and Sheng Yang and Shengbiao Meng and Shi Zhong and Shiyu Huang and Shuyuan Zhao and Siyan Xue and Tianshu Zhang and Tianwei Luo and Tianxiang Hao and Tianyu Tong and Wei Jia and Wenkai Li and Xiao Liu and Xiaohan Zhang and Xin Lyu and Xinyu Zhang and Xinyue Fan and Xuancheng Huang and Yadong Xue and Yanfeng Wang and Yanling Wang and Yanzi Wang and Yifan An and Yifan Du and Yiheng Huang and Yilin Niu and Yiming Shi and Yu Wang and Yuan Wang and Yuanchang Yue and Yuchen Li and Yusen Liu and Yutao Zhang and Yuting Wang and Yuxuan Zhang and Zhao Xue and Zhengxiao Du and Zhenyu Hou and Zihan Wang and Peng Zhang and Debing Liu and Bin Xu and Juanzi Li and Minlie Huang and Yuxiao Dong and Jie Tang},
      year={2026},
      eprint={2507.01006},
      archivePrefix={arXiv},
      primaryClass={cs.CV},
      url={https://arxiv.org/abs/2507.01006}, 
}

@misc{yang2025r4bincentivizinggeneralpurposeautothinking,
      title={R-4B: Incentivizing General-Purpose Auto-Thinking Capability in MLLMs via Bi-Mode Annealing and Reinforce Learning}, 
      author={Qi Yang and Bolin Ni and Shiming Xiang and Han Hu and Houwen Peng and Jie Jiang},
      year={2025},
      eprint={2508.21113},
      archivePrefix={arXiv},
      primaryClass={cs.CV},
      url={https://arxiv.org/abs/2508.21113}, 
}

@misc{singhal2022largelanguagemodelsencode,
      title={Large Language Models Encode Clinical Knowledge}, 
      author={Karan Singhal and Shekoofeh Azizi and Tao Tu and S. Sara Mahdavi and Jason Wei and Hyung Won Chung and Nathan Scales and Ajay Tanwani and Heather Cole-Lewis and Stephen Pfohl and Perry Payne and Martin Seneviratne and Paul Gamble and Chris Kelly and Nathaneal Scharli and Aakanksha Chowdhery and Philip Mansfield and Blaise Aguera y Arcas and Dale Webster and Greg S. Corrado and Yossi Matias and Katherine Chou and Juraj Gottweis and Nenad Tomasev and Yun Liu and Alvin Rajkomar and Joelle Barral and Christopher Semturs and Alan Karthikesalingam and Vivek Natarajan},
      year={2022},
      eprint={2212.13138},
      archivePrefix={arXiv},
      primaryClass={cs.CL},
      url={https://arxiv.org/abs/2212.13138}, 
}

@misc{sellergren2025medgemmatechnicalreport,
      title={MedGemma Technical Report}, 
      author={Andrew Sellergren and Sahar Kazemzadeh and Tiam Jaroensri and Atilla Kiraly and Madeleine Traverse and Timo Kohlberger and Shawn Xu and Fayaz Jamil and Cían Hughes and Charles Lau and Justin Chen and Fereshteh Mahvar and Liron Yatziv and Tiffany Chen and Bram Sterling and Stefanie Anna Baby and Susanna Maria Baby and Jeremy Lai and Samuel Schmidgall and Lu Yang and Kejia Chen and Per Bjornsson and Shashir Reddy and Ryan Brush and Kenneth Philbrick and Mercy Asiedu and Ines Mezerreg and Howard Hu and Howard Yang and Richa Tiwari and Sunny Jansen and Preeti Singh and Yun Liu and Shekoofeh Azizi and Aishwarya Kamath and Johan Ferret and Shreya Pathak and Nino Vieillard and Ramona Merhej and Sarah Perrin and Tatiana Matejovicova and Alexandre Ramé and Morgane Riviere and Louis Rouillard and Thomas Mesnard and Geoffrey Cideron and Jean-bastien Grill and Sabela Ramos and Edouard Yvinec and Michelle Casbon and Elena Buchatskaya and Jean-Baptiste Alayrac and Dmitry Lepikhin and Vlad Feinberg and Sebastian Borgeaud and Alek Andreev and Cassidy Hardin and Robert Dadashi and Léonard Hussenot and Armand Joulin and Olivier Bachem and Yossi Matias and Katherine Chou and Avinatan Hassidim and Kavi Goel and Clement Farabet and Joelle Barral and Tris Warkentin and Jonathon Shlens and David Fleet and Victor Cotruta and Omar Sanseviero and Gus Martins and Phoebe Kirk and Anand Rao and Shravya Shetty and David F. Steiner and Can Kirmizibayrak and Rory Pilgrim and Daniel Golden and Lin Yang},
      year={2025},
      eprint={2507.05201},
      archivePrefix={arXiv},
      primaryClass={cs.AI},
      url={https://arxiv.org/abs/2507.05201}, 
}

@misc{li2025reasoningmodelshallucinatemore,
      title={Reasoning Models Hallucinate More: Factuality-Aware Reinforcement Learning for Large Reasoning Models}, 
      author={Junyi Li and Hwee Tou Ng},
      year={2025},
      eprint={2505.24630},
      archivePrefix={arXiv},
      primaryClass={cs.CL},
      url={https://arxiv.org/abs/2505.24630}, 
}

@misc{tan2025gtpogrpostokensequencelevel,
      title={GTPO and GRPO-S: Token and Sequence-Level Reward Shaping with Policy Entropy}, 
      author={Hongze Tan and Zihan Wang and Jianfei Pan and Jinghao Lin and Hao Wang and Yifan Wu and Tao Chen and Zhihang Zheng and Zhihao Tang and Haihua Yang},
      year={2026},
      eprint={2508.04349},
      archivePrefix={arXiv},
      primaryClass={cs.CL},
      url={https://arxiv.org/abs/2508.04349}, 
}

@misc{parthasarathi2025grpolambdacreditassignmentimproves,
      title={GRPO-$\lambda$: Credit Assignment improves LLM Reasoning}, 
      author={Prasanna Parthasarathi and Mathieu Reymond and Boxing Chen and Yufei Cui and Sarath Chandar},
      year={2025},
      eprint={2510.00194},
      archivePrefix={arXiv},
      primaryClass={cs.LG},
      url={https://arxiv.org/abs/2510.00194}, 
}

@article{LandisKoch1977,
  title     = {The measurement of observer agreement for categorical data},
  author    = {Landis, J. Richard and Koch, Gary G.},
  journal   = {Biometrics},
  year      = {1977},
  volume    = {33},
  number    = {1},
  pages     = {159--174},
  issn      = {0006-341X},
  language  = {eng},
  doi       = {},
  pmid      = {843571}
}

@misc{su2025gmaivlr1harnessingreinforcementlearning,
      title={GMAI-VL-R1: Harnessing Reinforcement Learning for Multimodal Medical Reasoning}, 
      author={Yanzhou Su and Tianbin Li and Jiyao Liu and Chenglong Ma and Junzhi Ning and Cheng Tang and Sibo Ju and Jin Ye and Pengcheng Chen and Ming Hu and Shixiang Tang and Lihao Liu and Bin Fu and Wenqi Shao and Xiaowei Hu and Xiangwen Liao and Yuanfeng Ji and Junjun He},
      year={2025},
      eprint={2504.01886},
      archivePrefix={arXiv},
      primaryClass={cs.CV},
      url={https://arxiv.org/abs/2504.01886}, 
}

@misc{zhang2025pathor1multimodalreinforcementlearningbased,
      title={Patho-R1: A Multimodal Reinforcement Learning-Based Pathology Expert Reasoner}, 
      author={Wenchuan Zhang and Penghao Zhang and Jingru Guo and Tao Cheng and Jie Chen and Shuwan Zhang and Zhang Zhang and Yuhao Yi and Hong Bu},
      year={2025},
      eprint={2505.11404},
      archivePrefix={arXiv},
      primaryClass={cs.CV},
      url={https://arxiv.org/abs/2505.11404}, 
}

@misc{zhang2025medtvtr1multimodalllmempowering,
      title={MedTVT-R1: A Multimodal LLM Empowering Medical Reasoning and Diagnosis}, 
      author={Yuting Zhang and Kaishen Yuan and Hao Lu and Yutao Yue and Jintai Chen and Kaishun Wu},
      year={2025},
      eprint={2506.18512},
      archivePrefix={arXiv},
      primaryClass={eess.IV},
      url={https://arxiv.org/abs/2506.18512}, 
}

@misc{xie2025medtrinity25mlargescalemultimodaldataset,
      title={MedTrinity-25M: A Large-scale Multimodal Dataset with Multigranular Annotations for Medicine}, 
      author={Yunfei Xie and Ce Zhou and Lang Gao and Juncheng Wu and Xianhang Li and Hong-Yu Zhou and Sheng Liu and Lei Xing and James Zou and Cihang Xie and Yuyin Zhou},
      year={2025},
      eprint={2408.02900},
      archivePrefix={arXiv},
      primaryClass={cs.CV},
      url={https://arxiv.org/abs/2408.02900}, 
}

@misc{liu2023visualinstructiontuning,
      title={Visual Instruction Tuning}, 
      author={Haotian Liu and Chunyuan Li and Qingyang Wu and Yong Jae Lee},
      year={2023},
      eprint={2304.08485},
      archivePrefix={arXiv},
      primaryClass={cs.CV},
      url={https://arxiv.org/abs/2304.08485}, 
}

@article{Guo_2025,
   title={DeepSeek-R1 incentivizes reasoning in LLMs through reinforcement learning},
   volume={645},
   ISSN={1476-4687},
   url={http://dx.doi.org/10.1038/s41586-025-09422-z},
   DOI={10.1038/s41586-025-09422-z},
   number={8081},
   journal={Nature},
   publisher={Springer Science and Business Media LLC},
   author={Guo, Daya and Yang, Dejian and Zhang, Haowei and Song, Junxiao and Wang, Peiyi and Zhu, Qihao and Xu, Runxin and Zhang, Ruoyu and Ma, Shirong and Bi, Xiao and Zhang, Xiaokang and Yu, Xingkai and Wu, Yu and Wu, Z. F. and Gou, Zhibin and Shao, Zhihong and Li, Zhuoshu and Gao, Ziyi and Liu, Aixin and Xue, Bing and Wang, Bingxuan and Wu, Bochao and Feng, Bei and Lu, Chengda and Zhao, Chenggang and Deng, Chengqi and Ruan, Chong and Dai, Damai and Chen, Deli and Ji, Dongjie and Li, Erhang and Lin, Fangyun and Dai, Fucong and Luo, Fuli and Hao, Guangbo and Chen, Guanting and Li, Guowei and Zhang, H. and Xu, Hanwei and Ding, Honghui and Gao, Huazuo and Qu, Hui and Li, Hui and Guo, Jianzhong and Li, Jiashi and Chen, Jingchang and Yuan, Jingyang and Tu, Jinhao and Qiu, Junjie and Li, Junlong and Cai, J. L. and Ni, Jiaqi and Liang, Jian and Chen, Jin and Dong, Kai and Hu, Kai and You, Kaichao and Gao, Kaige and Guan, Kang and Huang, Kexin and Yu, Kuai and Wang, Lean and Zhang, Lecong and Zhao, Liang and Wang, Litong and Zhang, Liyue and Xu, Lei and Xia, Leyi and Zhang, Mingchuan and Zhang, Minghua and Tang, Minghui and Zhou, Mingxu and Li, Meng and Wang, Miaojun and Li, Mingming and Tian, Ning and Huang, Panpan and Zhang, Peng and Wang, Qiancheng and Chen, Qinyu and Du, Qiushi and Ge, Ruiqi and Zhang, Ruisong and Pan, Ruizhe and Wang, Runji and Chen, R. J. and Jin, R. L. and Chen, Ruyi and Lu, Shanghao and Zhou, Shangyan and Chen, Shanhuang and Ye, Shengfeng and Wang, Shiyu and Yu, Shuiping and Zhou, Shunfeng and Pan, Shuting and Li, S. S. and Zhou, Shuang and Wu, Shaoqing and Yun, Tao and Pei, Tian and Sun, Tianyu and Wang, T. and Zeng, Wangding and Liu, Wen and Liang, Wenfeng and Gao, Wenjun and Yu, Wenqin and Zhang, Wentao and Xiao, W. L. and An, Wei and Liu, Xiaodong and Wang, Xiaohan and Chen, Xiaokang and Nie, Xiaotao and Cheng, Xin and Liu, Xin and Xie, Xin and Liu, Xingchao and Yang, Xinyu and Li, Xinyuan and Su, Xuecheng and Lin, Xuheng and Li, X. Q. and Jin, Xiangyue and Shen, Xiaojin and Chen, Xiaosha and Sun, Xiaowen and Wang, Xiaoxiang and Song, Xinnan and Zhou, Xinyi and Wang, Xianzu and Shan, Xinxia and Li, Y. K. and Wang, Y. Q. and Wei, Y. X. and Zhang, Yang and Xu, Yanhong and Li, Yao and Zhao, Yao and Sun, Yaofeng and Wang, Yaohui and Yu, Yi and Zhang, Yichao and Shi, Yifan and Xiong, Yiliang and He, Ying and Piao, Yishi and Wang, Yisong and Tan, Yixuan and Ma, Yiyang and Liu, Yiyuan and Guo, Yongqiang and Ou, Yuan and Wang, Yuduan and Gong, Yue and Zou, Yuheng and He, Yujia and Xiong, Yunfan and Luo, Yuxiang and You, Yuxiang and Liu, Yuxuan and Zhou, Yuyang and Zhu, Y. X. and Huang, Yanping and Li, Yaohui and Zheng, Yi and Zhu, Yuchen and Ma, Yunxian and Tang, Ying and Zha, Yukun and Yan, Yuting and Ren, Z. Z. and Ren, Zehui and Sha, Zhangli and Fu, Zhe and Xu, Zhean and Xie, Zhenda and Zhang, Zhengyan and Hao, Zhewen and Ma, Zhicheng and Yan, Zhigang and Wu, Zhiyu and Gu, Zihui and Zhu, Zijia and Liu, Zijun and Li, Zilin and Xie, Ziwei and Song, Ziyang and Pan, Zizheng and Huang, Zhen and Xu, Zhipeng and Zhang, Zhongyu and Zhang, Zhen},
   year={2025},
   month=sep, pages={633–638} }

@misc{hu2021loralowrankadaptationlarge,
      title={LoRA: Low-Rank Adaptation of Large Language Models}, 
      author={Edward J. Hu and Yelong Shen and Phillip Wallis and Zeyuan Allen-Zhu and Yuanzhi Li and Shean Wang and Lu Wang and Weizhu Chen},
      year={2021},
      eprint={2106.09685},
      archivePrefix={arXiv},
      primaryClass={cs.CL},
      url={https://arxiv.org/abs/2106.09685}, 
}

@article{PhysioNet-mimic-ext-mimic-cxr-vqa-1.0.0,
  author = {Bae, Seongsu and Kyung, Daeun and Ryu, Jaehee and Cho, Eunbyeol and Lee, Gyubok and Kweon, Sunjun and Oh, Jungwoo and JI, Lei and Chang, Eric and Kim, Tackeun and Choi, Edward},
  title = {{MIMIC-Ext-MIMIC-CXR-VQA: A Complex, Diverse, And Large-Scale Visual Question Answering Dataset for Chest X-ray Images}},
  journal = {{PhysioNet}},
  year = {2024},
  month = jul,
  note = {Version 1.0.0},
  doi = {10.13026/deqx-d943},
  url = {https://doi.org/10.13026/deqx-d943}
}

@misc{butsanets2026radimagenetvqalargescalectmri,
      title={RadImageNet-VQA: A Large-Scale CT and MRI Dataset for Radiologic Visual Question Answering}, 
      author={Léo Butsanets and Charles Corbière and Julien Khlaut and Pierre Manceron and Corentin Dancette},
      year={2026},
      eprint={2512.17396},
      archivePrefix={arXiv},
      primaryClass={cs.CV},
      url={https://arxiv.org/abs/2512.17396}, 
}

@misc{liu2026gdpogrouprewarddecouplednormalization,
      title={GDPO: Group reward-Decoupled Normalization Policy Optimization for Multi-reward RL Optimization}, 
      author={Shih-Yang Liu and Xin Dong and Ximing Lu and Shizhe Diao and Peter Belcak and Mingjie Liu and Min-Hung Chen and Hongxu Yin and Yu-Chiang Frank Wang and Kwang-Ting Cheng and Yejin Choi and Jan Kautz and Pavlo Molchanov},
      year={2026},
      eprint={2601.05242},
      archivePrefix={arXiv},
      primaryClass={cs.CL},
      url={https://arxiv.org/abs/2601.05242}, 
}

@misc{yun2025medprmmedicalreasoningmodels,
      title={Med-PRM: Medical Reasoning Models with Stepwise, Guideline-verified Process Rewards}, 
      author={Jaehoon Yun and Jiwoong Sohn and Jungwoo Park and Hyunjae Kim and Xiangru Tang and Yanjun Shao and Yonghoe Koo and Minhyeok Ko and Qingyu Chen and Mark Gerstein and Michael Moor and Jaewoo Kang},
      year={2025},
      eprint={2506.11474},
      archivePrefix={arXiv},
      primaryClass={cs.CL},
      url={https://arxiv.org/abs/2506.11474}, 
}

@misc{yang2026medreflmedicalreasoningenhancement,
      title={Med-REFL: Medical Reasoning Enhancement via Self-Corrected Fine-grained Reflection}, 
      author={Zongxian Yang and Jiayu Qian and Zegao Peng and Haoyu Zhang and Yu-An Huang and KC Tan and Zhi-An Huang},
      year={2026},
      eprint={2506.13793},
      archivePrefix={arXiv},
      primaryClass={cs.AI},
      url={https://arxiv.org/abs/2506.13793}, 
}

@misc{zhi2025medgr2breakingdatabarrier,
      title={MedGR$^2$: Breaking the Data Barrier for Medical Reasoning via Generative Reward Learning}, 
      author={Weihai Zhi and Jiayan Guo and Shangyang Li},
      year={2025},
      eprint={2508.20549},
      archivePrefix={arXiv},
      primaryClass={cs.LG},
      url={https://arxiv.org/abs/2508.20549}, 
}

@misc{zhou2025automatingexpertlevelmedicalreasoning,
      title={Automating Expert-Level Medical Reasoning Evaluation of Large Language Models}, 
      author={Shuang Zhou and Wenya Xie and Jiaxi Li and Zaifu Zhan and Meijia Song and Han Yang and Cheyenna Espinoza and Lindsay Welton and Xinnie Mai and Yanwei Jin and Zidu Xu and Yuen-Hei Chung and Yiyun Xing and Meng-Han Tsai and Emma Schaffer and Yucheng Shi and Ninghao Liu and Zirui Liu and Rui Zhang},
      year={2025},
      eprint={2507.07988},
      archivePrefix={arXiv},
      primaryClass={cs.CL},
      url={https://arxiv.org/abs/2507.07988}, 
}

\appendix

\section{Related Work}
\label{app:related_work}

\subsection{Reasoning in Medical Multimodal Large Language Models}
\label{sec:medical_mllms}
Multimodal large language models~\cite{vteam2025glm45vglm41vthinkingversatilemultimodal, yang2025r4bincentivizinggeneralpurposeautothinking, zhu2025internvl3exploringadvancedtraining} have significantly advanced visual understanding through instruction-based learning, a paradigm widely adapted to medical vision--language tasks~\cite{singhal2022largelanguagemodelsencode, sellergren2025medgemmatechnicalreport}. Early medical MLLMs primarily relied on supervised fine-tuning (SFT) with carefully curated or synthetic medical data~\cite{sun2025chirono1ignitingmultimodallarge, Kim2025, xie2025medtrinity25mlargescalemultimodaldataset}. LLaVA-Med~\cite{li2023llavamedtraininglargelanguageandvision} fine-tuned LLaVA~\cite{liu2023visualinstructiontuning} on PubMed Central data for biomedical image understanding. 

Following the success of DeepSeek-R1~\cite{Guo_2025} in enhancing reasoning through GRPO~\cite{shao2024deepseekmathpushinglimitsmathematical}, subsequent work has increasingly adapted RL to the medical domain~\cite{zhang2025pathor1multimodalreinforcementlearningbased, zhang2025medtvtr1multimodalllmempowering}.
Med-R1~\cite{lai2025medr1reinforcementlearninggeneralizable}, MedVLM-R1~\cite{pan2025medvlmr1incentivizingmedicalreasoning}, and GMAI-VL-R1~\cite{su2025gmaivlr1harnessingreinforcementlearning} use GRPO-based training to enhance medical multimodal reasoning. However, most of these approaches compute advantages at the sequence level~\cite{liu2025breakingrewardcollapseadaptive, su2025gmaivlr1harnessingreinforcementlearning}, applying a uniform learning signal across all tokens that cannot distinguish which reasoning step led to an incorrect answer.

\begin{table*}[t]
\centering
\small
\setlength{\tabcolsep}{5pt}
\renewcommand{\arraystretch}{1.08}

% ---------- Subtable (a) ----------
\begin{subtable}[t]{0.47\textwidth}
\centering
\begin{tabular}{lccc}
\toprule
\multirow{2}{*}{\textbf{Dataset}} &
\multirow{2}{*}{\textbf{N}} &
\multicolumn{2}{c}{\textbf{Human--LLM Alignment}} \\
\cmidrule(lr){3-4}
& & \textbf{Cohen's $\kappa$} & \textbf{Agreement Rate} \\
\midrule
VQA-RAD  & 200 & 0.732 & 87.4\% \\
SLAKE    & 200 & 0.725 & 86.2\% \\
PathVQA  & 200 & 0.629 & 85.3\% \\
\midrule
\textbf{Total} & 600 & 0.713 & 86.4\% \\
\bottomrule
\end{tabular}
\vspace{0.1cm}
\caption{Answer correctness alignment.}
\label{tab:human-eval-answer}
\end{subtable}
\hfill
% ---------- Subtable (b) ----------
\begin{subtable}[t]{0.50\textwidth}
\centering
\begin{tabular}{lccc}
\toprule
\multirow{2}{*}{\textbf{Criteria}} &
\multirow{2}{*}{\textbf{N}} &
\multicolumn{2}{c}{\textbf{Human--LLM Alignment}} \\
\cmidrule(lr){3-4}
& & \textbf{Cohen's $\kappa$} & \textbf{Agreement Rate} \\
\midrule
Gold Alignment       & 355 & 0.714 & 85.2\% \\
Answer Contribution  & 355 & 0.668 & 82.9\% \\
\midrule
\textbf{\makecell[l]{Reasoning Process\\Reward}} & 355 & 0.709 & 84.8\% \\
\bottomrule
\end{tabular}
\vspace{0.1cm}
\caption{Step-wise reasoning quality alignment.}
\label{tab:human-eval-reasoning}
\end{subtable}

\vspace{0.1cm}
\caption{
\textbf{Human--LLM evaluator alignment.}
We report alignment between human judgments and LLM-based evaluation for answer correctness and step-wise reasoning quality.
For answer correctness, the overall Cohen's $\kappa$ is 0.713 across 600 samples.
For step-wise reasoning quality, both criteria achieve $\kappa > 0.66$, and the reasoning process reward achieves $\kappa = 0.709$, indicating substantial agreement with human judgments.
}
\label{tab:human-eval-alignment}
\vspace{-0.3cm}
\end{table*}

\subsection{Process Supervision for Medical Reasoning}
\label{sec:process_supervision}
To address this, a growing body of work has moved toward evaluating individual reasoning steps in the medical domain, employing the resulting signal in various ways.
Some approaches integrate step-level process rewards directly into RL training. ChestX-Reasoner~\cite{fan2025chestxreasoneradvancingradiologyfoundation} mines step-by-step reasoning from clinical reports to guide a two-stage SFT-then-RL pipeline. MedGR$^2$~\cite{zhi2025medgr2breakingdatabarrier} trains a generative reward model whose composite reward, combining reasoning quality and answer correctness, supervises GRPO.
Others convert step-level assessments into offline preferences, as in Med-REFL~\cite{yang2026medreflmedicalreasoningenhancement}, which distills tree-of-thoughts reflection values into the policy through direct preference optimization (DPO).
Still others train a process reward model applied at inference time, as in Med-PRM~\cite{yun2025medprmmedicalreasoningmodels}, which verifies each step against retrieved clinical guidelines to select traces.

The importance of step-level errors is thus widely recognized, and prior work~\cite{yun2025medprmmedicalreasoningmodels, zhou2025automatingexpertlevelmedicalreasoning} has emphasized that identifying and correcting errors at specific reasoning steps is essential for reliable clinical decision making.
However, none of these methods redistributes the learning signal according to where a failure occurs. In all these cases the resulting signal is summed into the reward and applied at the sequence level~\cite{fan2025chestxreasoneradvancingradiologyfoundation, zhi2025medgr2breakingdatabarrier}, or distilled into trajectory-level preferences~\cite{yang2026medreflmedicalreasoningenhancement}, and thus does not directly correct the step at which the failure occurs during training.

\subsection{Step-wise Credit Assignment}
\label{sec:credit_assign}
The limitation above, that step-level signals are not redistributed according to where a failure occurs, has been studied more directly in general-domain reasoning. There, recent work explores token-level credit assignment~\cite{tan2025gtpogrpostokensequencelevel, parthasarathi2025grpolambdacreditassignmentimproves, xie2025capoenhancingllmreasoning}
to allocate learning signals selectively across the reasoning trajectory.
For example, FSPO~\cite{li2025reasoningmodelshallucinatemore} uses step-wise factual verification to reward supported steps and penalize hallucinated ones. CAPO~\cite{xie2025capoenhancingllmreasoning} leverages an LLM to generate step-wise critiques for token-level rewards, though both apply penalties in a position-agnostic manner regardless of where in the trajectory the failure occurs. MRPO combines this token-level credit assignment with medical step-level evaluation. By assigning exponentially stronger penalties to earlier invalid steps, MRPO corrects faulty reasoning more effectively than prior medical process supervision and steers the model toward valid reasoning.

\section{Reliability of LLM-as-Judge Evaluation}
\label{app:llm_for_judge}

\subsection{Human–LLM Evaluator Alignment}
\label{app:human_eval}

To validate the reliability of our LLM-as-judge evaluation for \textbf{answer correctness} (Section~\ref{sec:prelim} and \ref{sec:experimental_setup}) and \textbf{step-wise reasoning quality} (Sections~\ref{sec:prelim}, \ref{sec:reward_structure}, and \ref{sec:reasoning_quality_analysis}), we conduct a comparison study with a single medical-student human evaluator.

We randomly sample 200 instances from the test set of each of VQA-RAD, SLAKE, and PathVQA, and generate responses using Qwen2.5-VL-7B-Instruct. Both the LLM and human evaluators independently assess these responses. Answer correctness is evaluated using the prompt in Appendix~\ref{app:appendix_prompts_answer} and step-wise reasoning quality using the prompt in Appendix~\ref{app:appendix_process_reward_prompt}, all with GPT-5-mini as the LLM evaluator. The human evaluator is provided with the same criteria specified in each prompt.

\paragraph{Metrics.}
We measure inter-rater agreement between the two evaluators using \textbf{agreement rate} and \textbf{Cohen's $\kappa$}. Agreement rate represents the proportion of samples on which both evaluators made identical judgments. However, this metric can be misleading under class imbalance, as high agreement may occur by chance. To address this, we additionally report Cohen's $\kappa$, which adjusts for chance agreement. Following the guidelines of \citet{LandisKoch1977}, a Cohen's $\kappa$ above 0.61 indicates substantial agreement.

\paragraph{Answer Correctness.}

Table~\ref{tab:human-eval-answer} reports the alignment between the LLM evaluator and the human evaluator on the 600 selected instances, broken down by benchmark. While the agreement rate exceeds 85\% across all three benchmarks, we additionally report Cohen's $\kappa$ to account for class imbalance. For VQA-RAD and SLAKE, Cohen's $\kappa$ exceeds 0.7, indicating substantial alignment. For PathVQA, Cohen's $\kappa$ is lower due to more severe class imbalance, where the proportion of incorrect answers is higher. Under such imbalance, Cohen's $\kappa$ can be reduced even when agreement rate remains high. Additionally, PathVQA involves longer answers and higher question difficulty, making correctness judgments more challenging and resulting in lower inter-evaluator agreement.
Finally, when aggregating results across all three benchmarks, we obtain Cohen’s $\kappa = 0.713$, indicating strong alignment between the human evaluator and the LLM evaluator.

\paragraph{Step-wise Reasoning Quality.}

For step-wise reasoning quality evaluation, we first segment each generated rationale into sentence-level steps and then score each step based on two metrics: Gold Alignment and Answer Contribution, which are derived from the process reward criteria introduced in Section~\ref{sec:prelim}, \ref{sec:reward_structure}, and Section~\ref{sec:reasoning_quality_analysis}.
Since each instance contains an average of 3.9 reasoning steps, evaluating all steps for the full set of 600 instances would be prohibitively costly. We therefore randomly sample 100 instances from the 600 and evaluate all reasoning steps within these instances, resulting in a total of 355 reasoning steps.

Table~\ref{tab:human-eval-reasoning} reports the alignment between the LLM evaluator and the human evaluator on the 355 reasoning steps. Agreement rates exceed 80\% for both metrics, and Cohen's $\kappa$ exceeds 0.66 for both Gold Alignment and Answer Contribution. We further measure alignment on the \textbf{reasoning process reward} derived from these two metrics. As described in Section~\ref{sec:reward_structure}, this reward is computed binarily, assigning 1 when either metric receives a score of 1 and 0 otherwise. Both evaluators independently score the two metrics and compute the corresponding process reward. We obtain an agreement rate of 84.8\% and Cohen's $\kappa = 0.709$, indicating substantial agreement between the two evaluators.

\begin{table}[t]
\centering
\small
\setlength{\tabcolsep}{6pt}
\renewcommand{\arraystretch}{1.12}

\begin{tabular}{lccc}
\toprule
\multirow{2}{*}{\textbf{Method}} &
\multicolumn{3}{c}{\textbf{Judge Model}} \\
\cmidrule(lr){2-4}
& \textbf{GPT-5-mini} & \textbf{GPT-5.4} & \textbf{Claude-4.5-haiku} \\
\midrule

\multicolumn{4}{@{}l}{\textit{Qwen2.5-VL-7B-Instruct}} \\
GRPO & 26.06 & 24.37 & 24.96 \\
MRPO & 26.79 & 25.01 & 26.17 \\

\midrule
\multicolumn{4}{@{}l}{\textit{Qwen3-VL-8B-Instruct}} \\
GRPO & 28.69 & 26.91 & 28.05 \\
MRPO & 29.09 & 27.57 & 28.73 \\

\midrule
\multicolumn{4}{@{}l}{\textit{Qwen3-VL-8B-Thinking}} \\
GRPO & 29.74 & 28.19 & 29.52 \\
MRPO & 30.48 & 28.49 & 30.76 \\

\midrule
\multicolumn{4}{@{}l}{\textit{InternVL3-8B-Instruct}} \\
GRPO & 30.84 & 29.32 & 28.28 \\
MRPO & 31.94 & 30.69 & 29.15 \\

\bottomrule
\end{tabular}

\vspace{0.1cm}
\caption{
\textbf{Performance comparison under different judge models.}
Average accuracy of GRPO and MRPO across four backbones, evaluated by 
GPT-5-mini, GPT-5.4, and Claude-4.5-haiku.
}
\label{tab:judge-accuracy}
\vspace{-0.3cm}
\end{table}

\subsection{Cross-Judge Evaluation}
\label{app:judge_difference}

Since MRPO employs GPT-5-mini both as the process reward judge during training and as the evaluator for answer correctness, a concern arises as to whether the observed gains in both answer accuracy and reasoning quality stem from evaluator-aligned overfitting to a single judge rather than genuine reasoning improvement. To address this, we repeat the evaluations from Section~\ref{sec:main_results} and Section~\ref{sec:reasoning_quality_analysis}, replacing the judge with GPT-5.4 and Claude-4.5-haiku to re-assess answer accuracy and reasoning step evaluation across all four backbones and benchmarks.

\begin{table}[t]
\centering
\small
\setlength{\tabcolsep}{6pt}
\renewcommand{\arraystretch}{1.12}

\begin{tabular}{lccc}
\toprule
\multirow{2}{*}{\textbf{Method}} &
\multicolumn{3}{c}{\textbf{Judge Model}} \\
\cmidrule(lr){2-4}
& \textbf{GPT-5-mini} & \textbf{GPT-5.4} & \textbf{Claude-4.5-haiku} \\
\midrule

\multicolumn{4}{@{}l}{\textit{Early-Stage}} \\
Baseline & 58.6 & 52.3 & 54.8 \\
GRPO     & 21.6 & 24.1 & 23.3 \\
MRPO     & 13.4 & 19.4 & 16.7 \\

\midrule
\multicolumn{4}{@{}l}{\textit{Mid-Stage}} \\
Baseline & 29.1 & 33.5 & 30.4 \\
GRPO     & 34.1 & 36.6 & 35.9 \\
MRPO     & 40.1 & 37.7 & 37.0 \\

\midrule
\multicolumn{4}{@{}l}{\textit{Late-Stage}} \\
Baseline & 12.3  & 14.2 & 14.8 \\
GRPO     & 44.3 & 39.3 & 40.8 \\
MRPO     & 46.5 & 42.9 & 46.3 \\

\bottomrule
\end{tabular}

\vspace{0.1cm}
\caption{
\textbf{Sample distribution across First Failure Point (FFP) stages under different judge models.}
Proportions of early, mid, and late-stage failures for the baseline, GRPO, and MRPO under different judge models.
}
\label{tab:judge-failure-stage}
\vspace{-0.3cm}
\end{table}

\paragraph{Answer Accuracy.}
Table~\ref{tab:judge-accuracy} reports answer accuracy under the three judges. Using each alternative judge, we re-evaluate answer accuracy on all benchmarks from Section~\ref{sec:main_results}, and report the average score across benchmarks in the table. While the absolute scores vary across judges, reflecting differences in their scoring strictness, the relative ordering between GRPO and MRPO is preserved in every case. Across all four backbones and all three judges, MRPO consistently outperforms GRPO without a single exception. This consistency indicates that the advantage of MRPO is not an artifact of the specific judge used during training, but rather a substantive improvement in answer quality that holds independently of the evaluating judge.

\paragraph{First Failure Point (FFP) Analysis.}
Under each judge, we compute the proportion of samples across FFP stages (Early, Mid, and Late-Stage), following Section~\ref{sec:reasoning_quality_analysis}.

Table~\ref{tab:judge-failure-stage} reports the resulting proportion of samples falling into each FFP stage for the baseline, GRPO, and MRPO under each judge. Although the exact magnitudes differ across judges, the same monotonic trend emerges consistently: moving from the baseline to GRPO and then to MRPO progressively reduces the share of early-stage failures while shifting failures toward later stages. Under all three judges, MRPO records the lowest early-stage failure proportion and the highest late-stage proportion, confirming that its effect of delaying the first failure point is preserved regardless of the judge. Taken together, these results demonstrate that MRPO's mitigation of cascading failures is a robust property of the method rather than a judge-specific phenomenon.

\section{Experimental Setup}
\label{app:exp_setup}

\subsection{Dataset}
\label{app:dataset}

\paragraph{Training Dataset.}
Our training set is derived from the training splits of medical VQA benchmarks, VQA-RAD~\cite{vqarad2018}, SLAKE~\cite{liu2021slakesemanticallylabeledknowledgeenhanceddataset}, and PathVQA~\cite{he2020pathvqa30000questionsmedical}. 

\begin{itemize}
  \item \textbf{VQA-RAD.} VQA-RAD is the first manually constructed radiology VQA dataset, containing 315 images from the MedPix database paired with 2,247 clinician-generated QA pairs. The images cover head CT and MRI, chest X-ray, and abdominal CT, with both open-ended and closed-ended questions.

  \item \textbf{SLAKE.} SLAKE is a bilingual English-Chinese medical VQA dataset comprising 642 radiology images including CT, MRI, and X-ray, paired with 14,028 question-answer pairs annotated by experienced physicians. It covers five body regions. We use only the English subset in our experiments.

  \item \textbf{PathVQA.} PathVQA is the first pathology VQA dataset, containing 4,289 pathology images and 32,632 question-answer pairs across eight categories. The dataset was constructed using a semi-automated pipeline from pathology textbooks and digital libraries, with all pairs manually verified. The majority of questions are open-ended.
\end{itemize}

From these sources, we exclude binary and multiple-choice questions to focus on open-ended instances, where answers are provided as free-form text. Gold reasoning annotations are obtained from MedThink~\cite{gai2024medthinkexplainingmedicalvisual}, a rationale-augmented resource built on the same benchmarks. To ensure every training example is paired with a gold reasoning annotation, we align our training instances to MedThink by exact matching on the image, question, and answer triple. This procedure yields a one-to-one mapping between the final training set and the gold reasoning annotations. After alignment, the training set contains 13{,}381 open-ended QA instances over 3{,}556 unique images.

\paragraph{Evaluation Dataset.}
For evaluation, we use two types of benchmarks: \textbf{in-distribution benchmarks}, consisting of the test splits of our training datasets, and five \textbf{out-of-distribution benchmarks}. For the in-distribution benchmarks, we filter the test splits of the training datasets, VQA-RAD~\cite{vqarad2018}, SLAKE~\cite{liu2021slakesemanticallylabeledknowledgeenhanceddataset}, and PathVQA~\cite{he2020pathvqa30000questionsmedical}, to open-ended QA only, yielding 200, 706, and 3{,}357 samples respectively, for a total of 4{,}263 samples. For the out-of-distribution evaluation, we additionally adopt five benchmarks, PMC-VQA~\cite{zhang2024pmcvqavisualinstructiontuning}, VQA-Med-2021~\cite{ImageCLEF-VQA-Med2021}, Quilt-VQA~\cite{seyfioglu2025quiltllavavisualinstructiontuning}, RadImageNet-VQA~\cite{butsanets2026radimagenetvqalargescalectmri}, and MIMIC-Ext-MIMIC-CXR-VQA~\cite{PhysioNet-mimic-ext-mimic-cxr-vqa-1.0.0}, that span diverse imaging modalities unseen during training.

\begin{itemize}
\item \textbf{PMC-VQA.} PMC-VQA is a large-scale medical VQA dataset comprising 227K question--answer pairs over 149K images, automatically constructed from figure--caption pairs in PubMed Central articles via a scalable pipeline. The dataset spans diverse imaging modalities and diseases, and although it provides multiple-choice options for every question, we evaluate in an open-ended setting in our experiments. Specifically, we use the 2{,}000-sample manually-verified clean test set.
\item \textbf{VQA-Med-2021.} VQA-Med-2021 is a radiology VQA benchmark released as part of the ImageCLEF 2021 challenge, with a pronounced focus on questions about abnormalities in radiology images. Its test set consists of 500 radiology images paired with 500 abnormality questions, and the reference answers of the test set were manually validated by a medical doctor to ensure data quality. Among these, we use only the 425 QA pairs that have a single ground-truth answer.
\item \textbf{Quilt-VQA.} Quilt-VQA is a histopathology VQA benchmark comprising 985 images with 1{,}283 human-generated question--answer pairs, extracted from educational histopathology videos on YouTube. The benchmark covers a wide range of diagnostic concepts and supports both open-ended and closed-ended questions for evaluating pathology-focused multimodal models. As Quilt-VQA is released as an evaluation set, in our experiments we use only the 724 open-ended QA pairs.
\item \textbf{RadImageNet-VQA.} RadImageNet-VQA is a large-scale CT and MRI VQA dataset built from expert annotations, providing 750K images paired with 7.5M question--answer samples. It covers three diagnostic tasks, abnormality detection, anatomy recognition, and pathology identification across 8 anatomical regions and 97 pathology categories, and supports open-ended, closed-ended, and multiple-choice questions. For evaluation, the authors provide a stratified benchmark of 1{,}000 images with 9{,}000 QA pairs. From this benchmark, we use only the 2{,}000 open-ended questions in our experiments.
\item \textbf{MIMIC-Ext-MIMIC-CXR-VQA.} MIMIC-Ext-MIMIC-CXR-VQA is a complex and large-scale chest radiograph VQA dataset comprising approximately 377K question-answer pairs derived from the MIMIC-CXR database. The dataset is constructed using 48 question templates that involve set and logical operations, designed to evaluate diverse aspects of chest X-ray interpretation. From the test set, we filter to open-ended questions of the \textit{query} semantic type, and we further restrict these to the 1{,}724 test samples that have a single ground-truth answer.
\end{itemize}

\subsection{Implementation Detail}
\label{app:implementation_detail}
All experiments are conducted with PyTorch~\cite{paszke2019pytorchimperativestylehighperformance} on 8$\times$NVIDIA A100 GPUs, and we adopt FlashAttention-2~\cite{dao2023flashattention2fasterattentionbetter} to improve training efficiency. Our implementation builds on VLM-R1~\cite{shen2025vlmr1stablegeneralizabler1style}, an open-source GRPO framework for VLMs. 

We train MRPO on four backbones, Qwen2.5-VL-7B-Instruct, Qwen3-VL-8B-Instruct, Qwen3-VL-8B-Thinking and InternVL3-8B-Instruct. To quantify MRPO's effectiveness and compare it against other methods, all RL methods including MRPO, GRPO, and GDPO are trained on the same training dataset under identical settings. Each method is trained for 1 epoch with a batch size of 64 and a learning rate of $10^{-6}$, sampling 8 rollouts per prompt with a single run per configuration. Following the standard GRPO configuration~\cite{shao2024deepseekmathpushinglimitsmathematical}, we set the KL coefficient $\beta = 0.04$ and the clipping range $\epsilon = 0.2$ for all RL methods.

For SFT, we train on the same training dataset augmented with gold reasoning annotations, so the model jointly learns to produce the reasoning trace and the final answer. We employ LoRA~\cite{hu2021loralowrankadaptationlarge} with rank 8, alpha 32, and dropout 0.05, with a learning rate of $2 \times 10^{-5}$ for 3 epochs.

\begin{table}[t]
\centering
\small
\setlength{\tabcolsep}{5pt}
\renewcommand{\arraystretch}{1.15}

\resizebox{\columnwidth}{!}{%
\begin{tabular}{lccc}
\toprule
\textbf{Metric} & \textbf{GRPO} & \textbf{GDPO} & \textbf{MRPO} \\
\midrule
Training Time  & 110h 25min & 113h 39min & 120h 54min \\
Input Tokens   & 234.3M     & 245.0M     & 273.3M \\
Output Tokens  & 73.0M      & 72.8M      & 78.7M \\
\arrayrulecolor{gray!35}
\midrule
\arrayrulecolor{black}
Total Cost     & \$192.96      & \$201.74      & \$215.48 \\
\bottomrule
\end{tabular}
}

\vspace{0.1cm}
\caption{
\textbf{Training resource comparison across RL methods.}
We compare training time, token usage, and total cost for GRPO, GDPO, and MRPO.
}
\label{tab:training-resource-comparison}
\vspace{-0.3cm}
\end{table}

\subsection{Training Cost and Efficiency}
\label{app:api_cost}

We compare of training time, token usage, and 
total cost across RL methods on Qwen2.5-VL-7B-Instruct in 
Table~\ref{tab:training-resource-comparison}. MRPO, GRPO, and GDPO all 
issue only a single API call per rollout to jointly evaluate all 
reasoning sentences. With 13K samples and 8 rollouts, this amounts to 
roughly 107K calls per epoch across all configurations.

As shown in Table~\ref{tab:training-resource-comparison}, the total cost ranges from \$192.96 (GRPO) to \$215.48 (MRPO), with GDPO in between. MRPO incurs only a marginal increase over GRPO ($\approx$12\%), which arises not from additional judge queries but from MRPO generating longer, more structured reasoning traces that consume more tokens. This indicates that step-wise process supervision is practically affordable at this scale. For larger-scale training, the results in Appendix~\ref{app:local_prm} show that while MRPO maintains consistent gains over GRPO even with open-source judges, their absolute performance still falls short of API-based judges, as the quality of step-level evaluation governs the magnitude of RL gains. This motivates the development of dedicated local process reward models that match frontier-judge quality. We leave this as future work, which would enable both stronger RL-driven gains and the complete elimination of API dependency.

\begin{table*}[t]
\centering
\fontsize{8.8pt}{10pt}\selectfont
\setlength{\tabcolsep}{2.7pt}
\renewcommand{\arraystretch}{1.05}

\resizebox{\textwidth}{!}{%
\begin{tabular}{lcccccccccc}
\toprule
\multirow{2}{*}{\textbf{Method}} &
\multicolumn{4}{c}{\textbf{In-Distribution}} &
\multicolumn{6}{c}{\textbf{Out-of-Distribution}} \\
\cmidrule(lr){2-5}\cmidrule(lr){6-11}
& {\scriptsize \textbf{VQA-RAD}}
& {\scriptsize \textbf{SLAKE}}
& {\scriptsize \textbf{PathVQA}}
& {\scriptsize \textbf{AVG}}
& {\scriptsize \textbf{PMC-VQA}}
& {\scriptsize \textbf{VQA-Med}}
& {\scriptsize \textbf{Quilt-VQA}}
& {\scriptsize \textbf{Rad-VQA}}
& {\scriptsize \textbf{MIMIC-VQA}}
& {\scriptsize \textbf{AVG}} \\

\midrule

\multicolumn{11}{@{}l}{\textit{Qwen2.5-VL-7B-Instruct}} \\
Base
& 42.00 & 58.29 & 17.10 & 25.10
& 28.35 & 5.18 & 19.06 & 27.20 & 15.08 & 22.28 \\
SFT
& 38.50 & 63.88 & 18.83 & 26.30
& 28.50 & 4.71 & 18.37 & 26.20 & 15.02 & 21.91 \\
GRPO
& \textbf{44.00} & 65.27 & 20.52 & 29.06
& \underline{31.05} & 3.76 & 20.99 & \underline{29.30} & \underline{16.71} & \underline{24.20} \\
\rowcolor{cyan!12}
MRPO
& 41.50 & 65.89 & 21.30 & 29.63
& 30.85 & 5.65 & \textbf{22.79} & \textbf{29.65} & \textbf{18.65} & \textbf{25.04} \\
SFT + GRPO
& 42.00 & \underline{66.82} & \underline{22.34} & \underline{29.67}
& 30.40 & \textbf{6.82} & 21.13 & 26.65 & 13.81 & 22.71 \\
SFT + MRPO
& \underline{42.50} & \textbf{68.68} & \textbf{23.44} & \textbf{30.85}
& \textbf{31.15} & \underline{6.59} & \underline{22.24} & 26.90 & 14.79 & 23.35 \\

\midrule
\multicolumn{11}{@{}l}{\textit{Qwen3-VL-8B-Instruct}} \\
Base
& \underline{43.00} & 59.69 & 21.48 & 28.81
& \underline{31.00} & 9.41 & \underline{23.90} & 33.15 & 15.31 & 25.61 \\
SFT
& \textbf{45.00} & 66.29 & 21.27 & 29.84
& 30.35 & 7.76 & 19.48 & 33.40 & 15.20 & 24.89 \\
GRPO
& \textbf{45.00} & 67.14 & 20.11 & 29.06
& 30.25 & \underline{9.65} & \textbf{24.03} & \textbf{40.60} & \textbf{18.79} & \underline{28.46} \\
\rowcolor{cyan!12}
MRPO
& 41.50 & \underline{68.27} & 20.43 & 29.35
& \textbf{33.00} & \textbf{12.00} & \underline{23.90} & \underline{40.20} & \underline{17.46} & \textbf{28.94} \\
SFT + GRPO
& 40.00 & 67.42 & \underline{25.95} & \underline{33.47}
& 28.25 & 6.59 & 19.75 & 35.20 & 15.43 & 24.82 \\
SFT + MRPO
& 40.50 & \textbf{68.41} & \textbf{26.57} & \textbf{34.15}
& 29.20 & 7.53 & 19.48 & 35.75 & 14.50 & 25.05 \\

\bottomrule
\end{tabular}
}

\vspace{0.1cm}
\caption{
\textbf{Ablation on SFT cold-start initialization.}
We compare six configurations across two backbones, Qwen2.5-VL-7B-Instruct and Qwen3-VL-8B-Instruct, including the base model, SFT, GRPO, MRPO, SFT followed by GRPO, and SFT followed by MRPO.
VQA-Med denotes VQA-Med-2021, Rad-VQA denotes RadImageNet-VQA, and MIMIC-VQA denotes MIMIC-Ext-MIMIC-CXR-VQA.}
\label{tab:sft_cold_start}
\vspace{-0.3cm}
\end{table*}

\section{Ablation Study}
\label{app:ablation_study}

To better understand the contribution of each design choice in MRPO, we conduct four ablation studies on two backbones, Qwen2.5-VL-7B-Instruct and Qwen3-VL-8B-Instruct. Specifically, we examine the SFT cold-start initialization prior to RL training, the token-level shaping function for advantage reshaping, the advantage reweighting strategy, and the choice of process reward model. All ablations are conducted on the same training dataset and evaluated on three in-distribution and five out-of-distribution medical VQA benchmarks.

\subsection{SFT Cold-Start Initialization}
\label{app:sft_cold_start}
Table~\ref{tab:sft_cold_start} reports the effect of initializing the model with supervised fine-tuning prior to RL training. We compare six configurations against the base model, including SFT alone, GRPO alone, MRPO alone, SFT followed by GRPO, and SFT followed by MRPO. The cold-start SFT is trained under the same setting as SFT alone, with training details provided in Appendix~\ref{app:implementation_detail}.

A consistent pattern emerges across both backbones. SFT cold-start substantially improves in-distribution performance but degrades out-of-distribution performance. For Qwen3-VL-8B-Instruct, combining SFT with MRPO further improves PathVQA from 20.43 to 26.57 and SLAKE from 68.27 to 68.41, achieving the highest in-distribution average among all configurations. However, the same configuration drops the out-of-distribution average from 28.94 to 25.05, falling below MRPO without SFT cold-start. The same pattern holds for Qwen2.5-VL-7B-Instruct, where SFT+MRPO yields gains on PathVQA from 21.30 to 23.44 and SLAKE from 65.89 to 68.68 in the in-distribution setting, but underperforms MRPO alone on out-of-distribution benchmarks.

This trade-off suggests that SFT on gold reasoning annotations overfits the model to the specific reasoning patterns of the training distribution, which improves performance on the matched in-distribution test sets but weakens the transferable reasoning capability that RL alone is able to induce. The effect is particularly pronounced on PathVQA, where SFT cold-start brings the largest gains, since the gold reasoning annotations from MedThink most directly align with the PathVQA test distribution. In contrast, the out-of-distribution benchmarks such as PMC-VQA, VQA-Med, and Quilt-VQA show consistent degradation, indicating that the reasoning patterns learned through SFT do not generalize to unseen imaging modalities and question styles.
Based on this finding, we adopt direct RL training without SFT cold-start in our main experiments, prioritizing the generalization capability that better reflects the demands of real-world clinical deployment, where models routinely encounter queries beyond the training distribution.

\begin{table*}[t]
\centering
\fontsize{8.8pt}{10pt}\selectfont
\setlength{\tabcolsep}{3.0pt}
\renewcommand{\arraystretch}{1.05}

\resizebox{\textwidth}{!}{%
\begin{tabular}{lccccccccc}
\toprule
\multirow{2}{*}{\textbf{Shaping Function}} &
\multicolumn{3}{c}{\textbf{In-Distribution}} &
\multicolumn{5}{c}{\textbf{Out-of-Distribution}} &
\multirow{2}{*}{\textbf{AVG}} \\
\cmidrule(lr){2-4}\cmidrule(lr){5-9}
& {\scriptsize \textbf{VQA-RAD}}
& {\scriptsize \textbf{SLAKE}}
& {\scriptsize \textbf{PathVQA}}
& {\scriptsize \textbf{PMC-VQA}}
& {\scriptsize \textbf{VQA-Med}}
& {\scriptsize \textbf{Quilt-VQA}}
& {\scriptsize \textbf{Rad-VQA}}
& {\scriptsize \textbf{MIMIC-VQA}}
& \\

\midrule

\multicolumn{10}{@{}l}{\textit{Qwen2.5-VL-7B-Instruct}} \\
Constant (CAPO)
& 40.50 & 63.31 & 18.59
& 29.75 & 4.47 & \underline{20.99}
& 27.90 & 15.78
& 24.68 \\
Uniform (FSPO)
& \textbf{43.00} & 61.47 & 18.17
& 29.20 & 4.47 & 20.72
& 27.10 & 15.43
& 24.16 \\
Linear
& \underline{42.50} & 62.46 & \underline{19.66}
& \underline{29.95} & 5.41 & 19.89
& \underline{28.70} & 16.13
& \underline{25.18} \\
Quadratic
& 40.50 & \underline{64.45} & 18.59
& 29.25 & \textbf{5.88} & 20.17
& 28.35 & \underline{17.81}
& 25.05 \\
\rowcolor{cyan!12}
Exponential (MRPO)
& 41.50 & \textbf{65.89} & \textbf{21.30}
& \textbf{30.85} & \underline{5.65} & \textbf{22.79}
& \textbf{29.65} & \textbf{18.65} & \textbf{26.79} \\

\midrule
\multicolumn{10}{@{}l}{\textit{Qwen3-VL-8B-Instruct}} \\
Constant (CAPO)
& \underline{41.00} & 62.46 & 18.92
& 30.20 & 6.12 & \underline{22.51}
& 38.70 & 16.65
& 27.05 \\
Uniform (FSPO)
& 36.00 & 61.33 & 19.24
& 28.60 & \underline{8.00} & 20.17
& 33.35 & 13.63
& 25.19 \\
Linear
& \textbf{41.50} & \underline{62.75} & \underline{19.72}
& 30.30 & 7.53 & 21.55
& \textbf{40.55} & \underline{16.94}
& \underline{27.70} \\
Quadratic
& 40.50 & 62.61 & 19.48
& \underline{31.50} & 7.76 & 22.10
& 39.80 & 14.50
& 27.35 \\
\rowcolor{cyan!12}
Exponential (MRPO)
& \textbf{41.50} & \textbf{68.27} & \textbf{20.43}
& \textbf{33.00} & \textbf{12.00} & \textbf{23.90}
& \underline{40.20} & \textbf{17.46} & \textbf{29.09} \\

\bottomrule
\end{tabular}
}

\vspace{0.1cm}
\caption{
\textbf{Ablation study on token shaping strategies.}
We compare five token-level shaping functions, namely constant(CAPO), uniform(FSPO), linear, quadratic, and exponential, on two backbones, Qwen2.5-VL-7B-Instruct and Qwen3-VL-8B-Instruct.
VQA-Med denotes VQA-Med-2021, Rad-VQA denotes RadImageNet-VQA, and MIMIC-VQA denotes MIMIC-Ext-MIMIC-CXR-VQA.
The exponential function corresponds to the proposed shaping strategy in MRPO.
}
\label{tab:token-shaping-ablation}
\vspace{-0.3cm}
\end{table*}

\begin{table*}[t]
\centering
\fontsize{8.8pt}{10pt}\selectfont
\setlength{\tabcolsep}{3.0pt}
\renewcommand{\arraystretch}{1.05}

\resizebox{\textwidth}{!}{%
\begin{tabular}{lccccccccc}
\toprule
\multirow{2}{*}{\textbf{Reweighting Strategy}} &
\multicolumn{3}{c}{\textbf{In-Distribution}} &
\multicolumn{5}{c}{\textbf{Out-of-Distribution}} &
\multirow{2}{*}{\textbf{AVG}} \\
\cmidrule(lr){2-4}\cmidrule(lr){5-9}
& {\scriptsize \textbf{VQA-RAD}}
& {\scriptsize \textbf{SLAKE}}
& {\scriptsize \textbf{PathVQA}}
& {\scriptsize \textbf{PMC-VQA}}
& {\scriptsize \textbf{VQA-Med}}
& {\scriptsize \textbf{Quilt-VQA}}
& {\scriptsize \textbf{Rad-VQA}}
& {\scriptsize \textbf{MIMIC-VQA}}
& \\

\midrule

\multicolumn{10}{@{}l}{\textit{Qwen2.5-VL-7B-Instruct}} \\
Soft Reweighting
& \textbf{45.00} & \textbf{66.71} & 19.63
& 28.50 & \textbf{6.35} & \underline{21.13}
& 28.55 & 14.10
& 25.00 \\
Full Reweighting
& 35.50 & 64.02 & \underline{20.67}
& \underline{29.65} & 4.71 & 19.20
& \textbf{30.40} & \underline{15.55}
& \underline{25.55} \\
\rowcolor{cyan!12}
Selective Reweighting (MRPO)
& \underline{41.50} & \underline{65.89} & \textbf{21.30}
& \textbf{30.85} & \underline{5.65} & \textbf{22.79}
& \underline{29.65} & \textbf{18.65} & \textbf{26.79} \\

\midrule
\multicolumn{10}{@{}l}{\textit{Qwen3-VL-8B-Instruct}} \\
Soft Reweighting
& \textbf{45.00} & \underline{64.45} & \textbf{20.88}
& \underline{30.20} & 6.59 & 22.38
& 36.95 & \underline{14.62}
& 27.29 \\
Full Reweighting
& \underline{41.50} & \underline{64.45} & \underline{20.52}
& 30.15 & \underline{8.00} & \textbf{25.83}
& \underline{38.80} & 11.95
& \underline{27.34} \\
\rowcolor{cyan!12}
Selective Reweighting (MRPO)
& \underline{41.50} & \textbf{68.27} & 20.43
& \textbf{33.00} & \textbf{12.00} & \underline{23.90}
& \textbf{40.20} & \textbf{17.46} & \textbf{29.09} \\

\bottomrule
\end{tabular}
}

\vspace{0.1cm}
\caption{
\textbf{Ablation study on advantage reweighting strategies.}
We compare three strategies, namely Soft Reweighting, Full Reweighting, and Selective Reweighting, on two backbones, Qwen2.5-VL-7B-Instruct and Qwen3-VL-8B-Instruct.
VQA-Med denotes VQA-Med-2021, Rad-VQA denotes RadImageNet-VQA, and MIMIC-VQA denotes MIMIC-Ext-MIMIC-CXR-VQA.
The selective reweighting strategy corresponds to the proposed method in MRPO, which applies advantage reshaping only to instances with incorrect final answers.
}
\label{tab:answer-reweighting-ablation}
\vspace{-0.3cm}
\end{table*}

\subsection{Token-Level Shaping Function}
\label{app:token_shaping}

Table~\ref{tab:token-shaping-ablation} reports the effect of different token-level shaping functions in MRPO's advantage reshaping.
We compare five credit assignment schemes for penalizing tokens in failed reasoning steps: constant, uniform, linear, quadratic, and exponential.
Notably, the first two are not merely synthetic shaping variants but
correspond to existing credit assignment methods proposed in the general domain: constant reproduces CAPO~\cite{xie2025capoenhancingllmreasoning} and uniform reproduces FSPO~\cite{li2025reasoningmodelshallucinatemore}, which we adapt to our setting as representative prior baselines.
For a failed reasoning step at relative position $\frac{k-1}{K-1}$, where $k$ denotes the step index and $K$ denotes the total number of steps, the penalty multipliers are defined as $-1$ for uniform (FSPO), $-\left(1-\frac{k-1}{K-1}\right)$ for linear, $-\left(1-\frac{k-1}{K-1}\right)^2$ for quadratic, and $-\exp\!\left(1-\frac{k-1}{K-1}\right)$ for exponential.
Each multiplier is applied to $|A_i|$.
For constant (CAPO), a fixed penalty $-W_{\mathrm{process}}$ is instead subtracted from the token rewards of failed steps prior to normalization, independent of both the position $\frac{k-1}{K-1}$ and the advantage magnitude $|A_i|$.

Uniform (FSPO) applies the same penalty regardless of position, and constant (CAPO) subtracts a fixed penalty from failed-step token rewards, while the other three apply progressively stronger penalties to tokens in earlier failed steps, with exponential providing the steepest decay. 
A clear pattern emerges across both backbones. All position-aware shaping functions consistently outperform the position-agnostic methods (FSPO, CAPO), confirming that penalizing early failures more strongly than later ones is beneficial. Among them, exponential shaping yields the largest improvement, gaining 2.63 and 3.90 points on Qwen2.5-VL-7B-Instruct and Qwen3-VL-8B-Instruct over the uniform baseline, and achieves the highest average score on both backbones, reaching 26.79 and 29.09 respectively. CAPO additionally shows reward hacking, collapsing its average step count to 1.4 to evade the position-independent penalty.
Based on this finding, we adopt exponential shaping as the default token-level function in MRPO, as it provides the strongest signal for correcting the root causes of reasoning failures. This also positions our contribution relative to existing step-level credit assignment methods: reflecting where a failure occurs, rather than only whether a step is valid, is what drives the gains.

\begin{table*}[t]
\centering
\fontsize{8.8pt}{10pt}\selectfont
\setlength{\tabcolsep}{3.0pt}
\renewcommand{\arraystretch}{1.05}

\resizebox{\textwidth}{!}{%
\begin{tabular}{llccccccccc}
\toprule
\multirow{2}{*}{\textbf{RL method}} &
\multirow{2}{*}{\textbf{PRM}} &
\multicolumn{3}{c}{\textbf{In-Distribution}} &
\multicolumn{5}{c}{\textbf{Out-of-Distribution}} &
\multirow{2}{*}{\textbf{AVG}} \\
\cmidrule(lr){3-5}\cmidrule(lr){6-10}
&
& {\scriptsize \textbf{VQA-RAD}}
& {\scriptsize \textbf{SLAKE}}
& {\scriptsize \textbf{PathVQA}}
& {\scriptsize \textbf{PMC-VQA}}
& {\scriptsize \textbf{VQA-Med}}
& {\scriptsize \textbf{Quilt-VQA}}
& {\scriptsize \textbf{Rad-VQA}}
& {\scriptsize \textbf{MIMIC-VQA}}
& \\

\midrule

\multicolumn{11}{@{}l}{\textit{Qwen2.5-VL-7B-Instruct}} \\
- (Base Model) & -
& \underline{42.00} & 58.29 & 17.10
& 28.35 & 5.18 & 19.06
& 27.20 & 15.08
& 23.36 \\

\arrayrulecolor{gray!35}
\midrule
\arrayrulecolor{black}

GRPO & Med-PRM
& 40.50 & 65.01 & 17.27
& 29.80 & 6.12 & 16.99
& 26.15 & 14.21
& 23.64 \\
GRPO & MedGemma-27B
& 41.00 & 64.73 & 18.41
& 30.30 & \underline{6.35} & 20.30
& 28.85 & 17.34
& 25.26 \\
GRPO & Qwen3-VL-30B
& 39.00 & 62.18 & 18.50
& \underline{31.20} & 3.29 & 19.06
& \underline{29.95} & 15.66
& 24.99 \\
\rowcolor{gray!10}
GRPO & GPT-5-mini
& \textbf{44.00} & 65.27 & \underline{20.52}
& 31.05 & 3.76 & \underline{20.99}
& 29.30 & 16.71 & \underline{26.06} \\

\arrayrulecolor{gray!35}
\midrule
\arrayrulecolor{black}

MRPO & Med-PRM
& 39.50 & 62.32 & 16.98
& 28.60 & \underline{6.35} & 18.23
& 25.85 & 14.04
& 23.16 \\
MRPO & MedGemma-27B
& \underline{42.00} & \textbf{66.29} & 19.06
& 30.60 & 4.94 & \underline{20.99}
& 28.50 & 16.94
& 25.49 \\
MRPO & Qwen3-VL-30B
& 38.50 & 61.05 & 18.98
& \textbf{31.55} & \textbf{6.82} & 18.23
& \textbf{30.95} & \underline{18.04}
& 25.75 \\
\rowcolor{cyan!12}
MRPO & GPT-5-mini
& 41.50 & \underline{65.89} & \textbf{21.30}
& 30.85 & 5.65 & \textbf{22.79}
& 29.65 & \textbf{18.65} & \textbf{26.79} \\

\midrule
\multicolumn{11}{@{}l}{\textit{Qwen3-VL-8B-Instruct}} \\
- (Base Model) & -
& 43.00 & 59.69 & \textbf{21.48}
& 31.00 & 9.41 & \underline{23.90}
& 33.15 & 15.31
& 26.83 \\

\arrayrulecolor{gray!35}
\midrule
\arrayrulecolor{black}

GRPO & Med-PRM
& 43.00 & 64.87 & 12.98
& 27.25 & 7.53 & 15.19
& 32.00 & 18.62
& 23.59 \\
GRPO & MedGemma-27B
& \textbf{45.00} & 65.86 & 19.42
& 31.20 & \underline{11.76} & 21.69
& 38.45 & \underline{18.97}
& 28.14 \\
GRPO & Qwen3-VL-30B
& 43.50 & 64.72 & 20.34
& 30.65 & 8.94 & 23.62
& 37.20 & 17.63
& 27.81 \\
\rowcolor{gray!10}
GRPO & GPT-5-mini
& \textbf{45.00} & \underline{67.14} & 20.11
& 30.25 & 9.65 & \textbf{24.03}
& \textbf{40.60} & 18.79 & \underline{28.69} \\

\arrayrulecolor{gray!35}
\midrule
\arrayrulecolor{black}

MRPO & Med-PRM
& 40.00 & 66.29 & 13.40
& 27.00 & 5.18 & 16.99
& 33.25 & \textbf{21.00}
& 24.34 \\
MRPO & MedGemma-27B
& \underline{44.00} & 66.71 & 19.54
& 31.05 & \underline{11.76} & 22.38
& 38.65 & 18.91
& 28.26 \\
MRPO & Qwen3-VL-30B
& 43.50 & 65.30 & 19.48
& \underline{32.15} & \underline{11.76} & \underline{23.90}
& 37.00 & 18.39
& 28.06 \\
\rowcolor{cyan!12}
MRPO & GPT-5-mini
& 41.50 & \textbf{68.27} & \underline{20.43}
& \textbf{33.00} & \textbf{12.00} & \underline{23.90}
& \underline{40.20} & 17.46 & \textbf{29.09} \\

\bottomrule
\end{tabular}
}

\vspace{0.1cm}
\caption{
\textbf{Ablation study on process reward models.}
We compare four process reward models, Med-PRM, MedGemma-27B, Qwen3-VL-30B-A3B-Instruct(denoted as Qwen3-VL-30B), and GPT-5-mini, under both GRPO and MRPO training on two backbones, Qwen2.5-VL-7B-Instruct and Qwen3-VL-8B-Instruct.
VQA-Med denotes VQA-Med-2021, Rad-VQA denotes RadImageNet-VQA, and MIMIC-VQA denotes MIMIC-Ext-MIMIC-CXR-VQA.
GPT-5-mini is the default process reward model in MRPO.
}
\label{tab:local-prm-ablation}
\vspace{-0.3cm}
\end{table*}

\subsection{Advantage Reweighting Strategy}
\label{app:reweigthing}

Table~\ref{tab:answer-reweighting-ablation} reports the effect of different advantage reweighting strategies. We compare three strategies, namely Soft Reweighting, Full Reweighting, and the proposed selective strategy in MRPO. Full Reweighting applies step-wise advantage reshaping to all training instances regardless of answer correctness, Soft Reweighting applies the same reshaping scaled by a factor of $0.5$, and MRPO applies the reshaping only when the answer reward falls below the threshold $\tau$, leaving the advantages of correctly answered trajectories unchanged.

On both backbones, MRPO consistently outperforms both Full and Soft Reweighting. On Qwen3-VL-8B-Instruct, MRPO achieves an average of 29.09, surpassing Full Reweighting at 27.34 and Soft Reweighting at 27.29 by 1.75 and 1.80 points respectively. The same pattern holds for Qwen2.5-VL-7B-Instruct, where MRPO reaches 26.79 compared to 25.55 and 25.00 for the two alternatives. These results indicate that applying advantage reshaping indiscriminately to all training instances harms performance by disrupting reasoning trajectories that already lead to correct answers. Restricting reshaping to failed predictions allows the model to retain the reasoning patterns that already work while concentrating the learning signal on the root causes of failure. Based on this finding, we adopt selective reweighting in MRPO, applying advantage reshaping only to instances where the final answer is judged incorrect.

\subsection{Process Reward Model}
\label{app:local_prm}
Table~\ref{tab:local-prm-ablation} reports the effect of different process reward models (PRMs) used to compute the step-wise reasoning reward $R_{\mathrm{proc}}$.
We compare four alternatives under both GRPO and MRPO training.
MedGemma-27B is a strong open-source medical multimodal LLM repurposed as a local judge, Med-PRM is a process reward model for medical reasoning evaluation that takes only the question and reasoning text as input without the image, Qwen3-VL-30B-A3B-Instruct(denoted as Qwen3-VL-30B) is a strong open-source generalist multimodal LLM, and GPT-5-mini is a frontier general-purpose LLM accessed via API.

\begin{figure*}[t]
  \centering

  \begin{subfigure}[t]{0.24\textwidth}
    \centering
    \includegraphics[width=\linewidth]{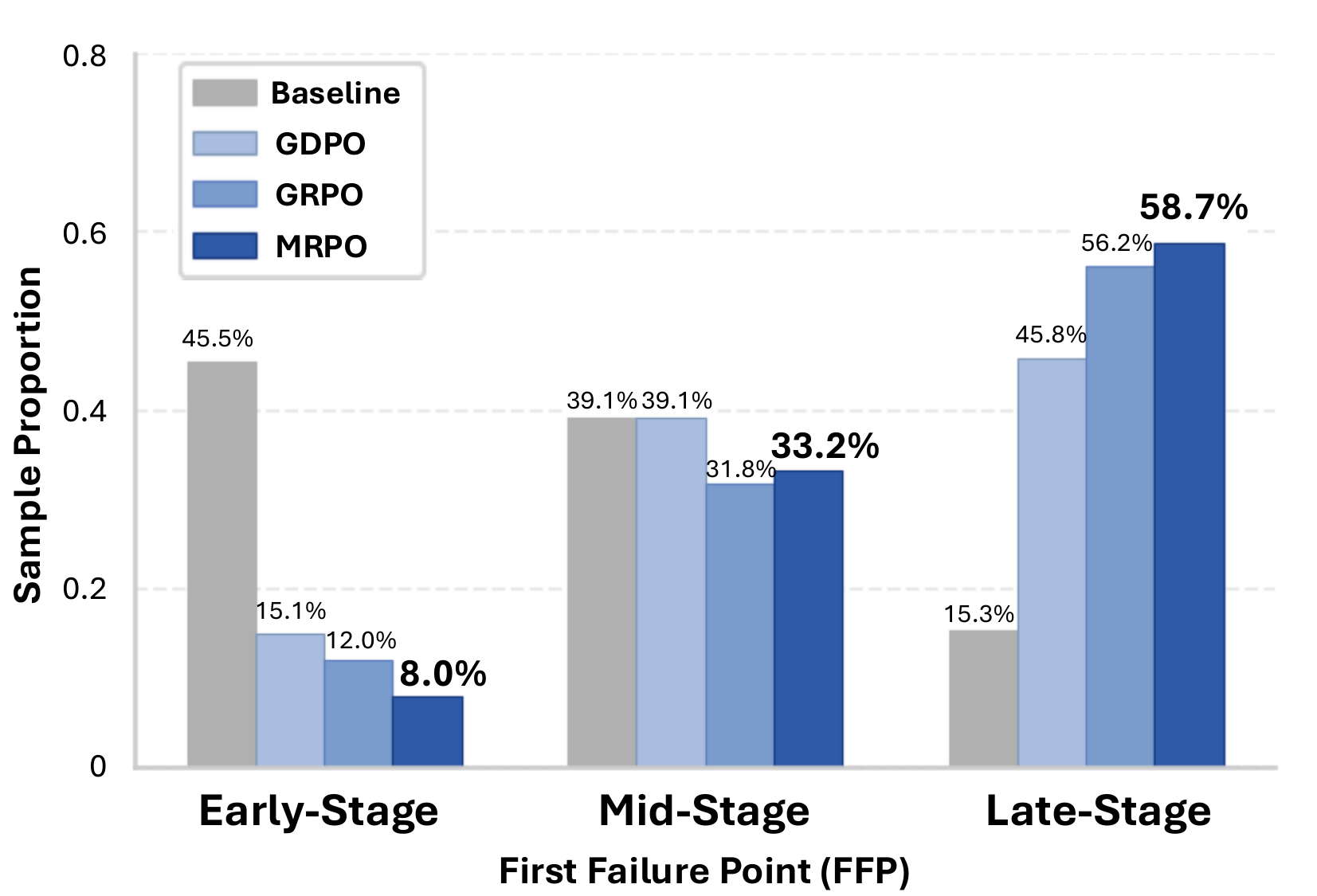}
    \caption{Qwen2.5-VL-7B-Instruct}
    \label{fig:qwen25-vl-7b-instruct}
  \end{subfigure}
  \hfill
  \begin{subfigure}[t]{0.24\textwidth}
    \centering
    \includegraphics[width=\linewidth]{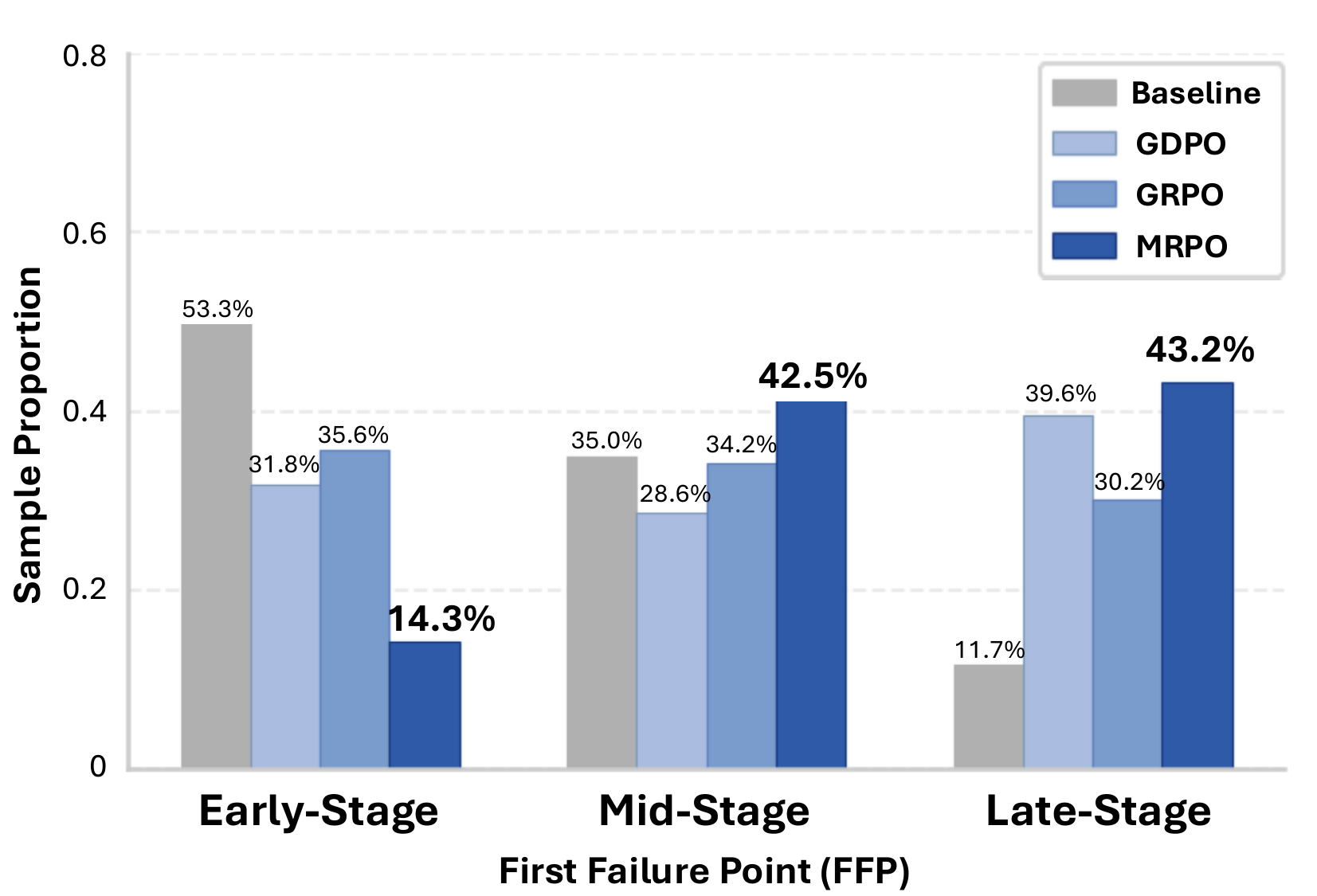}
    \caption{Qwen3-VL-8B-Instruct}
    \label{fig:qwen3-vl-8b-instruct}
  \end{subfigure}
  \hfill
  \begin{subfigure}[t]{0.24\textwidth}
    \centering
    \includegraphics[width=\linewidth]{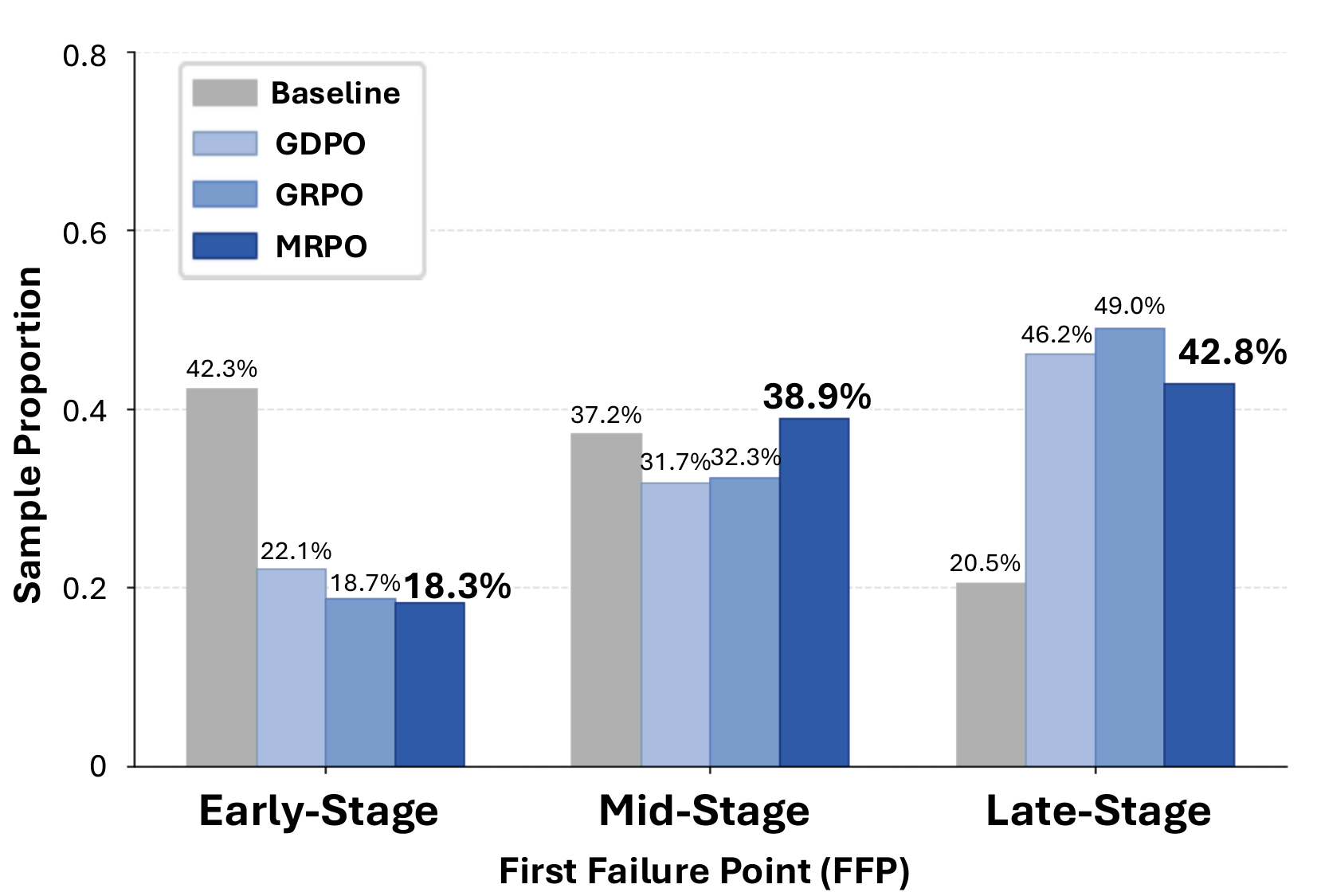}
    \caption{Qwen3-VL-8B-Thinking}
    \label{fig:qwen3-VL-8B-thinking}
  \end{subfigure}
  \hfill
  \begin{subfigure}[t]{0.24\textwidth}
    \centering
    \includegraphics[width=\linewidth]{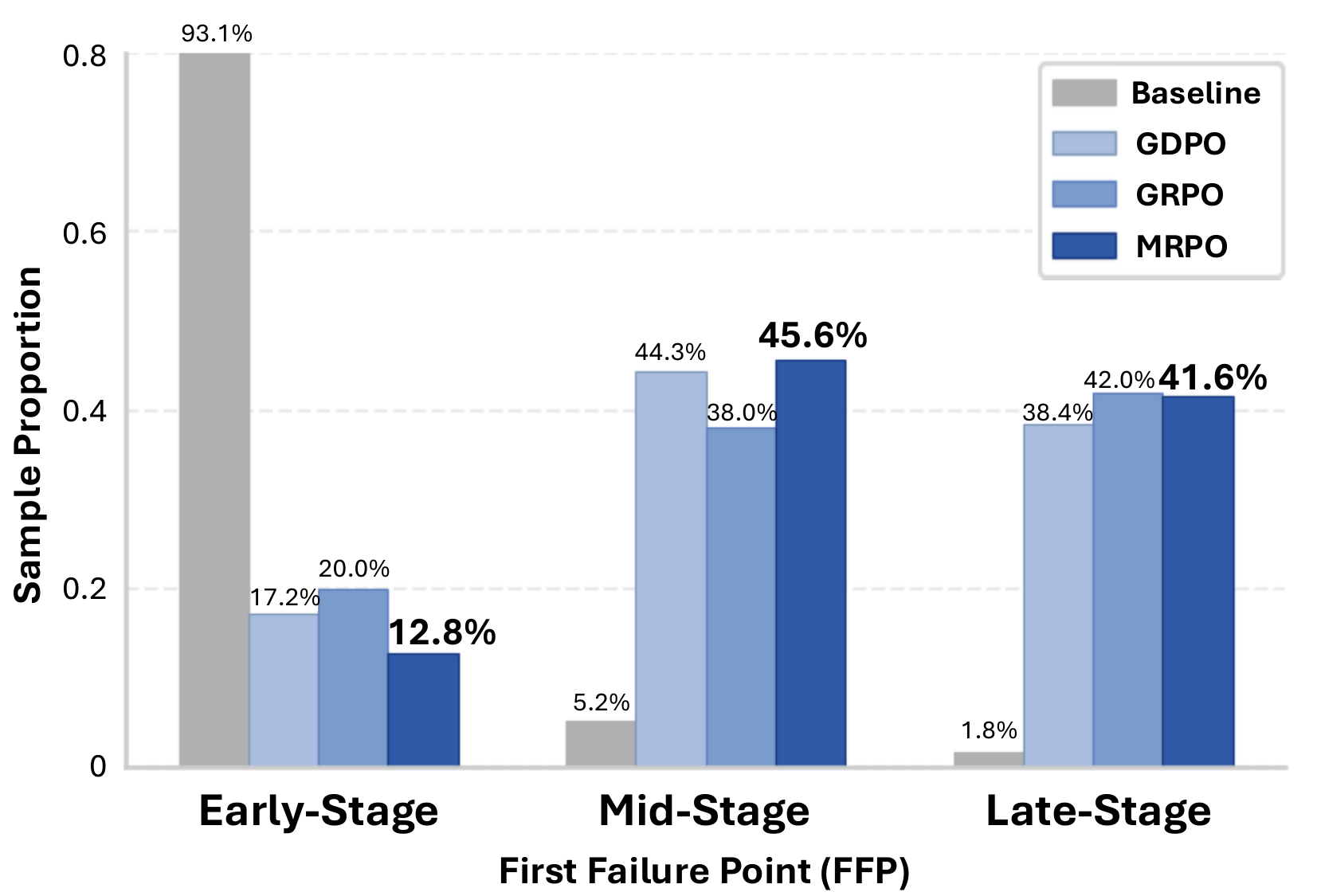}
    \caption{InternVL3-8B-Instruct}
    \label{fig:internvl3-8b-instruct}
  \end{subfigure}

  \caption{
    \textbf{Sample distribution across First Failure Point (FFP) stages 
    for each backbone.}
    Proportions of samples falling into Early, Mid, and Late-Stage FFP ranges for the baseline, GRPO, GDPO, and MRPO on Qwen2.5-VL-7B-Instruct, Qwen3-VL-8B-Instruct, Qwen3-VL-8B-Thinking, and InternVL3-8B-Instruct.
  }
  \label{fig:four-backbone-comparison-ffp}
  \vspace{-0.3cm}
\end{figure*}

\begin{figure*}[t]
  \centering

  \begin{subfigure}[t]{0.24\textwidth}
    \centering
    \includegraphics[width=\linewidth]{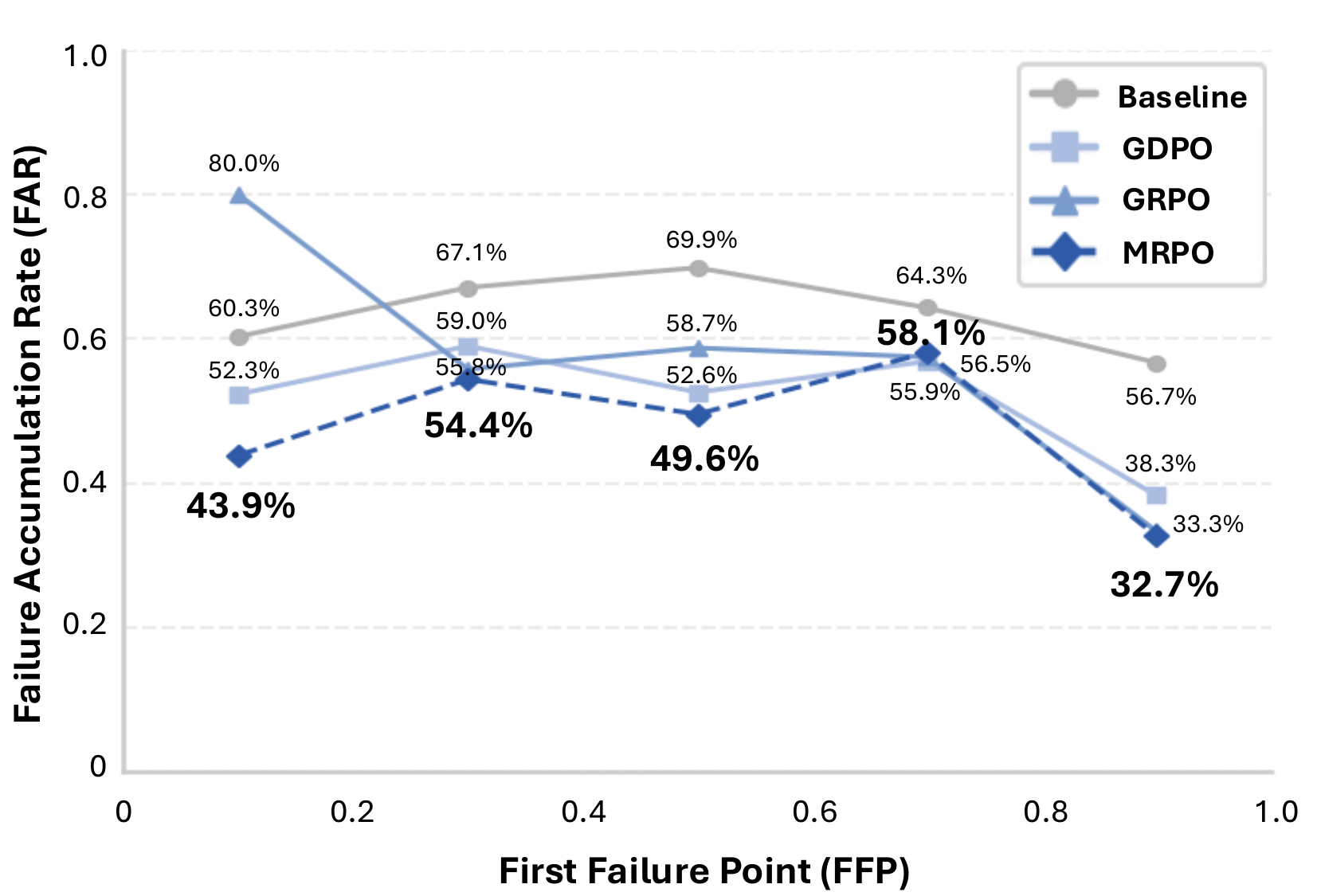}
    \caption{Qwen2.5-VL-7B-Instruct}
    \label{fig:qwen25-vl-7b-instruct}
  \end{subfigure}
  \hfill
  \begin{subfigure}[t]{0.24\textwidth}
    \centering
    \includegraphics[width=\linewidth]{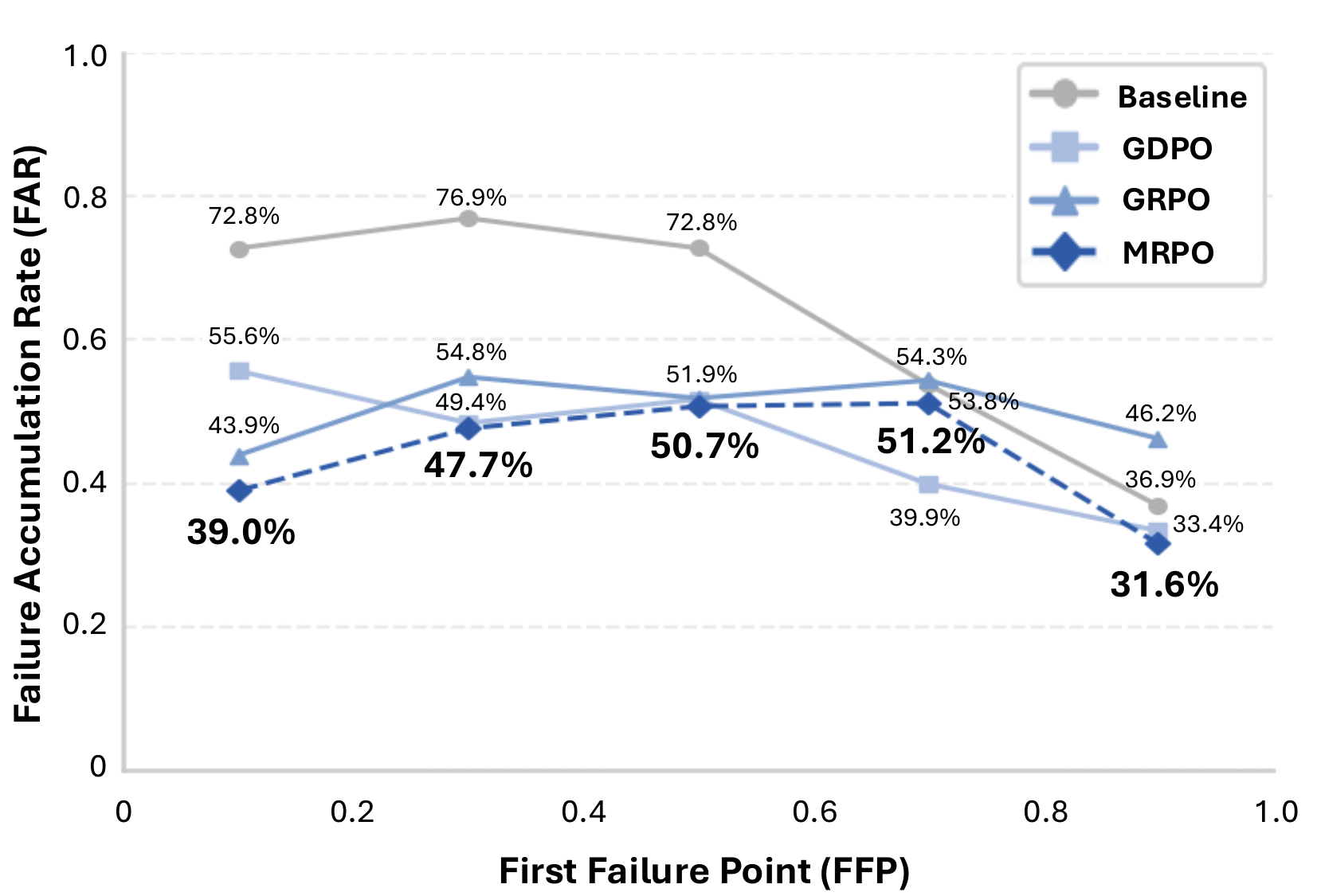}
    \caption{Qwen3-VL-8B-Instruct}
    \label{fig:qwen3-vl-8b-instruct}
  \end{subfigure}
  \hfill
  \begin{subfigure}[t]{0.24\textwidth}
    \centering
    \includegraphics[width=\linewidth]{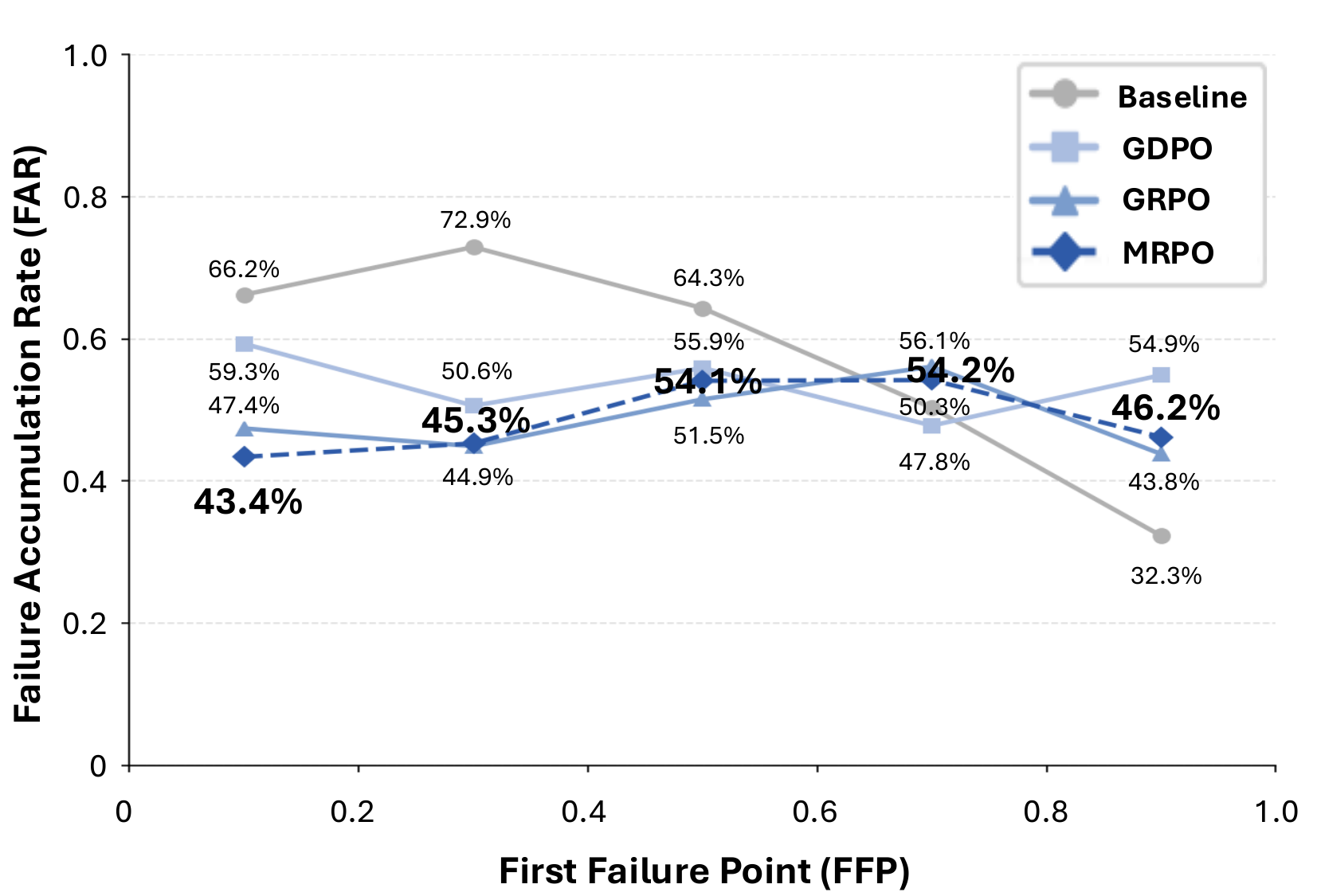}
    \caption{Qwen3-VL-8B-Thinking}
    \label{fig:qwen3-VL-8B-thinking}
  \end{subfigure}
  \hfill
  \begin{subfigure}[t]{0.24\textwidth}
    \centering
    \includegraphics[width=\linewidth]{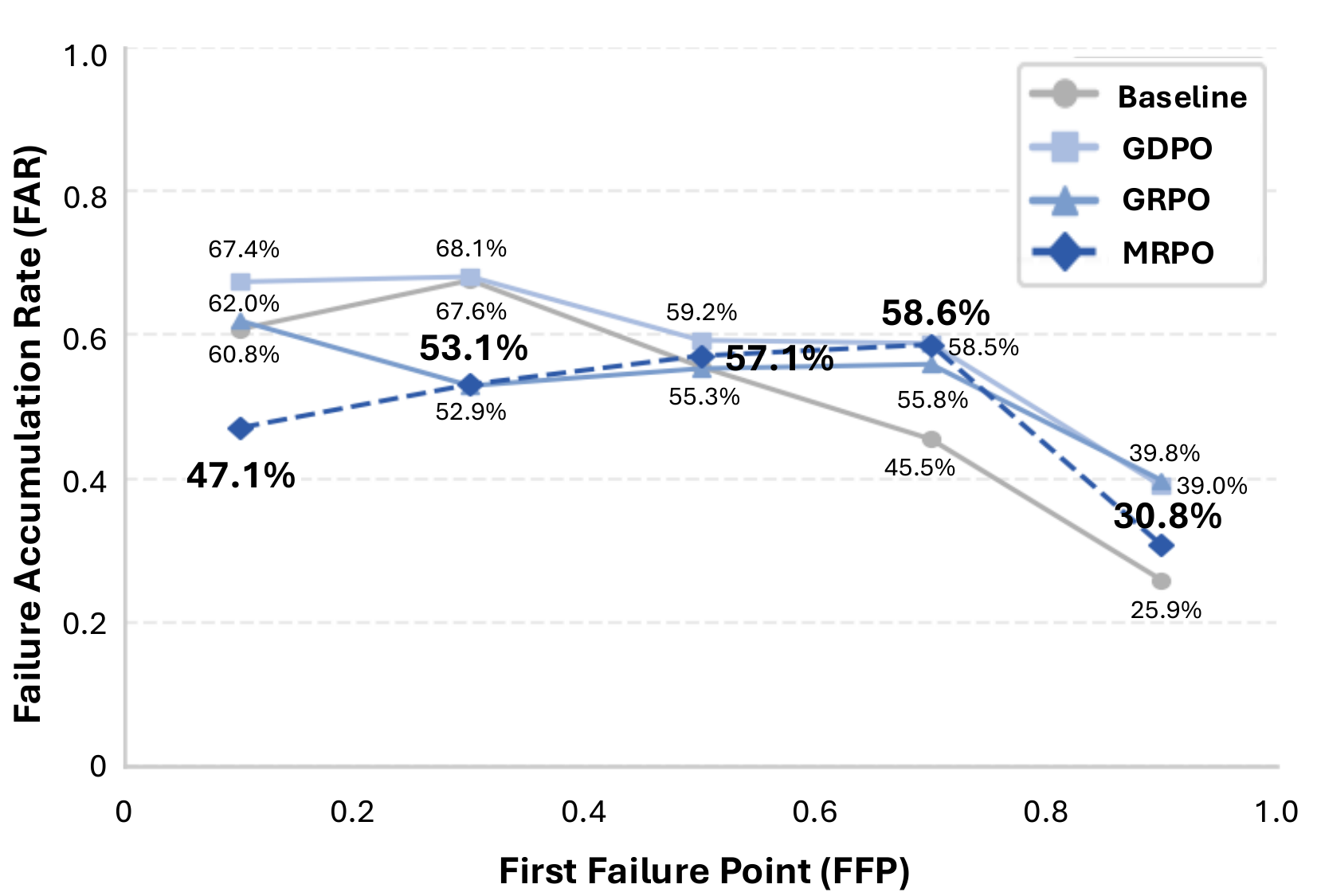}
    \caption{InternVL3-8B-Instruct}
    \label{fig:internvl3-8b-instruct}
  \end{subfigure}

  \caption{
    \textbf{Failure Accumulation Rate (FAR) across FFP bins for each backbone.}
    FAR across First Failure Point (FFP) bins for the baseline, GRPO, GDPO, and MRPO on Qwen2.5-VL-7B-Instruct, Qwen3-VL-8B-Instruct, Qwen3-VL-8B-Thinking, and InternVL3-8B-Instruct.
  }
  \label{fig:four-backbone-comparison-far}
  \vspace{-0.3cm}
\end{figure*}

Two clear patterns emerge from the results.
First, MRPO outperforms GRPO in most configurations, indicating that step-wise advantage reshaping benefits across a wide range of process reward models.
On Qwen3-VL-8B-Instruct, MRPO outperforms GRPO with all four PRMs, with gains of 0.75, 0.12, 0.25, and 0.40 points with Med-PRM, MedGemma-27B, Qwen3-VL-30B, and GPT-5-mini, respectively.
The same trend holds on Qwen2.5-VL-7B-Instruct with MedGemma-27B, Qwen3-VL-30B, and GPT-5-mini, while only with the weakest judge, Med-PRM, does MRPO underperform GRPO by a small margin.
This suggests that MRPO's core benefit, the step-wise reshaping mechanism, is largely robust to the choice of judge model, as long as it provides sufficiently reliable step-level signals.

Second, GPT-5-mini achieves the highest average across both backbones and training paradigms, which is why we adopt it as the default process reward model in MRPO: it provides the highest performance ceiling, reaching the best average on every backbone.
On Qwen2.5-VL-7B-Instruct, MRPO with GPT-5-mini reaches 26.79, surpassing Qwen3-VL-30B at 25.75, MedGemma-27B at 25.49, and Med-PRM at 23.16.
On Qwen3-VL-8B-Instruct, MRPO with GPT-5-mini reaches 29.09, surpassing MedGemma-27B at 28.26, Qwen3-VL-30B at 28.06, and Med-PRM at 24.34.
We attribute this to step-level evaluation quality. GPT-5-mini is a stronger evaluator of reasoning steps, achieving substantial agreement with human judgments (Cohen's $\kappa$ above 0.7), as shown in our human-LLM alignment study in Appendix~\ref{app:human_eval}. In contrast, MedGemma-27B is a weaker judge, and Med-PRM receives only the question and reasoning text without the image, so it cannot properly assess visual reasoning and consequently fails to identify which reasoning steps are invalid. Since the effectiveness of MRPO's step-wise advantage reshaping fundamentally depends on accurately identifying which reasoning steps are invalid, a judge with stronger reasoning-evaluation capability translates directly into stronger downstream RL performance.

Based on these findings, we adopt GPT-5-mini as the default process reward model in MRPO.
Notably, both the medically fine-tuned MedGemma-27B and the strong open-source generalist Qwen3-VL-30B narrow the gap to GPT-5-mini, with MedGemma slightly ahead on Qwen3-VL-8B-Instruct and Qwen3-VL-30B on Qwen2.5-VL-7B-Instruct. This indicates that MRPO's benefit does not depend on a proprietary judge, and that either a capable medical judge or a strong generalist offers a practical open-source alternative when API costs are a concern. We leave developing dedicated medical VQA process reward models that match frontier-LLM judgment quality without API dependency as a promising direction for future work.

\section{Reasoning Analysis}
\label{app:reasoning_analysis}

\subsection{Reasoning Analysis Across Backbones}
\label{app:reasoning_analysis_across_backbones}

To examine the results of Section~\ref{sec:reasoning_quality_analysis} in greater detail, we decompose the First Failure Point (FFP) analysis by backbone. As shown in Figure~\ref{fig:four-backbone-comparison-ffp}, across all four backbones, the most consistent finding is the suppression of early-stage, cascade-inducing failures. The baseline concentrates failures in the early stage, and while all RL methods reduce this, MRPO lowers it the most on every backbone, reaching 8.0\% on Qwen2.5-VL-7B-Instruct, 14.3\% on Qwen3-VL-8B-Instruct, 18.3\% on Qwen3-VL-8B-Thinking, and 12.8\% on InternVL3-8B-Instruct, the lowest among all methods and below both GRPO and GDPO in each case. The removed early failures are correspondingly redistributed to later stages, most prominently as a clean early-to-late shift on Qwen2.5-VL-7B-Instruct, where the late stage rises from 15.3\% to 58.7\%. Overall, the directional pattern of fewer early-stage failures and more late-stage failures holds consistently across four backbones from different model families, demonstrating that MRPO's distinctive contribution, the suppression of early cascade failures, generalizes across architectures.

Figure~\ref{fig:four-backbone-comparison-far} reports the Failure Accumulation Rate (FAR) across FFP bins for each backbone, where MRPO attains the lowest FAR in the earliest bin, FFP 0.0–0.2, on all four backbones, reaching 43.9\%, 39.0\%, 43.4\%, and 47.1\% for Qwen2.5-VL-7B-Instruct, Qwen3-VL-8B-Instruct, Qwen3-VL-8B-Thinking, and InternVL3-8B-Instruct, below every competing method and well under the baseline. In other words, even when an early failure does occur, MRPO is the most effective at preventing it from cascading through the remaining steps, indicating improved recovery from early errors. This advantage is concentrated in the early bins, consistent with MRPO's design: because the exponential penalty targets the earliest failed steps, its strongest effect on failure accumulation emerges precisely where the penalty is applied. Together with the suppression of early-stage failures above, this shows that MRPO addresses cascading failures along two complementary and architecture-agnostic axes, delaying the onset of failures and improving recovery once they begin.

\subsection{Paired Comparison of GRPO and MRPO}
\label{app:reasoning_analysis_paired_comparison}

Since MRPO is derived directly from GRPO by reshaping step-wise advantages, we conduct a paired, instance-level comparison between the two methods to isolate the effect of this reshaping on individual predictions. We pool the reasoning traces generated on the test splits of the three in-distribution benchmarks (VQA-RAD, SLAKE, and PathVQA) across four backbones (Qwen2.5-VL-7B-Instruct, Qwen3-VL-8B-Instruct, Qwen3-VL-8B-Thinking, and InternVL3-8B-Instruct), and evaluate each instance under both methods.
Table~\ref{tab:grpo-mrpo-consistency} summarizes the resulting prediction consistency. The two methods agree on the large majority of instances, with 25.2\% jointly correct and 62.0\% jointly incorrect. The discordant cases, where exactly one method succeeds, are of primary interest: MRPO is uniquely correct on 6.8\% of instances while GRPO is uniquely correct on 6.0\%. We analyze these two disjoint groups separately. For the MRPO-only-correct group, we ask how MRPO recovers instances that GRPO fails, by examining where reasoning failures occurred under GRPO and how they are redistributed under MRPO. For the GRPO-only-correct group, we conversely characterize the nature of the failures that MRPO newly introduces. Together, these two analyses reveal not only the net change in accuracy but the underlying shift in reasoning behavior that GRPO and MRPO induce on the same inputs.

\begin{table}[t]
\centering
\small
\setlength{\tabcolsep}{8pt}
\renewcommand{\arraystretch}{1.15}

\begin{tabular}{l!{\color{gray!35}\vrule}c!{\color{gray!35}\vrule}c}
\toprule
\textbf{GRPO \textbackslash{} MRPO} & \textbf{Correct} & \textbf{Wrong} \\
\arrayrulecolor{gray!35}
\midrule
\arrayrulecolor{black}
\textbf{Correct} & 4,294 (25.2\%) & 1,024 (6.0\%) \\
\arrayrulecolor{gray!35}
\midrule
\arrayrulecolor{black}
\textbf{Wrong}   & 1,167 (6.8\%)   & 10,567 (62.0\%) \\
\bottomrule
\end{tabular}

\vspace{0.1cm}
\caption{
\textbf{Prediction consistency between GRPO and MRPO.}
Rows indicate GRPO predictions and columns indicate MRPO predictions.
}
\label{tab:grpo-mrpo-consistency}
\vspace{-0.3cm}
\end{table}

\begin{table}[t]
\centering
\small
\setlength{\tabcolsep}{3.2pt}
\renewcommand{\arraystretch}{1.12}

\resizebox{\columnwidth}{!}{%
\begin{tabular}{
l
!{\color{gray!35}\vrule}c
!{\color{gray!35}\vrule}c
!{\color{gray!35}\vrule}c
!{\color{gray!35}\vrule}c
!{\color{black}\vrule}c
}
\toprule
\diagbox[width=2.4cm,height=0.75cm]{\textbf{GRPO}}{\textbf{MRPO}}
& \textbf{No FFP}
& \textbf{Early}
& \textbf{Mid}
& \textbf{Late}
& \textbf{SUM} \\
\arrayrulecolor{gray!35}
\midrule
\arrayrulecolor{black}
\textbf{No FFP}
& 279 & 11 & 23 & 61 & 374 \\
\arrayrulecolor{gray!35}
\midrule
\arrayrulecolor{black}
\rowcolor{cyan!10}
\textbf{Early}
& \begin{tabular}{@{}c@{}}63\\[-1pt]{\scriptsize (30.6\%)}\end{tabular}
& \begin{tabular}{@{}c@{}}15\\[-1pt]{\scriptsize (7.2\%)}\end{tabular}
& \begin{tabular}{@{}c@{}}36\\[-1pt]{\scriptsize (17.5\%)}\end{tabular}
& \begin{tabular}{@{}c@{}}92\\[-1pt]{\scriptsize (44.7\%)}\end{tabular}
& 206 \\
\arrayrulecolor{gray!35}
\midrule
\arrayrulecolor{black}
\textbf{Mid}
& 120 & 32 & 44 & 47 & 243 \\
\arrayrulecolor{gray!35}
\midrule
\arrayrulecolor{black}
\textbf{Late}
& 184 & 23 & 56 & 81 & 344 \\
\arrayrulecolor{black}
\midrule
\textbf{SUM}
& 646 & 81 & 159 & 281 & 1,167 \\
\bottomrule
\end{tabular}
}

\vspace{0.1cm}
\caption{
    \textbf{Transition matrix of First Failure Point (FFP) stages between GRPO and MRPO for MRPO-only-correct instances.}
    Rows indicate the GRPO FFP stage and columns the MRPO FFP stage. The GRPO Early-stage row is highlighted, with percentages computed within that row.
}
\label{tab:grpo-mrpo-ffp-transition-early}
\vspace{-0.3cm}
\end{table}

To characterize how MRPO recovers instances that GRPO answers incorrectly, we track, for each MRPO-only-correct instances, the First Failure Point (FFP) stage of its reasoning trace under both methods. Table~\ref{tab:grpo-mrpo-ffp-transition-early} reports the resulting transition matrix, where rows index the GRPO FFP stage and columns index the MRPO FFP stage. The clearest signal is the reduction of early and mid-stage failures. Under GRPO, these instances exhibit a substantial concentration of failures in the early and mid stages, consistent with the cascading-failure pattern that drives incorrect predictions. Under MRPO, both are markedly reduced, with early-stage failures falling from 206 to 81 and mid-stage failures from 243 to 159, indicating that step-wise advantage reshaping corrects the early reasoning errors that GRPO fails to resolve. The redistribution is most pronounced for the early-stage GRPO row, highlighted in Table~\ref{tab:grpo-mrpo-ffp-transition-early}: of the 206 instances that fail early under GRPO, only 7.2\% remain in the early stage under MRPO, while the remaining 92.8\% move out of it. Specifically, 30.6\% are resolved without a detectable failure point, 17.5\% shift to the mid stage, and 44.7\% shift to the late stage. This indicates that MRPO rarely leaves an early failure in place; instead, it either eliminates the initial failure or pushes its onset to a later stage, where the residual error is far less likely to cascade. We further note that a portion of the MRPO-only-correct instances corresponds to GRPO traces with no detected failure point, the No-FFP row. These appear to be cases where the reasoning trajectory is essentially valid but the answer is marked incorrect due to surface-level expression mismatches rather than a genuine reasoning error, and thus fall outside the step-level reasoning behavior analyzed here.

\begin{table}[t]
\centering
\small
\setlength{\tabcolsep}{3.2pt}
\renewcommand{\arraystretch}{1.12}

\resizebox{\columnwidth}{!}{%
\begin{tabular}{
l
!{\color{gray!35}\vrule}c
!{\color{gray!35}\vrule}c
!{\color{gray!35}\vrule}c
!{\color{gray!35}\vrule}c
!{\color{black}\vrule}c
}
\toprule
\diagbox[width=2.4cm,height=0.75cm]{\textbf{GRPO}}{\textbf{MRPO}}
& \textbf{No FFP}
& \textbf{Early}
& \textbf{Mid}
& \textbf{Late}
& \textbf{SUM} \\
\arrayrulecolor{gray!35}
\midrule
\arrayrulecolor{black}
\textbf{No FFP}
& 336 & 33 & 88 & 140 & 597 \\
\arrayrulecolor{gray!35}
\midrule
\arrayrulecolor{black}
\textbf{Early}
& 15 & 17 & 8 & 41 & 81 \\
\arrayrulecolor{gray!35}
\midrule
\arrayrulecolor{black}
\textbf{Mid}
& 48 & 24 & 76 & 17 & 165 \\
\arrayrulecolor{gray!35}
\midrule
\arrayrulecolor{black}
\textbf{Late}
& 88 & 15 & 21 & 57 & 181 \\
\arrayrulecolor{black}
\midrule
\rowcolor{cyan!10}
\textbf{SUM}
& \begin{tabular}{@{}c@{}}487\\[-1pt]{\scriptsize (47.6\%)}\end{tabular}
& \begin{tabular}{@{}c@{}}89\\[-1pt]{\scriptsize (8.7\%)}\end{tabular}
& \begin{tabular}{@{}c@{}}193\\[-1pt]{\scriptsize (18.8\%)}\end{tabular}
& \begin{tabular}{@{}c@{}}255\\[-1pt]{\scriptsize (24.9\%)}\end{tabular}
& 1,024 \\
\bottomrule
\end{tabular}
}

\vspace{0.1cm}
\caption{
    \textbf{Transition matrix of First Failure Point (FFP) stages between GRPO and MRPO for GRPO-only-correct instances.}
    Rows indicate the GRPO FFP stage and columns the MRPO FFP stage. The MRPO SUM row is highlighted, with percentages computed over the total number of samples.
}
\label{tab:grpo-mrpo-ffp-transition}
\vspace{-0.3cm}
\end{table}

Table~\ref{tab:grpo-mrpo-ffp-transition} reports the symmetric analysis on the GRPO-only-correct group, characterizing the failures MRPO newly introduces. Since GRPO answers these correctly, the question is at which stage MRPO's failures occur when a failure point is detected. As shown in the highlighted SUM row, MRPO's failures concentrate in the late stage at 24.9\%, exceeding the mid (18.8\%) and early (8.7\%) stages, while many traces remain free of any detected failure point at 47.6\%. In other words, when MRPO answers incorrectly, the failure tends to occur late in the trajectory rather than as an early failure triggering a cascade. Taken together, the two matrices reveal a consistent asymmetry aligned with MRPO's accuracy gain: the instances it newly answers correctly are those where it removes a GRPO early-stage failure, while the failures it introduces shift to the non-propagating late stage.

\section{Qualitative Analysis}
\label{app:qualitative_analysis}

\subsection{Stage-wise failure taxonomy}
\label{app:failure_taxonomy}

We conduct reasoning evaluation on the test sets of VQA-RAD, SLAKE, and PathVQA across four backbones (Qwen2.5-VL-7B-Instruct, Qwen3-VL-8B-Instruct, Qwen3-VL-8B-Thinking, and InternVL3-8B-Instruct). We randomly sample 30 failed traces from each First Failure Point (FFP) bin of width 0.1, yielding 300 traces, and the authors manually examine the first invalid step of each. We find that traces with similar FFP positions exhibit similar failure patterns and error types, allowing them to be grouped into three contiguous ranges, 0.0-0.4, 0.4-0.7, and 0.7-1.0, with largely homogeneous error types within each. We refer to these three ranges as the early, mid, and late stages, respectively. The taxonomy reveals that failure types are clearly stratified by stage, and their position determines how severely they affect the final answer: early-stage failures correspond to errors in establishing the visual premise and mostly trigger a cascade that corrupts the entire downstream trajectory and leads to an incorrect answer; mid-stage failures arise during the interpretation of correctly perceived content and propagate only partially, often still permitting recovery in later steps; and late-stage failures are confined to terminological expression and rarely affect the validity of the preceding reasoning. We describe the failure types and patterns at each stage below.

\begin{figure*}[t]
  \centering
  \includegraphics[width=\textwidth]{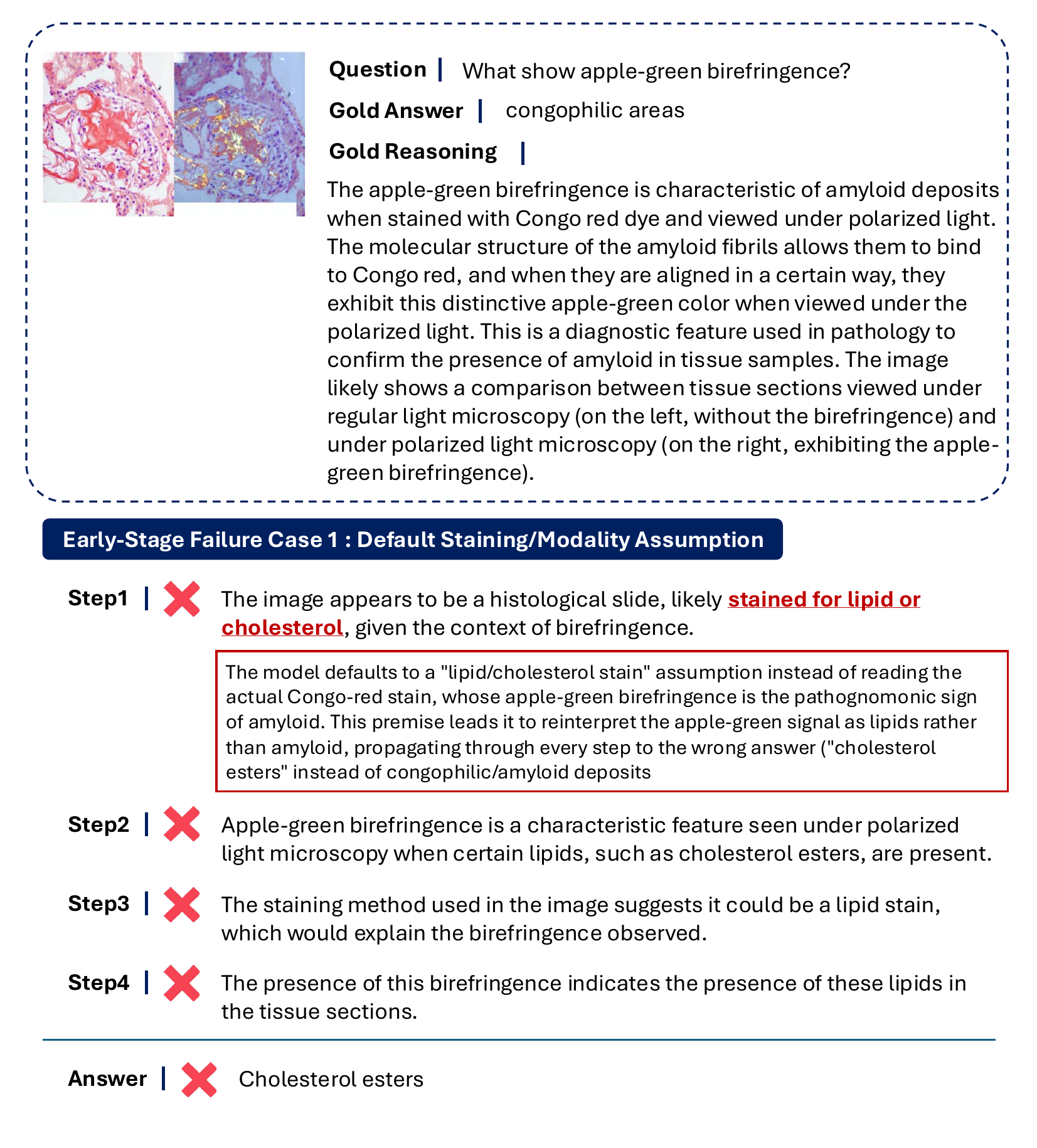}
  \caption{
    \textbf{Early-stage failure: Default Staining/Modality Assumption.}
    The model defaults to the most common modality (e.g., H\&E staining) instead of reading the actual image, grounding the entire trace in an unverified premise that corrupts all subsequent steps.
  }
  \label{fig:failure_1_early_1}
  \vspace{-0.3cm}
\end{figure*}

\begin{figure*}[t]
  \centering
  \includegraphics[width=\textwidth]{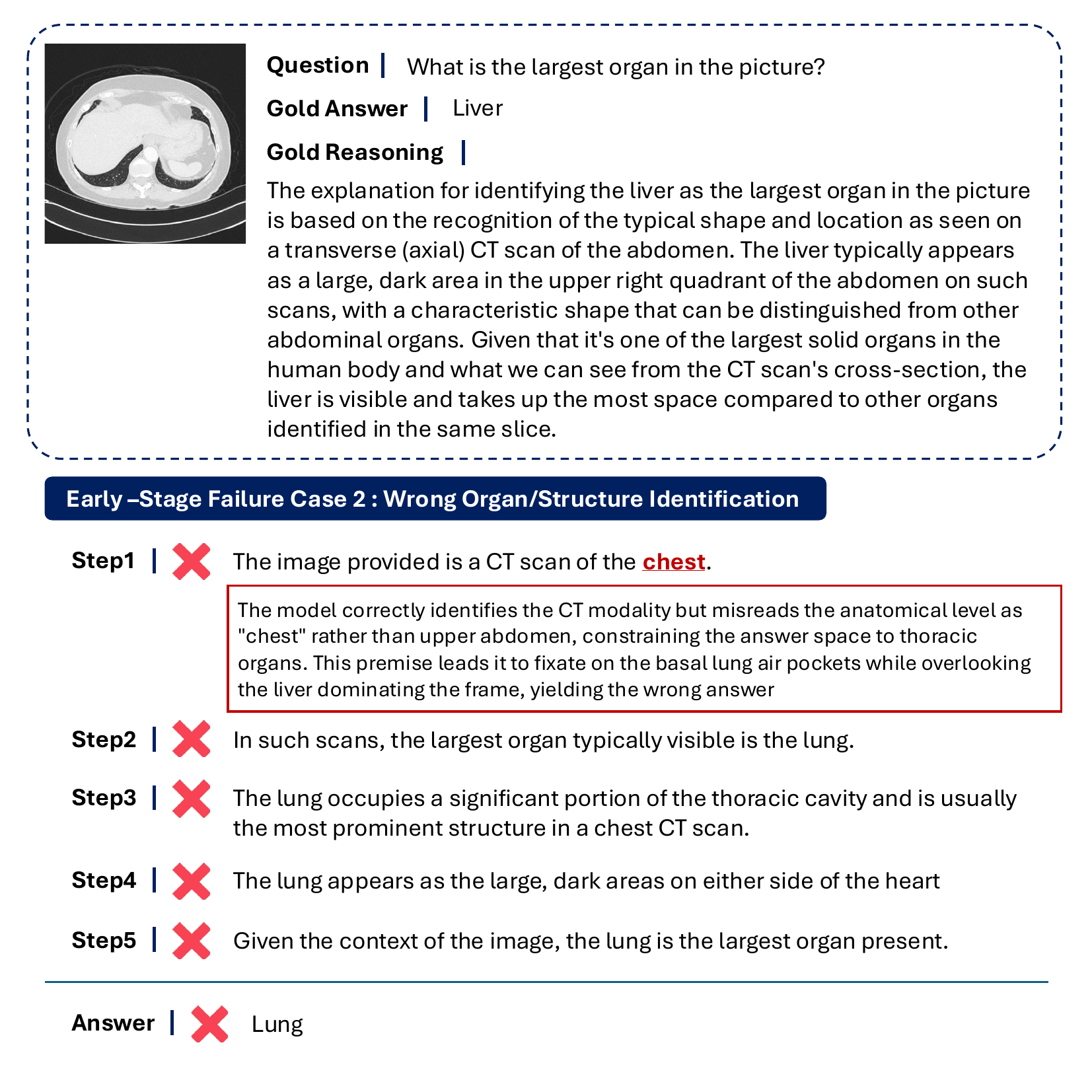}
  \caption{
    \textbf{Early-stage failure: Wrong Organ/Structure Identification.}
    The model recognizes the image type correctly but, in its first sentence, confidently specifies an incorrect organ or anatomical structure, leading the downstream reasoning to cascade into a wrong answer.
  }
  \label{fig:failure_1_early_2}
  \vspace{-0.3cm}
\end{figure*}

\paragraph{Early-stage errors.} Early-stage failures occur when the model fails to establish a correct visual premise at the outset of reasoning, corrupting every subsequent step that builds upon it. We identify two types.
\textbf{(1) Default Staining/Modality Assumption (Figure~\ref{fig:failure_1_early_1}):} rather than reading the actual image, the model begins reasoning by defaulting to the most common modality (e.g., H\&E staining), grounding the entire trace in an unverified assumption.
\textbf{(2) Wrong Organ/Structure Identification (Figure~\ref{fig:failure_1_early_2}):} the model correctly recognizes the image type but, in its first sentence, confidently specifies an incorrect organ or anatomical structure.

\begin{figure*}[t]
  \centering
  \includegraphics[width=\textwidth]{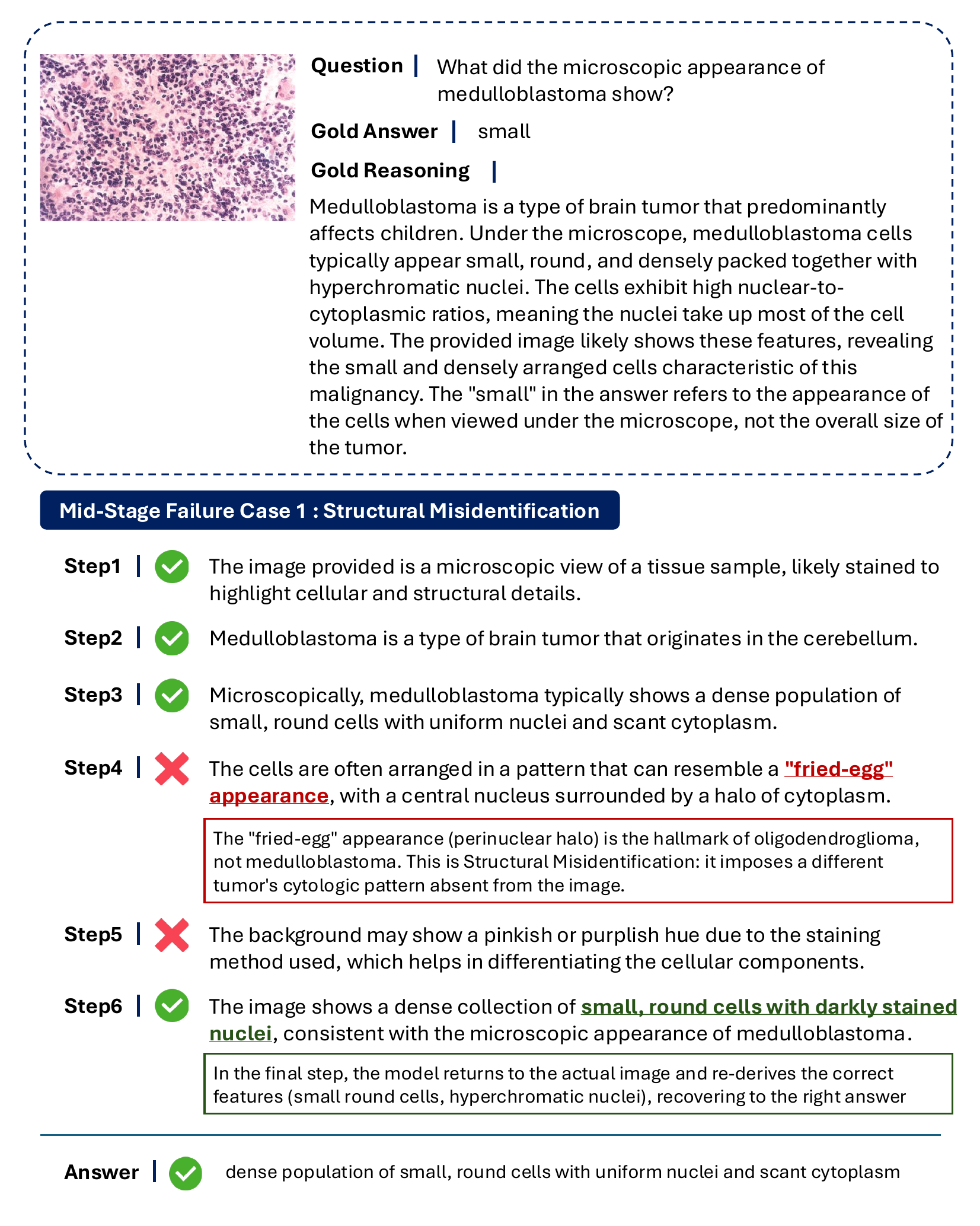}
  \caption{
    \textbf{Mid-stage failure: Structural Misidentification.}
    After establishing a correct premise, the model misrecognizes a microstructure as a different organ or structure, corrupting only the steps that depend on this misjudgment while later steps may still recover.
  }
  \label{fig:failure_2_mid_1}
  \vspace{-0.3cm}
\end{figure*}

\begin{figure*}[t]
  \centering
  \includegraphics[width=\textwidth]{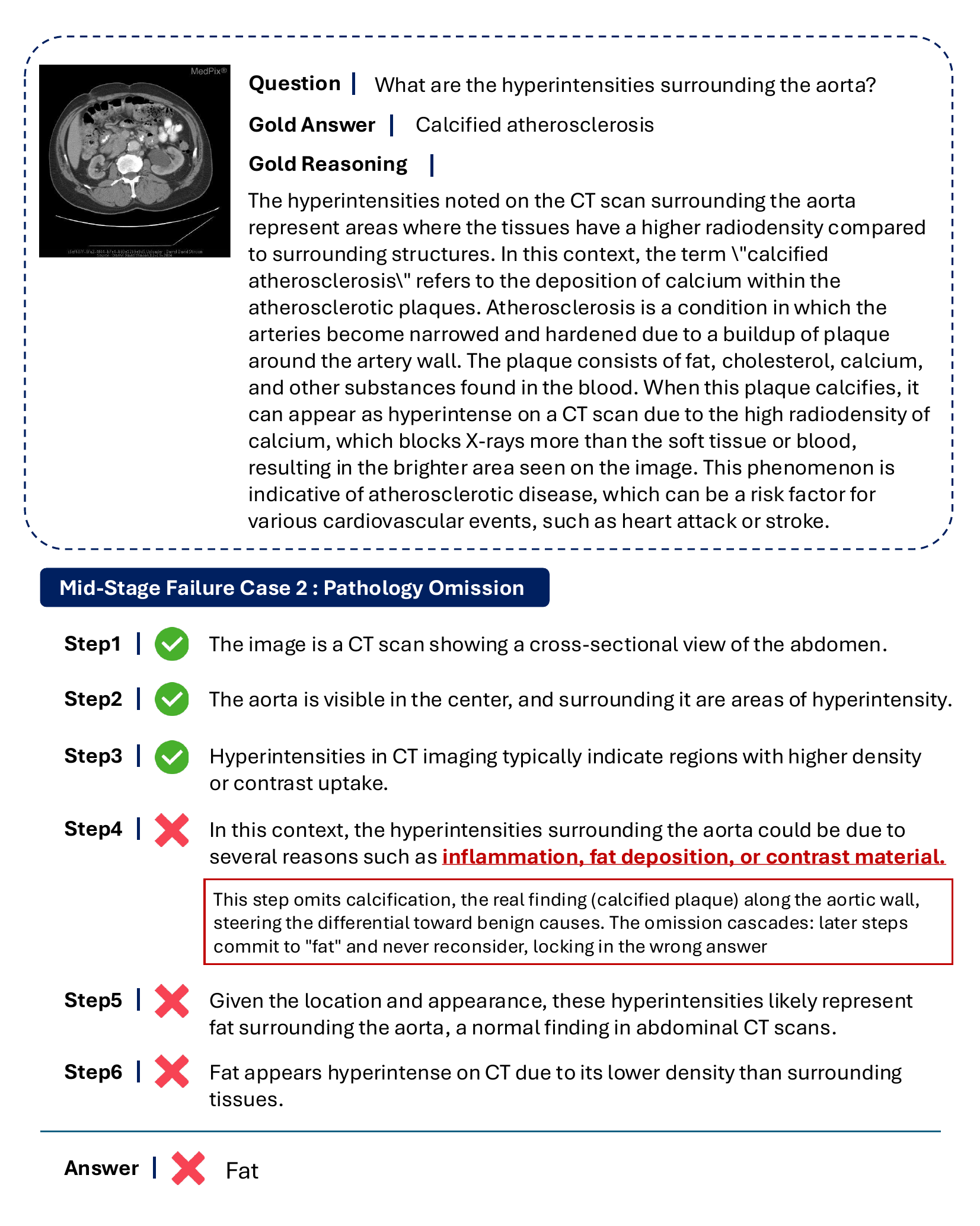}
  \caption{
    \textbf{Mid-stage failure: Pathology Omission.}
    Despite the presence of a lesion in the image, the model describes only normal findings, an interpretation error that arises after the visual premise is correctly established.
  }
  \label{fig:failure_2_mid_2}
  \vspace{-0.3cm}
\end{figure*}

\paragraph{Mid-stage errors.} Mid-stage failures arise after the visual premise is established, when the model engages with the image but errs in interpretation or diagnostic judgment. Because the premise is correct, these errors typically corrupt only the steps depending on the specific misjudgment rather than the entire trajectory, so later steps may still recover and reach the correct answer. We identify two types.
\textbf{(1) Structural Misidentification (Figure~\ref{fig:failure_2_mid_1}):} the model misrecognizes a microstructure as a different organ or structure.
\textbf{(2) Pathology Omission (Figure~\ref{fig:failure_2_mid_2}):} despite a lesion in the image, the model describes only normal findings.

\begin{figure*}[t]
  \centering
  \includegraphics[width=\textwidth]{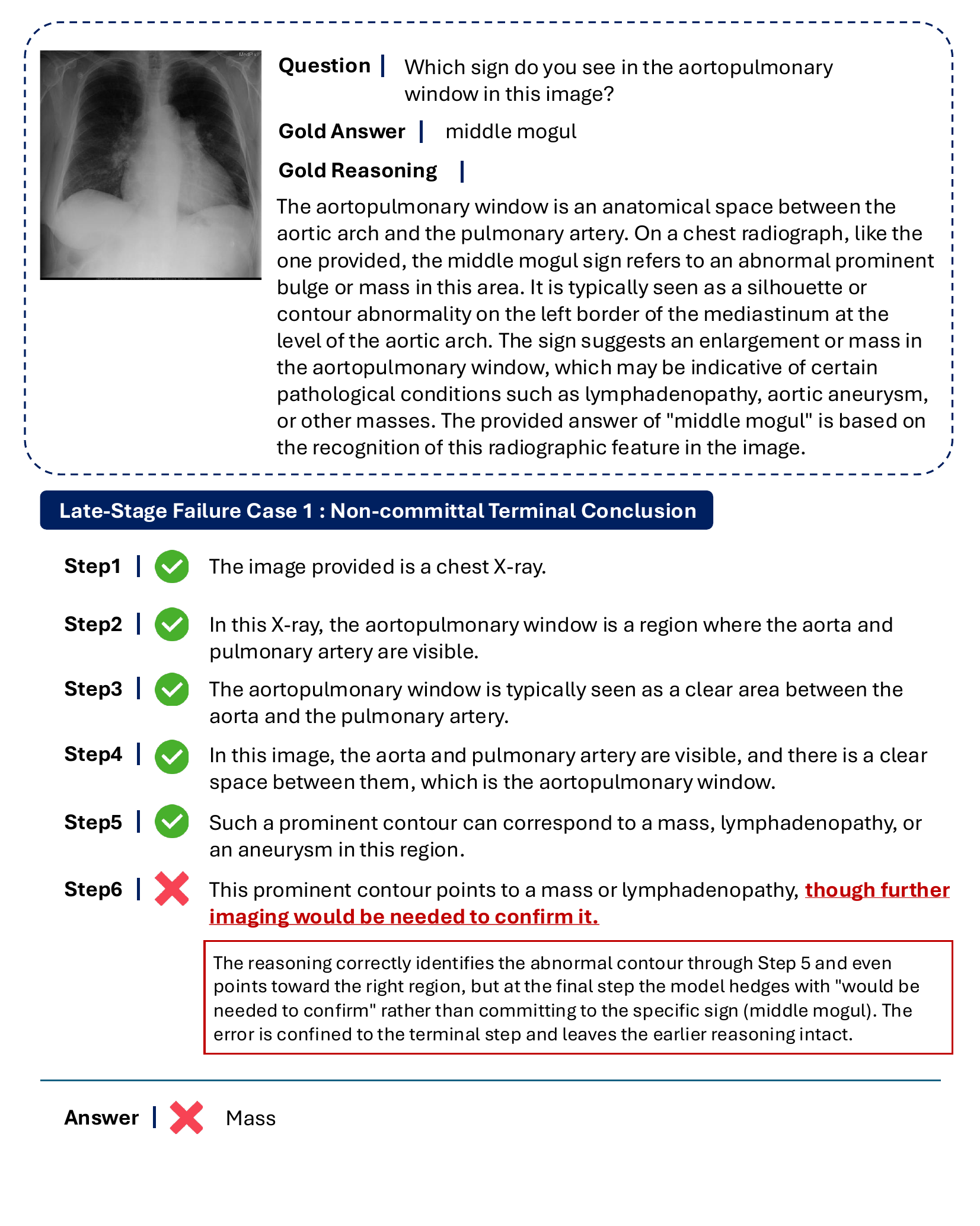}
  \caption{
    \textbf{Late-stage failure: Non-committal Terminal Conclusion.}
    The reasoning proceeds correctly until the late steps, but the model fails to converge on a specific answer and hedges with vague expressions such as ``consistent with'' or ``possibly,'' avoiding a definitive conclusion.
  }
  \label{fig:failure_3_late_1}
  \vspace{-0.3cm}
\end{figure*}

\begin{figure*}[t]
  \centering
  \includegraphics[width=\textwidth]{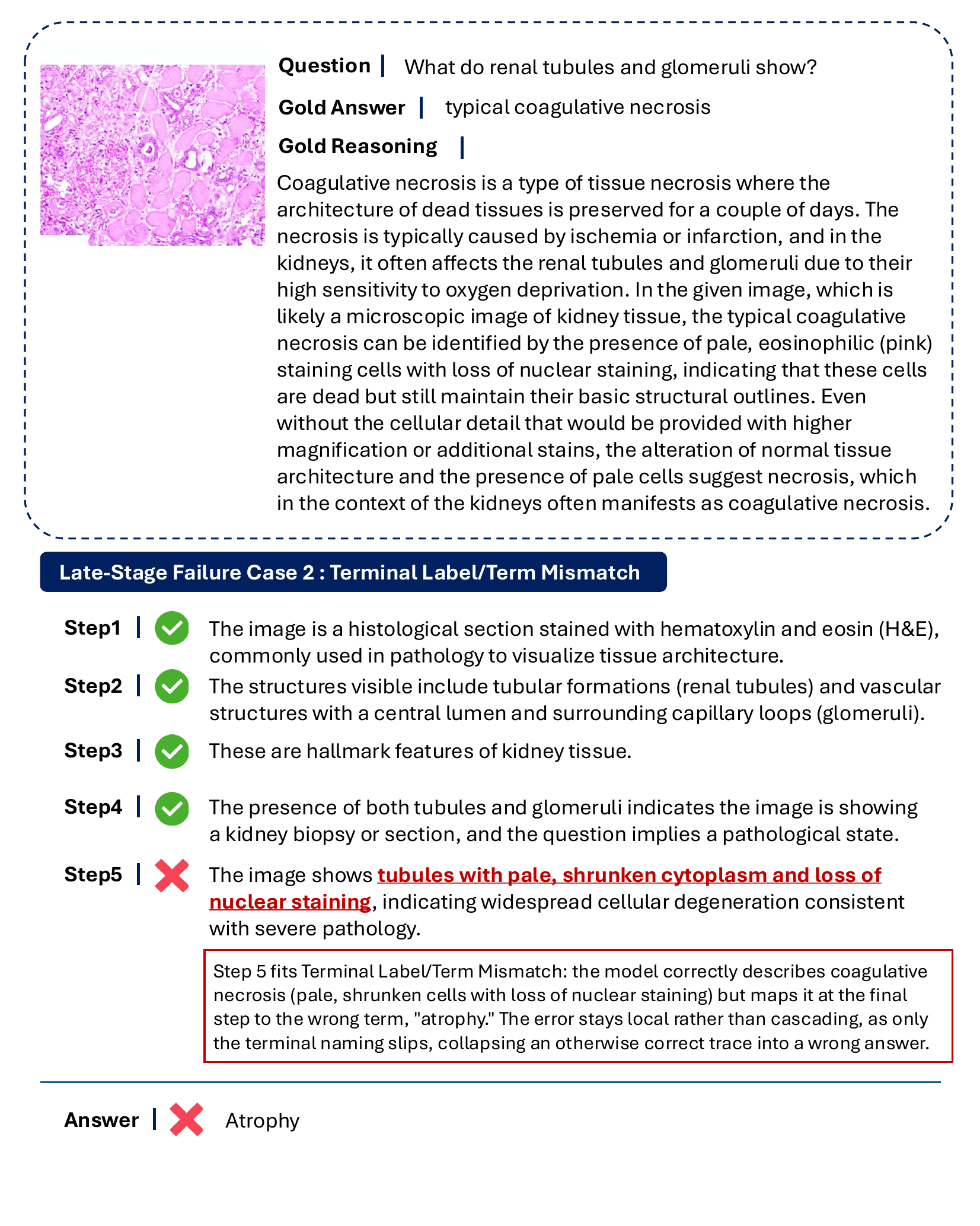}
  \caption{
    \textbf{Late-stage failure: Terminal Label/Term Mismatch.}
    Having correctly identified the relevant structure or finding, the model mismaps it at the final step to an incorrect name, laterality, or specific term, while the preceding reasoning remains intact.
  }
  \label{fig:failure_3_late_2}
  \vspace{-0.3cm}
\end{figure*}

\paragraph{Late-stage errors.} Late-stage failures occur after the model has correctly perceived the image and carried out largely valid reasoning, with the error confined to the final step. Because the preceding reasoning remains intact, these failures rarely propagate or induce a cascade; they instead reflect a local breakdown at the point of producing the final answer. We identify two types.
\textbf{(1) Non-committal Terminal Conclusion (Figure~\ref{fig:failure_3_late_1}):} although the reasoning proceeds correctly until the late steps, the model fails to converge on a specific answer and instead hedges with vague expressions such as "consistent with" or "possibly".
\textbf{(2) Terminal Label/Term Mismatch (Figure~\ref{fig:failure_3_late_2}):} having correctly identified the relevant structure or finding, the model mismaps it at the final step to an incorrect name, laterality (left/right), or specific term.

\subsection{Qualitative Case Studies}
\label{app:case_studies}

To complement the quantitative analyses, we qualitatively compare GRPO and MRPO reasoning traces on identical inputs. Each case pairs the two methods on the same question with every step annotated as valid or invalid, so that the divergence point is visible. We organize the cases around three mechanisms: \textbf{Case 1.} correcting early failures before they cascade, \textbf{Case 2.} recovering from early failures when they occur, and \textbf{Case 3.} the nature of the residual failures MRPO introduces.

\begin{figure*}[t]
  \centering
  \includegraphics[width=\textwidth]{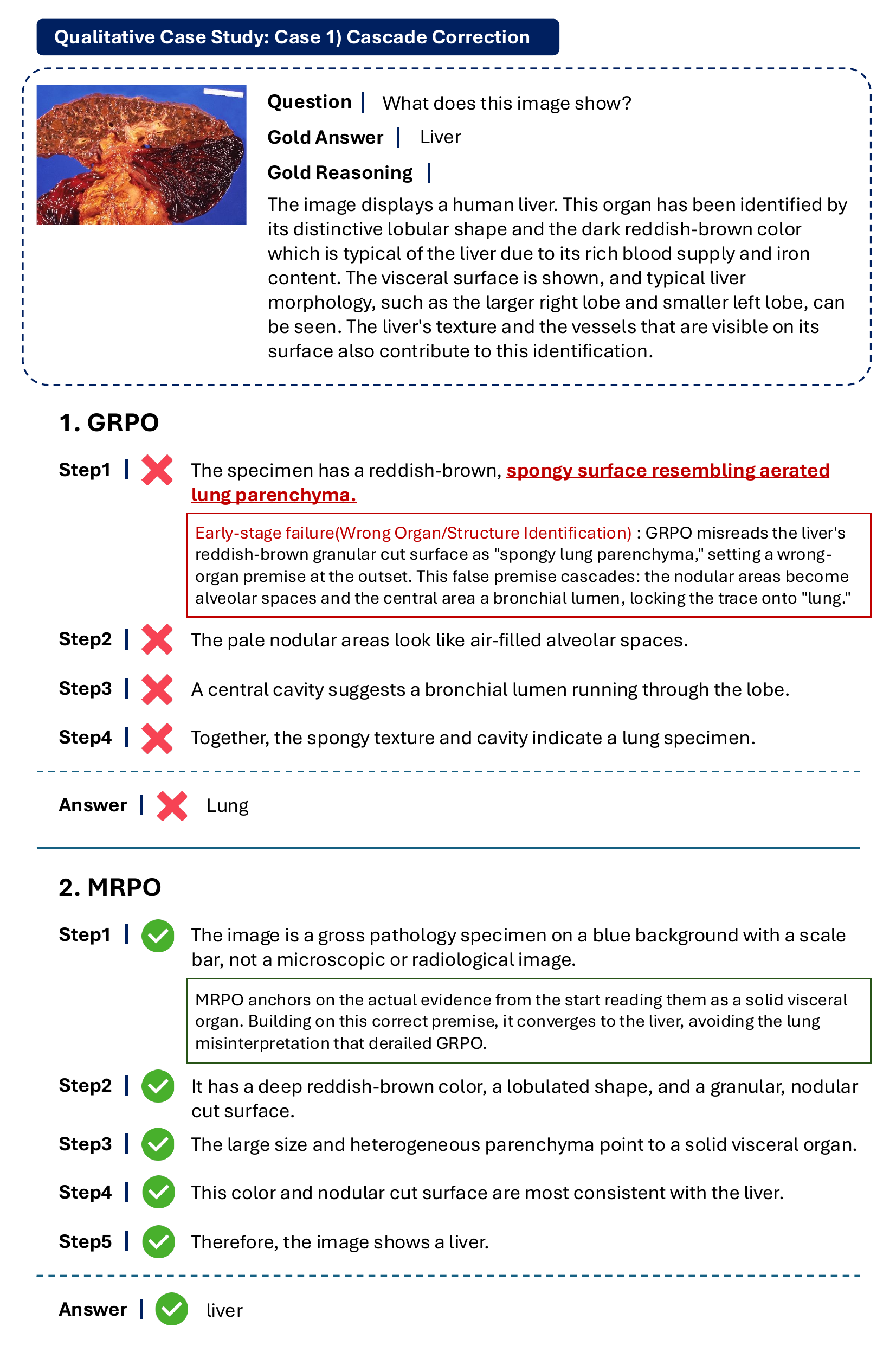}
  \caption{
    \textbf{Case 1: Cascade correction.}
    GRPO's incorrect premise in the first step cascades into a wrong organ identification, while MRPO anchors on the correct visual evidence and reaches the right answer.
  }
  \label{fig:failure_case_study_1_early_1}
  \vspace{-0.3cm}
\end{figure*}

\paragraph{Case 1: Cascade Correction.} The most direct effect of MRPO is correcting the early-stage error GRPO commits at the outset, before it propagates. Here GRPO establishes an incorrect visual premise in its first step, such as a Wrong Organ/Structure Identification, and every subsequent step inherits it, locking the trace onto a wrong answer. Given the same input, MRPO instead grounds its opening steps in the actual visual evidence, so the remaining reasoning builds on a correct premise and converges to the right answer. Figure~\ref{fig:failure_case_study_1_early_1} shows a gross pathology specimen where GRPO misreads the liver's reddish-brown granular cut surface as spongy lung parenchyma and cascades into an incorrect organ identification, answering Lung, while MRPO anchors on the correct premise of a solid visceral organ and recovers the right answer, Liver. This reflects the core mechanism behind MRPO's reduction of early-stage failures (Section~\ref{sec:reasoning_quality_analysis}): suppressing the first misperception suffices to redirect the entire trajectory, since downstream reasoning is conditioned on the corrected premise.

\begin{figure*}[t]
  \centering
  \includegraphics[width=\textwidth]{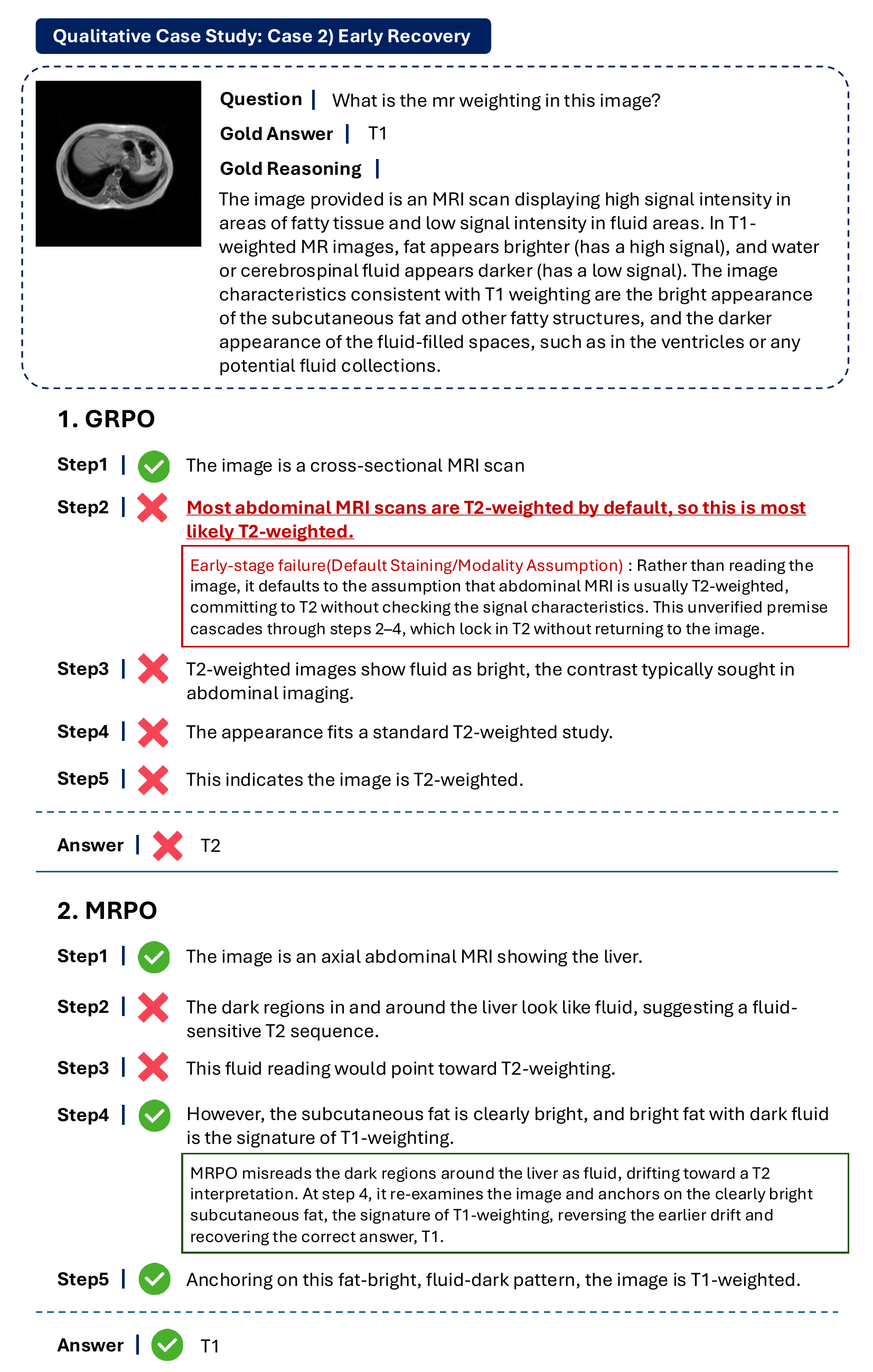}
  \caption{
    \textbf{Case 2: Early recovery.}
    GRPO defaults to a T2 assumption without inspecting the image and MRPO misreads dark regions as fluid; GRPO locks in the error, while MRPO re-anchors on the T1-characteristic bright subcutaneous fat and recovers the correct answer.
  }
  \label{fig:failure_case_study_2_early_1}
  \vspace{-0.3cm}
\end{figure*}

\paragraph{Case 2: Early Recovery.} Beyond preventing early failures, MRPO can also recover from a failure once it occurs, which GRPO rarely does. In these cases both methods reach the same wrong answer early in the trace, but for different reasons and with diverging subsequent behavior. GRPO commits to the mistaken premise and every following step inherits it, whereas MRPO re-examines the image and reverses the error before it reaches the answer. Figure~\ref{fig:failure_case_study_2_early_1} illustrates this on an MRI-weighting question. Both traces initially settle on T2 but for different reasons. GRPO defaults to the assumption that abdominal MRI is typically T2-weighted without inspecting the image, while MRPO engages the image but misreads the dark peri-hepatic regions as fluid. GRPO never returns to the image and locks in the wrong answer, whereas MRPO re-examines the scan, anchors on the clearly bright subcutaneous fat that signals T1-weighting, and reverses its drift to recover the correct answer. This behavior is the case-level counterpart of MRPO's lower Failure Accumulation Rate in the early bins (Section~\ref{sec:reasoning_quality_analysis}). Even when an early error arises, the step-wise penalty discourages the model from compounding it, allowing later steps to correct course rather than propagate the mistake.

\begin{figure*}[t]
  \centering
  \includegraphics[width=\textwidth]{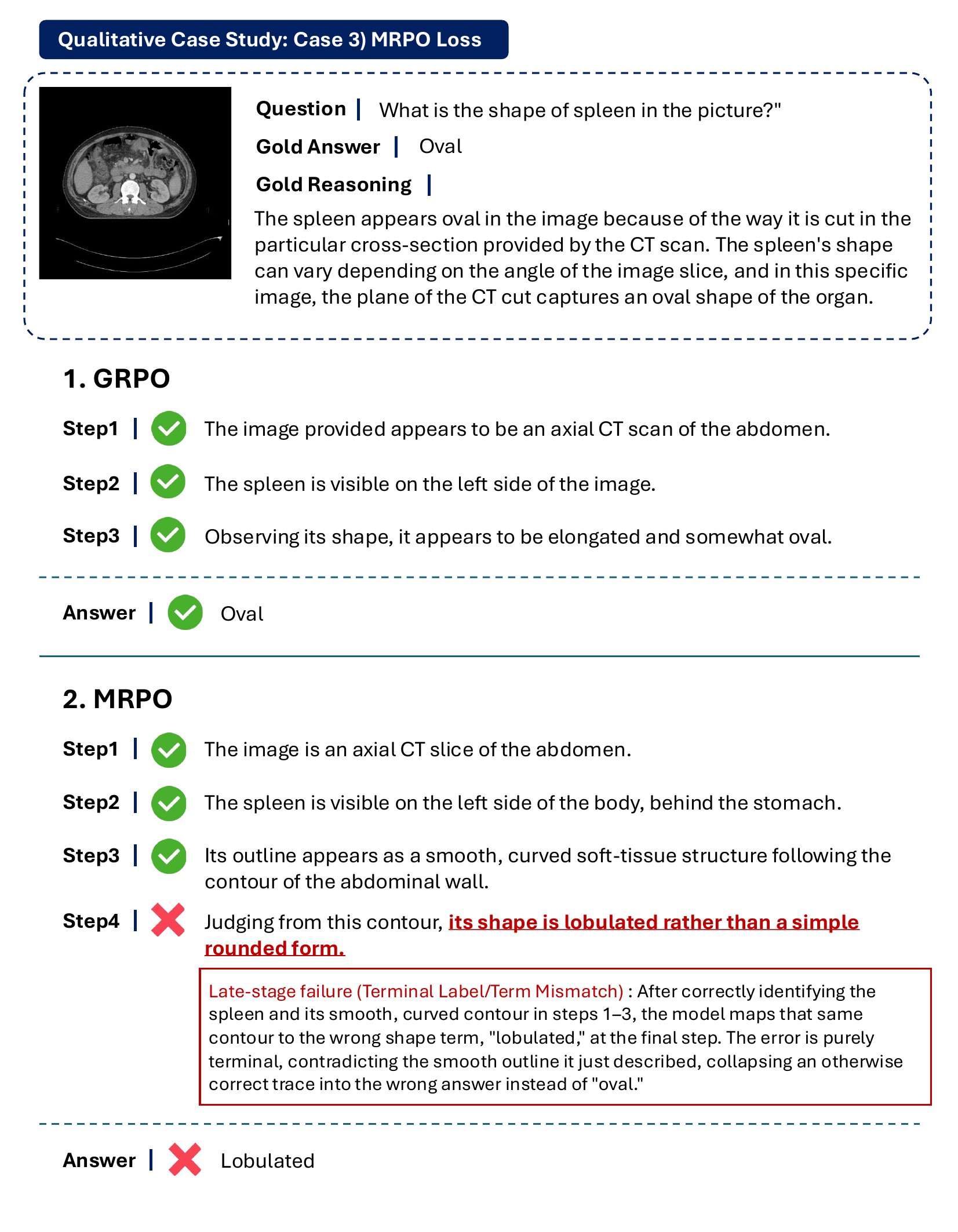}
  \caption{
    \textbf{Case 3: MRPO loss.}
    MRPO correctly identifies and describes the spleen but labels its shape as ``lobulated'' at the final step, a terminal term mismatch that leaves the preceding reasoning intact.
  }
  \label{fig:failure_case_study_3_early_1}
  \vspace{-0.3cm}
\end{figure*}

\paragraph{Case 3: MRPO Loss.} For completeness, we also examine the instances that GRPO answers correctly but MRPO does not. Consistent with the paired analysis in Appendix~\ref{app:reasoning_analysis_paired_comparison}, where MRPO's losses concentrate in the late stage, these failures are typically not cascading errors but local breakdowns at the final step, after the reasoning has otherwise proceeded correctly. Here MRPO perceives the image and reasons validly, but commits a Terminal Label/Term Mismatch when producing the answer, mapping a correctly described finding to the wrong term. Figure~\ref{fig:failure_case_study_3_early_1} shows a representative instance, where MRPO correctly identifies the spleen and describes its smooth, curved contour, yet labels the shape as ``lobulated'' at the final step, contradicting its own description and yielding the wrong answer where GRPO succeeds. Such losses indicate that MRPO's residual errors stem from terminal terminology rather than corrupted reasoning, leaving the preceding trajectory intact and far less likely to cascade. This points to refining terminal answer grounding, rather than the reasoning process itself, as a direction for further improvement.

\section{RL Training Plots}
\label{app:RL_training_plots}

\begin{figure*}[t]
  \centering
  \includegraphics[width=\textwidth]{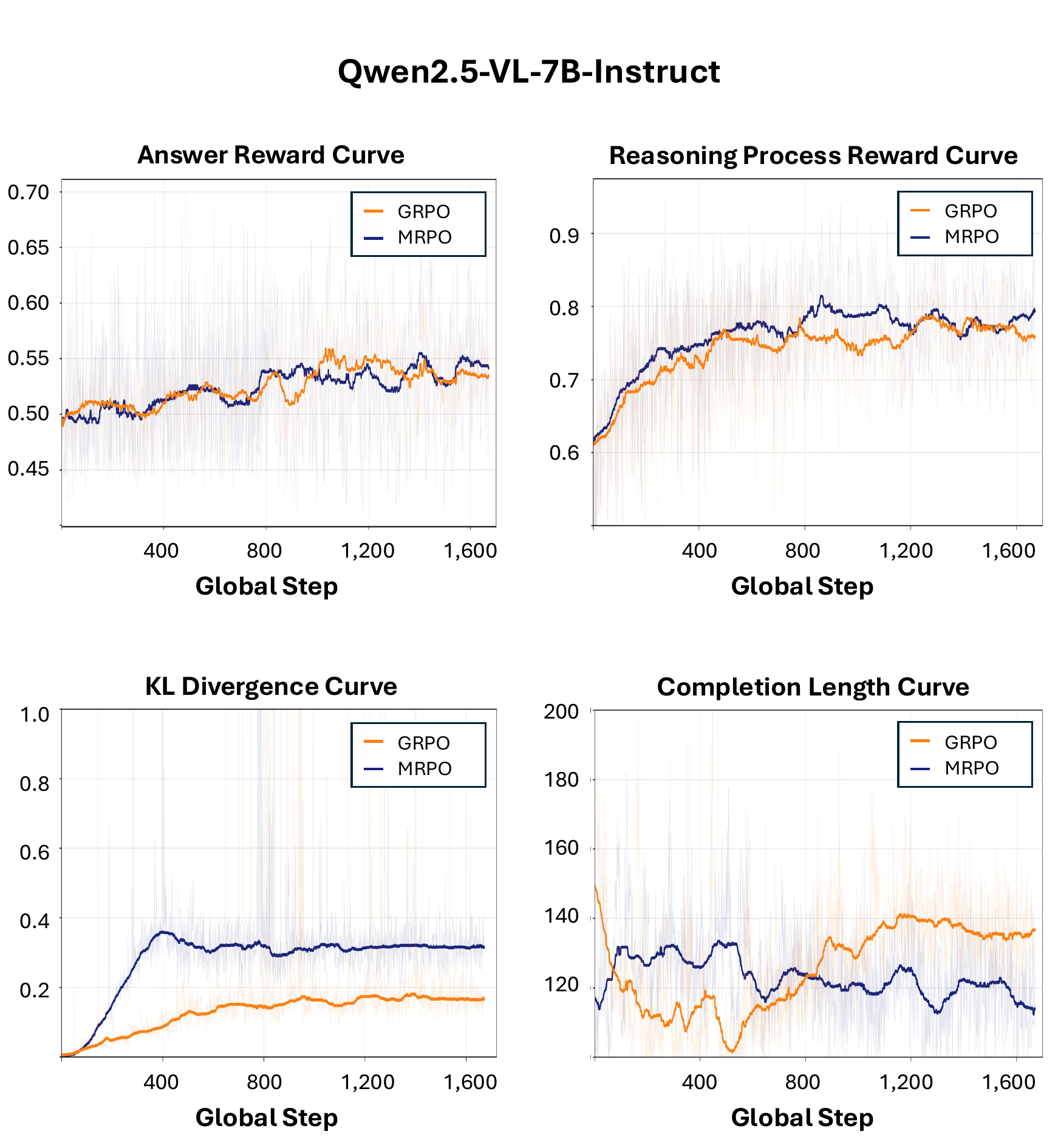}
  \caption{
    \textbf{Training dynamics of GRPO and MRPO on Qwen2.5-VL-7B-Instruct.}
    Curves show the answer reward, reasoning process reward, KL divergence, and completion length over training.
  }
  \label{fig:Training_Plot_1_1_new}
  \vspace{-0.3cm}
\end{figure*}

\begin{figure*}[t]
  \centering
  \includegraphics[width=\textwidth]{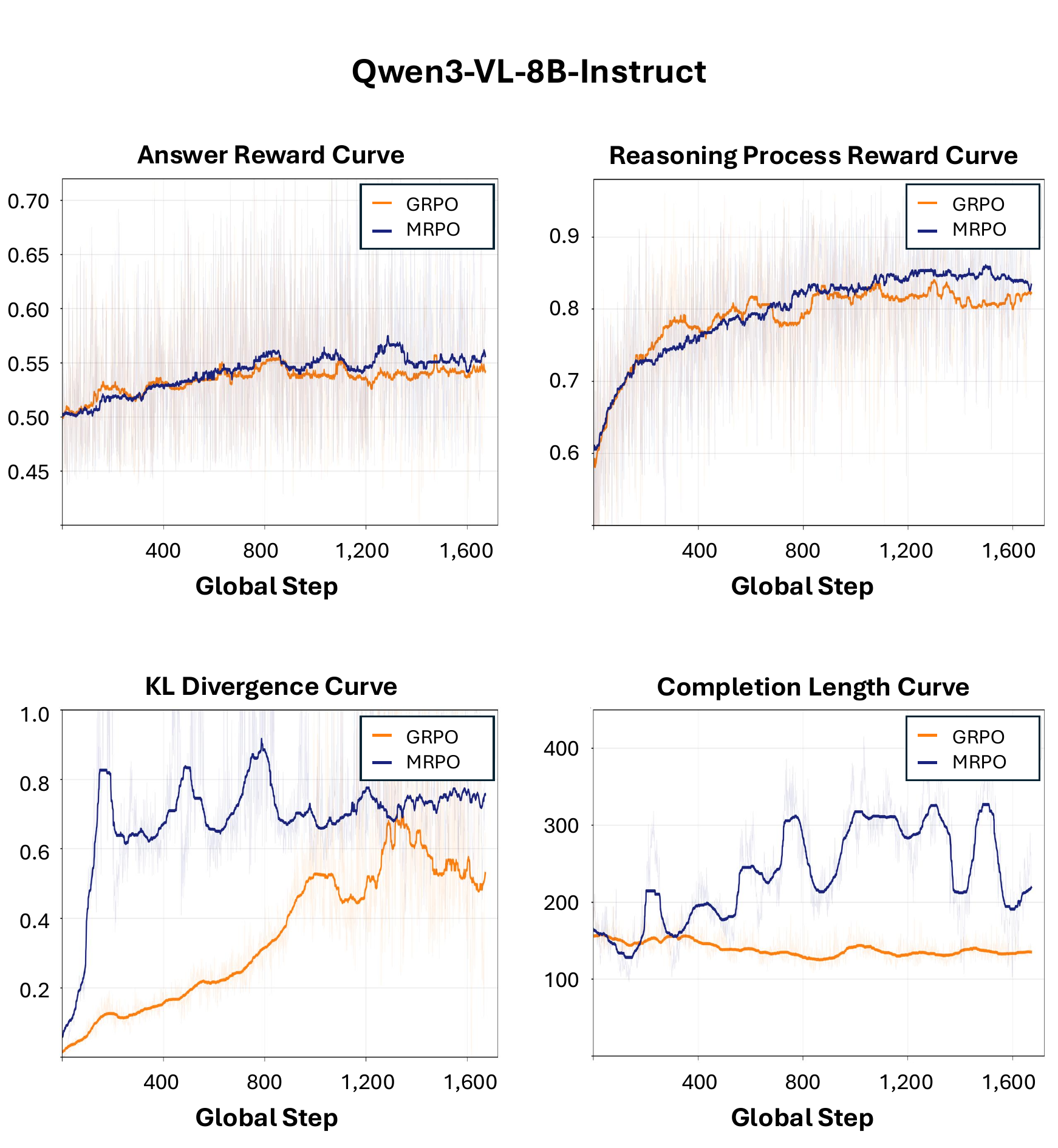}
  \caption{
    \textbf{Training dynamics of GRPO and MRPO on Qwen3-VL-8B-Instruct.}
    Curves show the answer reward, reasoning process reward, KL divergence, and completion length over training.
  }
  \label{fig:Training_Plot_2_1_new}
  \vspace{-0.3cm}
\end{figure*}

\begin{figure*}[t]
  \centering
  \includegraphics[width=\textwidth]{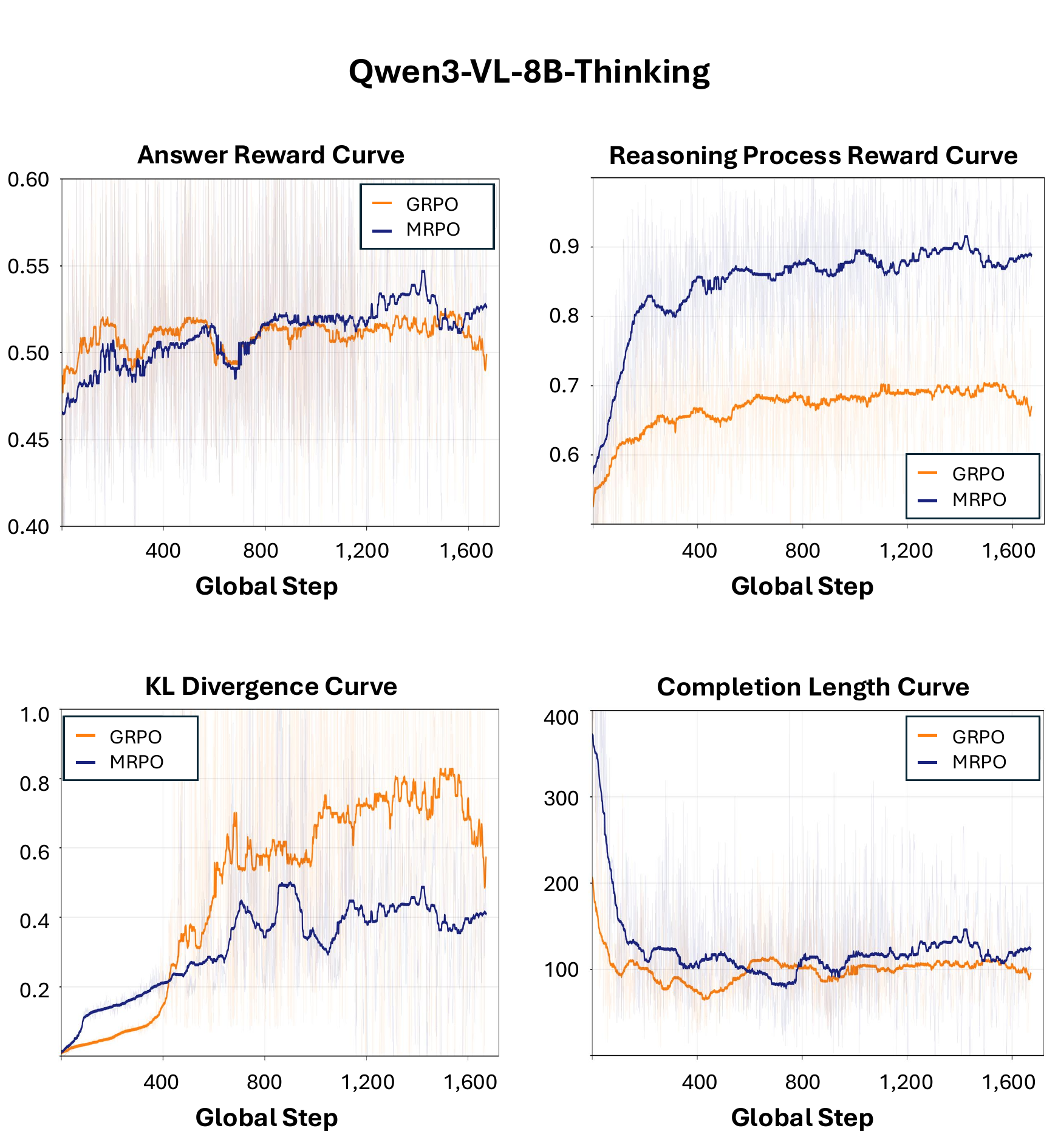}
  \caption{
    \textbf{Training dynamics of GRPO and MRPO on Qwen3-VL-8B-Thinking.}
    Curves show the answer reward, reasoning process reward, KL divergence, and completion length over training.
  }
  \label{fig:Training_Plot_2_2_new}
  \vspace{-0.3cm}
\end{figure*}

\begin{figure*}[t]
  \centering
  \includegraphics[width=\textwidth]{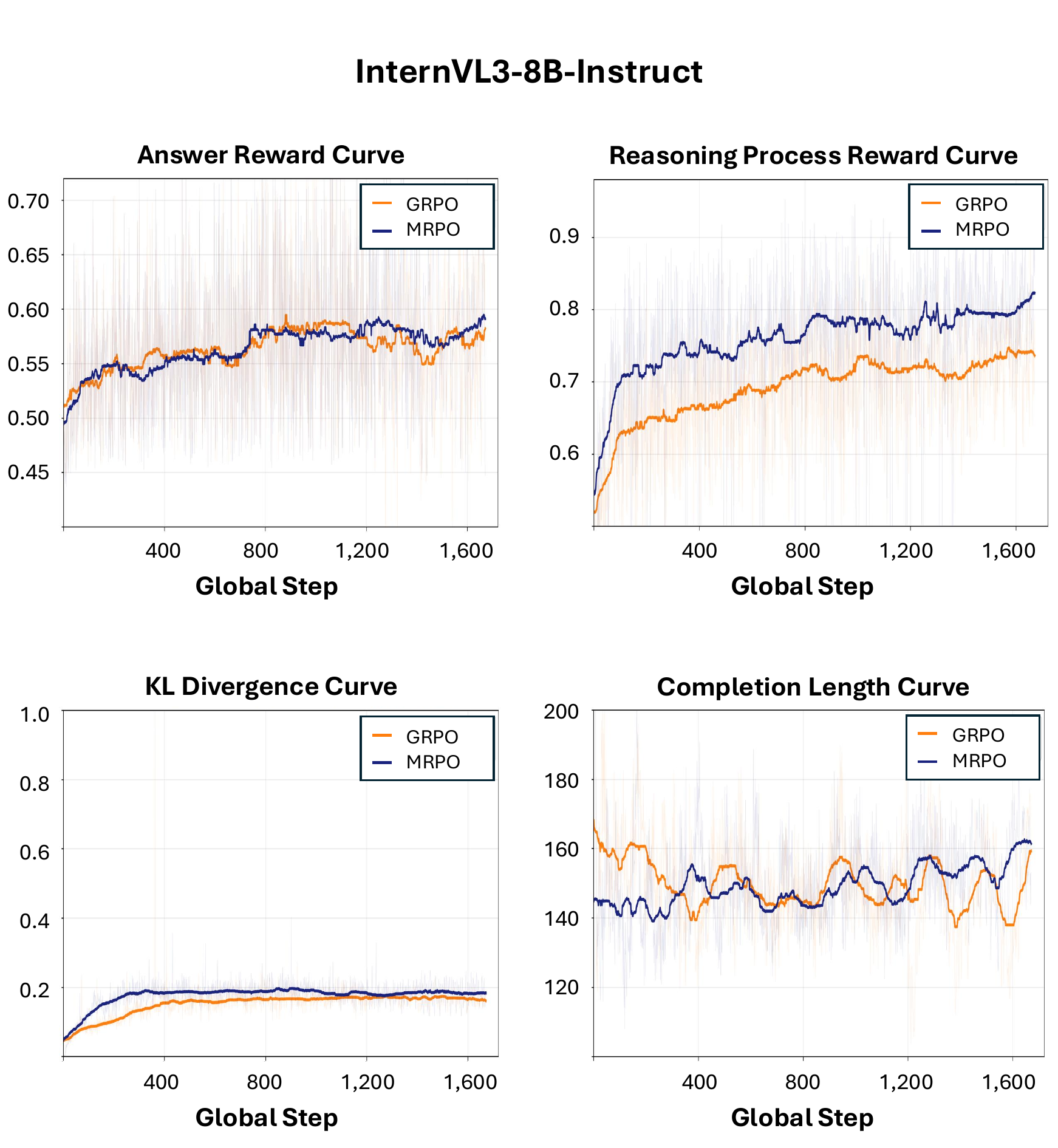}
  \caption{
    \textbf{Training dynamics of GRPO and MRPO on InternVL3-8B-Instruct.}
    Curves show the answer reward, reasoning process reward, KL divergence, and completion length over training.
  }
  \label{fig:Training_Plot_3_1_new}
  \vspace{-0.3cm}
\end{figure*}

We provide training-dynamics plots for all four backbones, Qwen2.5-VL-7B-Instruct (Figure~\ref{fig:Training_Plot_1_1_new}), Qwen3-VL-8B-Instruct (Figure~\ref{fig:Training_Plot_2_1_new}), Qwen3-VL-8B-Thinking (Figure~\ref{fig:Training_Plot_2_2_new}), and InternVL3-8B-Instruct (Figure~\ref{fig:Training_Plot_3_1_new}), covering answer reward, reasoning process reward, KL divergence, and completion length.

Across the four backbones, MRPO holds a slight edge over GRPO on answer reward and a clearer margin on reasoning process reward.
KL divergence runs somewhat higher for MRPO, a natural consequence of its stronger step-wise advantage reshaping, but in all cases it rises early and then settles rather than diverging, so training remains stable. Completion length varies by backbone and shows no consistent ordering except on Qwen3-VL-8B-Instruct, where MRPO produces clearly longer completions.

\section{Prompts}
\label{app:appendix_prompts}

\subsection{Answer Correctness Check Prompt}
\label{app:appendix_prompts_answer}

% \begin{figure}[t]
% \centering
\begin{tcolorbox}[
    enhanced,
    breakable,
    width=\columnwidth,
    colback=gray!3,
    colframe=black,
    colbacktitle=black,
    coltitle=white,
    fonttitle=\bfseries,
    title=Answer Correctness Check Prompt,
    boxrule=0.8pt,
    arc=2pt,
    left=8pt,
    right=8pt,
    top=6pt,
    bottom=6pt
]
Given a question about a medical image, there is a correct answer to the question and an answer to be determined.

\medskip
If the answer to be determined matches the correct answer or is a good enough answer to the question, output `O`; otherwise output `X`. Evaluate the answer to be determined (`O` or `X`).

\medskip
\textbf{Question}

\-\- - question about the medical image: \{problem\}

\-\- - Image: \{image\}

\medskip
\textbf{Answers}

\-\- - correct answer (ground truth): \{solution\}

\-\- - answer to be determined: \{generated answer\}

\medskip
Your response must be a single character: `O` (correct) or `X` (incorrect).
\end{tcolorbox}

% \captionof{figure}{
% \textbf{Answer correctness check prompt.}
% Prompt used to evaluate whether a generated answer matches the ground-truth answer.
% }
% \label{fig:appendix_prompts_answer}
% \end{figure}

\subsection{Step-wise Reasoning Evaluation Prompt}
\label{app:appendix_process_reward_prompt}

% \begin{figure}[t]
% \centering
\begin{tcolorbox}[
    enhanced,
    breakable,
    width=\columnwidth,
    colback=gray!3,
    colframe=black,
    colbacktitle=black,
    coltitle=white,
    fonttitle=\bfseries,
    title=Step-wise Reasoning Evaluation Prompt,
    boxrule=0.8pt,
    arc=2pt,
    left=8pt,
    right=8pt,
    top=6pt,
    bottom=6pt
]

You are a medical reasoning verification module.

\medskip

\section*{ Task}

\noindent You will be given:
\medskip

- Image: A medical image (provided separately)

- Problem: A medical VQA question about the image

- Ground\_Truth: The correct answer to the problem

- Gold\_Reasoning: Gold reasoning steps (reference standard)

- Reasoning\_Sentences: Generated reasoning sentences to evaluate

\medskip
For EACH generated sentence, output:
\medskip

- Gold Alignment: 0/1

- Answer Contribution: 0/1

\medskip
\section*{ General Rules (Apply First)}
\medskip

\textbf{Automatic 0 for BOTH Gold Alignment and Answer Contribution:}

- Meta-Commentary: "I will identify...", "Consider the possibilities..."

- Empty or meaningless statements

- Pure repetition of previous steps without new content

\medskip
Check these rules FIRST before evaluating Gold Alignment and Answer Contribution.

\medskip
\medskip
\section*{ Evaluation Criteria}

\medskip
\medskip
\noindent \textbf{\#\#\# Gold Alignment (Gold\_Reasoning Consistency)}
\medskip
\medskip

\textbf{First, extract Key Elements from Gold\_Reasoning:}

\-\- - Modality, Context, Key Findings, Anatomical Location (including laterality), Diagnostic Direction

\medskip
"Is this step consistent with Gold\_Reasoning?"
\medskip

Gold Alignment ONLY checks whether the step matches Gold\_Reasoning, regardless of contribution to answer.

\medskip
\medskip
- \textbf{Gold Alignment = 1:}
\medskip

\-\- - Early (First 1-2 steps) : Correctly identifies modality (e.g., "X-ray", "CT", "MRI", "electron microscopy"), tissue type (e.g., "histological section", "gross specimen"), or staining method (e.g., "H\&E", "immunostain") that matches Gold\_Reasoning

\-\- - Middle (Middle steps) : Identifies specific findings mentioned in Gold\_Reasoning, including: abnormalities, pathological features, key structures, AND correct anatomical location/laterality. Must match Gold's level of specificity.

\-\- - Later (Last 1-2 steps) : Reaches or clearly approaches the same diagnostic conclusion as Gold\_Reasoning. Must demonstrate diagnostic reasoning toward Ground\_Truth, not just restate observations.

\-\- - If Gold\_Reasoning contains diagnostic conclusions (e.g., "tuberculosis", "adenocarcinoma", "hemorrhage", "fracture"), generated reasoning MUST progress toward that diagnosis to get Alignment = 1
  
\medskip
\medskip
- \textbf{Gold Alignment = 0:}
\medskip

\-\- - Wrong Location/Laterality: Gold says "LEFT" but generated says "right"

\-\- - Contradiction: Directly contradicts Gold\_Reasoning

\-\- - Misdirection: Different diagnostic direction than Gold

\-\- - Content not mentioned or supported by Gold\_Reasoning

\-\- - Missing Critical Findings: Gold identifies pathological/abnormal findings but generated only describes normal or generic features

\-\- - Specificity Mismatch: Gold is specific (e.g., "lymphoma", "abnormal features") but generated is generic (e.g., "cellular structure", "tissue")

\-\- - If Gold identifies specific pathology (e.g., "granuloma", "infarction", "metastasis", "fibrosis") but generated only describes generic features (e.g., "tissue changes", "some abnormality", "lesion")

\-\- - Describing only normal-appearing structures when Gold identifies abnormalities

\medskip
\medskip
\noindent \textbf{\#\#\# Answer Contribution (Ground\_Truth Derivation)}
\medskip
\medskip

"Does this step directly help reach Ground\_Truth: '\{solution\}'?"

\medskip
\medskip
\textbf{Answer Contribution = 1:}
\medskip

\-\- - Directly mentions Ground\_Truth or semantically equivalent terms

\-\- - Identifies a finding that is explicitly required to derive Ground\_Truth

\-\- - States the specific diagnosis, location, or structure that matches Ground\_Truth

\medskip
\medskip
\textbf{Answer Contribution = 0:}
\medskip

\-\- - No direct relevance to Ground\_Truth

\-\- - Generic observation that applies to any image of this type

\-\- - Describes features not connected to Ground\_Truth

\-\- - Evasion: "Unknown", "cannot determine", "None"

\-\- - Context-only: states modality/setting without advancing toward Ground\_Truth

\medskip
\medskip
\section*{ Output Format (JSON ONLY):}
\medskip

Return JSON only, with this exact structure:

\medskip
\noindent \{\\
\- \- "Reasoning\_Check": \{\\
\- \- \- \- "step1": \{\\
\- \- \- \- \- \- "Gold Alignment": 1,\\
\- \- \- \- \- \- "Answer Contribution": 1\\
\- \- \- \- \- \},\\
\- \- \- \- "step2": \{ ... \}\\
\- \- \-\}\\
\-\}\\

\medskip
\medskip
\section*{ Rules:}
\medskip

\-\- - Apply General Rules FIRST (meta-commentary → both 0)

\-\- - Evaluate Gold Alignment and Answer Contribution INDEPENDENTLY

\-\- - Both values must be 0 or 1

\-\- - Do NOT output explanations outside JSON

\medskip
\medskip
\section*{ Inputs:}
\medskip

\-\- - "Problem": \{problem\}

\-\- - "Image": \{image\}

\-\- - Ground\_Truth: \{ground\_truth\}

\-\- - Gold\_Reasoning: \{gold\_reasoning\}

\-\- - "Reasoning\_Sentences": \{reasoning\_sentences\}
\end{tcolorbox}

% \captionof{figure}{
% \textbf{Answer correctness check prompt.}
% Prompt used to evaluate whether a generated answer matches the ground-truth answer.
% }
% \label{fig:appendix_process_reward_prompt}
% \end{figure}

\end{document}